\documentclass{book}
\usepackage[utf8]{inputenc}

\title{Extracting \& Learning a Dependency-Enhanced Type Lexicon for Dutch}
\author{kogkalidisk }
\date{May 2019}

\usepackage{xr}

\usepackage{hyperref}

\usepackage{amsmath}
\usepackage{amssymb}
\usepackage{mathtools}
\usepackage{proof}
\usepackage{stmaryrd}
\newlength{\arrow}
\newcommand*{\myrightarrow}[1]{\xrightarrow{\mathmakebox[\arrow]{\text{\scriptsize #1}}}}
\usepackage{rotating}

\usepackage{color}

\usepackage{graphicx}
\usepackage{float}
\usepackage{subcaption}

\usepackage{tikz}
\usepackage{pgf}
\usetikzlibrary{shapes,external}
\usetikzlibrary{shapes.multipart}
\usetikzlibrary{decorations.pathreplacing}
\usetikzlibrary{matrix}
\usetikzlibrary{calc}

\usepackage{linguex} 

\usepackage{epigraph}

\usepackage{tabularx}
\usepackage{booktabs}
\newcolumntype{m}{>{\hsize=.65\hsize}X}
\newcolumntype{s}{>{\hsize=.4\hsize}c}
\newcolumntype{u}{>{\hsize=.25\hsize}X}
\newcolumntype{U}{>{\hsize=.15\hsize}X}
\usepackage{multicol}
\usepackage{multirow}

\usepackage{algpseudocode}
\usepackage{algorithm}

\usepackage{setspace}

\newenvironment{abstract}%
    {\cleardoublepage\thispagestyle{empty}\null\vfill\begin{center}%
    \bfseries\abstractname\end{center}}%
    {\vfill\null}
\newcommand\abstractname{Abstract}
\newenvironment{acknowledgements}%
    {\cleardoublepage\thispagestyle{empty}\null\vfill\begin{center}%
    \bfseries Acknowledgements\end{center}}%
    {\vfill\null}

\begin{document}

\begin{titlepage}
	\centering
	\includegraphics[width=0.15\textwidth]{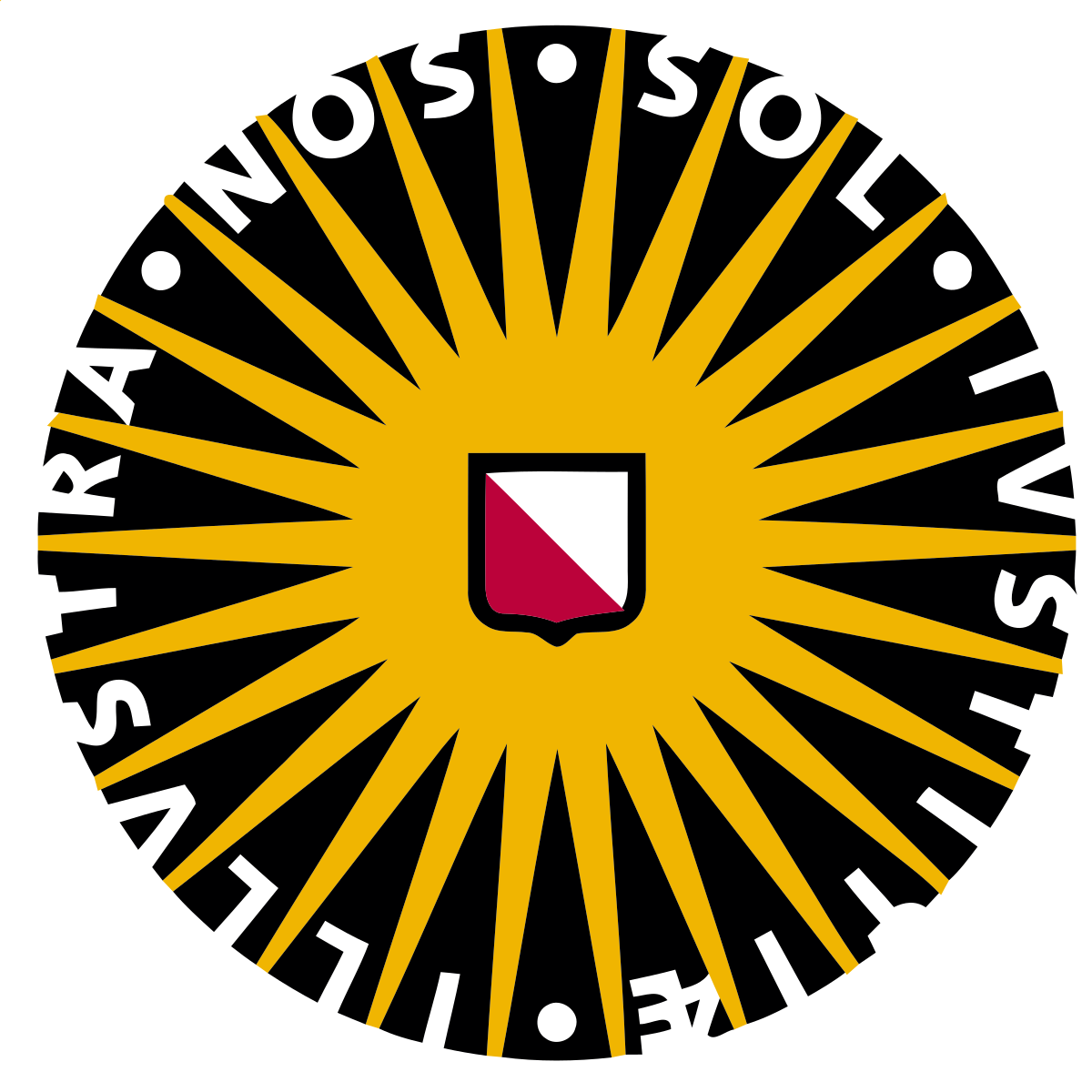}\par\vspace{1cm}
	%{\scshape\LARGE Utrecht University \par}
	%\vspace{1cm}
	%{\scshape\Large Final year project\par}
	\vspace{1.5cm}
	{\huge Extracting \& Learning a Dependency-Enhanced Type Lexicon for Dutch\par}
	\vspace{2cm}
	{\Large Submitted by \\
	\vspace{1.6em}
	 Konstantinos Kogkalidis}
	 \vspace{2cm}
	 
	 {In partial fullfillment of the requirements\\
	 \vspace{1.6em}
	 For the Master's Degree in: Artificial Intelligence  \\
	 \vspace{1.6em}
	 Utrecht University \\
	 \vspace{1.6em}
	 June 2019
	 }
	\vfill
	\begin{flushleft}
	Master's Committee: \\
	\quad\quad	Project Supervisor: Michael Moortgat \\
	\quad\quad	Second Examiner: Tejaswini Deoskar \\
	\quad\quad	External Expert: Richard Moot
	\end{flushleft}

\end{titlepage}

\begin{abstract}
This thesis is concerned with type-logical grammars and their practical applicability as tools of reasoning about sentence syntax and semantics.
The focal point is narrowed to Dutch, a language exhibiting a large degree of word order variability.
In order to overcome difficulties arising as a result of that variability, the thesis explores and expands upon a type grammar based on Multiplicative Intuitionistic Linear Logic, agnostic to word order but enriched with decorations that aim to reduce its proof-theoretic complexity.
An algorithm for the conversion of dependency-annotated sentences into type sequences is then implemented, populating the type logic with concrete, data-driven lexical types.
Two experiments are ran on the resulting grammar instantiation.
The first pertains to the learnability of the type-assignment process by a neural architecture. 
A novel application of a self-attentive sequence transduction model is proposed; contrary to established practices, it constructs types inductively by internalizing the type-formation syntax, thus exhibiting generalizability beyond a pre-specified type vocabulary.
The second revolves around a deductive parsing system that can resolve structural ambiguities by consulting both word and type information; preliminary results suggest both excellent computational efficiency and performance.
\end{abstract}

\begin{acknowledgements}
This thesis would not have been made possible without the ceaseless aid of my support committee;
Richard, who always took the time to listen to my ideas and assist in any problems I would encounter, but also kept a watchful eye for oversights I would otherwise be oblivious to, and
Tejaswini, who kept me motivated despite early failures, and was keen to point out novelty in results I considered trivial.
I could not overstate my luck in being able to work under Michael's supervision.
His expertise, resourcefulness and enthusiasm have inspired and permeated through every aspect of this work.

\vspace{1.6em}
I would like to also express my gratitude towards my family for their unconditional encouragement; my father, the most reliable tech support in all matters imaginable, my mother, who willingly endures my complaining, and my sister, who keeps finding new ways to tease and amuse me.

\vspace{1.6em}
Parts of the thesis have benefitted from fruitful discussions with Vasilis Bountris and Giorgos Tziafas, for which I am thankful.

\vspace{1.6em}
This work has been financially supported by an NWO grant under the scope of the project ``A composition calculus for vector-based semantic modelling with a localization for Dutch'' (360-89-070).
\end{acknowledgements}

\tableofcontents
\listoftables
\listoffigures
\listofalgorithms

\chapter{Introduction}
\label{chapter:intro}
This thesis is concerned with the parsing as deduction paradigm, as orchestrated by type-logical grammars, perceived through the lens of a data-driven experimental setting.
It seeks to bridge the gap between formal theory and empirical practice, integrating insights from half a century of progress in categorial type logics with recent advances in neural networks and natural language processing.

The key theme and the underlying goal behind this work is the development of a concrete, robust and widely applicable methodology for syntactic analysis that naturally lends itself to semantic uses.
The path towards this goal includes many twists and turns, forcing study of semantics to assume a secondary role throughout the thesis.
What is primarily addressed instead are exactly these twists and turns, the resolution of which brings us at an arm's reach of semantics by this thesis' conclusion.

Even though the methods applied are language-agnostic and highly general, the experiments performed focus on Dutch.
Dutch is a particularly challenging language to work with, owing to its many syntactic variations with respect to word order.
As such, it provides an excellent testing ground that puts our designs and hypotheses under rigorous tests, while soliciting novel approaches and creative solutions.

The thesis is organized in four major chapters.
Each chapter is largely autonomous, in the sense that it treats a different topic and seeks to answer a different research question.
However, there is a weak linear dependency between chapters as each one progressively expands upon its predecessors' results; thus, an exhaustive reading should best follow the document in the sequential order it is presented.
Chapters all share a similar structure; they have their own brief introductory overview and a summarizing conclusion which aims to concisely break down its scope and contribution.
A rapid but informative reading could be done on the basis of these summaries.

We begin in Chapter~\ref{chapter:tlg} by providing a brief account of Type-Logical Grammars and their most common varieties.
In considering Dutch, we notice practical issues caused by the language's peculiarities and seek out ways to address them.
The chapter's motif is the balance between formal rigor and pragmatic applicability of type-logical grammars for large-scale use in a language like Dutch.
The chapter concludes with a heedful compromise that retains the best of both worlds.

Having specified the grammar and its latent logic, Chapter~\ref{chapter:extraction} then sets out to populate a data-driven type lexicon.
The dependency-annotated sentences of the written Dutch corpus, Lassy-Small, is used to extrapolate type-logical derivations and phrasal type assignments.
The algorithmic process of this conversion is detailed, alongside the transformations necessitated by incompatibilities between the corpus' annotation philosophy and the grammar's specifications.

Chapter~\ref{chapter:sup} makes for a change of pace; it deals with supertagging, the statistical learning process through which type sequences may be assigned to sentences not included in the source corpus.
In reviewing the extracted type system, it notes a distinction to prior supertagging applications, related to the significantly larger size of the lexicon.
The question then turns to designing a system capable of overcoming this complication, accomplished through a simple reformulation of the problem statement.

Finally, Chapter~\ref{chapter:parsing} seeks to alleviate the proof-theoretic concessions made during the grammar's specification.
The topic revolves around the manipulation of an ambiguous type-logical proof structure utilizing proof-external information sources such as preferential biases exerted by semantic content.

A high-level synopsis plus a few concluding remarks are presented in Chapter~\ref{chapter:conclusion}.

\chapter{Grammar}
\label{chapter:tlg}

\section{Background}
The theoretical framework on top of which this thesis is built are type-logical grammars, a particular family of categorial grammars.
This chapter aims to provide a brief introductory background on type-logical grammars and their historical origins, positioning them within the broader context of categorial grammars and exposing their distinguishing characteristics.
Afterwards, an account of the specific grammar instantiation used for the current work will be given, in terms of its logical and and computational bases, together with the motivating reasons for its choice.
Key references for this chapter are the Stanford Encyclopedia of Philosophy entry for Type-Logical Grammars~\cite{sep-typelogical-grammar} and Moot and Retor{\'e}'s book on Categorial Type Logics~\cite{moot2012logic}.

\subsection{Overview}
\paragraph{Categorial Grammars}
Categorial grammar formalisms have their origins in the works of  Adjukiewicz~\cite{ajdukiewicz1935syntaktische} and Bar-Hillel~\cite{bar1953quasi}.
At their core and ever since their inception, they are defined on the basis of two simple components; a type system and a set of rules dictating type interactions.
The former is an inductive scheme for category (or type) construction, that utilizes a set of atomic types and a set of type-forming operators to provide the means for creating complex types.
The latter provides a number of schemata that describe what kinds of type combinations are permitted, and what the productions of these combinations are.

Categorial Grammars treat syntax as the formal process that dictates how phrases are gradually built by their components, which combine with one-another in terms of function-argument relations. 
They thus epitomize on the \textit{principle of compositionality}, which posits that the meaning of complex expressions is a production of the meaning of their parts and the rules used to compose them.

\paragraph{Parsing as Deduction}
The key insight of Type-Logical Grammars and their distinguishing feature is the logical take on the parsing process.
Lambek was the first to notice that categories may be perceived as logical formulas, and type-forming operators as logical connectives~\cite{lambek1958mathematics}.
Parsing is then lifted from arbitrary schematic rule application to a process of deductive inference, as driven by an underlying logic.
This yields a number of benefits which will be clarified later; for now, it is worth noting the flexibility inherent to such an approach.
Altering the choice of logic gives rise to a different grammar instance, so one may be designed for particular use-cases by a adopting an appropriate logic.
As such, Type-Logical Grammars form a wide landscape which encompasses many formalisms which may differ in their properties but all share a proof-theoretic perspective on parsing.

\subsection{Type-Logical Grammars}
Although a full exposition and comparison between the various incarnations of Type-Logical Grammars escapes the purposes of this work, it is still worthwhile to inspect their persistent aspects and their historical origins.

\paragraph{Lambek Calculus}
We begin with a brief description of what has come to be known as the Lambek Calculus (L)~\cite{lambek1958mathematics}, which built upon AB Grammars~\cite{bar1953quasi} in providing them with a logical formalization.
In the implication-only fragment which is of interest here, categories are defined as follows:
\[
\textsc{c} := \textsc{a} \ | \ \textsc{c}_1/\textsc{c}_2 \ | \ \textsc{c}_1 \backslash \textsc{c}_2
\]
This inductive scheme states that a valid category is either an atomic category $\textsc{a} \in \mathcal{A}$, where $\mathcal{A}$ a closed set of categories, or the result of either of the binary operators $/$, $\backslash$ (read as slash and backslash) on two valid categories.
Intuitively, an atomic element $\textsc{a}$ corresponds to a complete category, whereas a complex category $\textsc{a}/\textsc{b}$ ($\textsc{a} \backslash \textsc{b}$) correspond to an incomplete (or fractional) category that misses a $\textsc{b}$ to the right (left) to produce a full category $\textsc{a}$, with $/$ and $\backslash$ acting as directional \textit{implications}.
The corresponding type-logic contains four logical rules, presented here in Natural Deduction style:
\begin{align*}
    \begin{minipage}{0.5\textwidth}
    \begin{align*}
        \infer{\Gamma, \Delta \vdash B}{
            \Gamma \vdash B / A
            &
            \Delta \vdash A
        }\tag{/E}\\
        \\
        \infer{\Gamma \vdash B/A}{
            \Gamma, A \vdash B
        }\tag{/I}
    \end{align*}
    \end{minipage}
    \begin{minipage}{0.5\textwidth}
    \begin{align*}
        \infer{\Delta, \Gamma \vdash B}{
            \Delta \vdash A
            &
            \Gamma \vdash A\backslash B
        }\tag{$\backslash E$}\\
        \\
        \infer{\Gamma \vdash A\backslash B}{
            A, \Gamma \vdash B
        }\tag{$\backslash I$}
    \end{align*}
    \end{minipage}
\end{align*}
plus the identity Axiom:
\[
\infer{A \vdash A}{}\tag{Ax.}
\]
where $A$, $B$ are formulas (i.e. categories) and $\Gamma$, $\Delta$ are non-empty sequences of formulas. 
A statement of the form $\Gamma \vdash A$ is a judgement, expressing that from a sequence of \textit{assumptions} $\Gamma$ one can derive a \textit{conclusion} formula $A$.

The first line presents the slash and backslash Elimination rules, where $/E$ ($\backslash E$) states that if one has a proof of a formula $B/A$ ($A \backslash B$) from assumptions $\Gamma$ and a proof of a formula $A$ from assumptions $\Delta$, then from assumptions $\Gamma$, $\Delta$ ($\Delta$, $\Gamma$) one can derive a formula $B$.
Note that $\Gamma$, $\Delta$ refers to the concatenation of $\Gamma$ to $\Delta$ and is distinct from $\Delta$, $\Gamma$  --- that is, the order of items within a sequence plays a role in what constitutes a valid proof since our logical rules are non-commutative.
The second line presents the corresponding Introduction rules.
Now, $/I$ ($\backslash I$) state that if ones know a sequence $\Gamma$, $A$ to derive a $B$, then $A$ may be withdrawn, allowing one to derive $B/A$ ($B \backslash A$) from $\Gamma$ alone.
Elimination rules are dual to Introduction rules; the first allow the removal of an implicational type by applying it to its argument, whereas the latter create implicational types by abstracting arguments away, giving our logic access to \textit{hypothetical reasoning}.

To illustrate the linguistic relevance of such a grammar, we will take a second  to review how implication types may be used to convey information on sentence structure.
First off, atomic categories may be seen as structurally complete, independent phrases.
Phrasal composition is coordinated by phrasal heads, which are assigned complex categories.
Heads are then functors which consume the categories of their dependants, producing as a result the wider phrasal category.
At the bottom level, categories are provided by a \textit{lexicon}, a binary relation which associates lexical entries (i.e. words) with one or more potential categories.

To get the point across, we can devise a minimal grammar capable of modeling the syntactic structure of a small set of example sentences.
 Let us first initialize an atomic category set consisting of the elements $\textsc{n}$ for noun, $\textsc{np}$ for noun-phrase and $\textsc{s}$ for sentence, and a corresponding lexicon as follows:
 \[
 \begin{array}{cc}
 \text{Word} & \text{Category} \\
 \hline
 \text{girl}, \ \text{apple} & \textsc{n} \\
 \text{children} & \textsc{np} \\
 \text{the}, \ \text{a(n)} & \textsc{np}/ \textsc{n} \\
 \text{play(s)} & \textsc{np}\backslash \textsc{s} \\
 \text{ate} & (\textsc{np} \backslash \textsc{s}) / \textsc{np} \\ 
 \text{who} & (\textsc{np} \backslash \textsc{np}) / (\textsc{np} \backslash \textsc{s}) \\
 \text{which} & (\textsc{np} \backslash \textsc{np})/(\textsc{s} / \textsc{np})
 \end{array}
 \]

Equipped with the above lexicon, we can use the Lambek Calculus to provide derivations for a number of simple examples as shown in Figure~\ref{fig:lambek_en}, involving usage of our versions of intransitive ``play'' and transitive ``ate'' both in primary and embedded clauses.
Notice how the introduction rules bypass the need for explicit combinatory rules for peripheral extraction, as seen in the object-relativisation example of~\ref{fig:lambek_en:obj}.

 \begin{figure}
     \begin{subfigure}[b]{1\textwidth}
         \centering
         \[
         \infer[\backslash E]{\text{children}, \text{play} \vdash \textsc{s}}{
             \infer[L]{\text{children} \vdash \textsc{np}}{}
             &
             \infer[L]{\text{play} \vdash \textsc{np}\backslash \textsc{s}}{}
         } 
         \]
         \caption{Simple intransitive verb derivation.}
     \end{subfigure}
     \begin{subfigure}[b]{1\textwidth}
         \centering
         \[
         \infer[\backslash E]{\text{the}, \text{girl}, \text{ate}, \text{an}, \text{apple} \vdash \textsc{s}}{
             \infer[/E]{\text{the}, \text{girl} \vdash \textsc{np}}{
                 \infer[L]{\text{the} \vdash \textsc{np} / \textsc{n}}{}
                 &
                 \infer[L]{\text{girl} \vdash \textsc{n}}{}
             }
             &
             \infer[/E]{\text{ate}, \text{an}, \text{apple} \vdash \textsc{np} \backslash \textsc{s}}{
                 \infer[L]{\text{ate} \vdash (\textsc{np} \backslash \textsc{s})/\textsc{np}}{}
                 &
                 \infer[/E]{\text{an}, \text{apple} \vdash \textsc{np}}{
                     \infer[L]{\text{an} \vdash \textsc{np}/ \textsc{n}}{}
                     &
                     \infer[L]{\text{apple} \vdash \textsc{n}}{}
                 }
             }
         }
         \]
         \caption{Simple transitive verb phrase derivation.}
     \end{subfigure}
     \begin{subfigure}[b]{1\textwidth}
     \[
     \infer[/ E]{\text{who}, \text{play} \vdash \textsc{np} \backslash \textsc{np}}{
     	\infer[L]{\text{who} \vdash (\textsc{np}\backslash \textsc{np}) / (\textsc{np} \backslash \textsc{s})}{}
     	&
     	\infer[L]{\text{play} \vdash \textsc{np} \backslash \textsc{s}}{}
     }
     \]
     \caption{Subject-relative intransitive verb derivation.}
     \end{subfigure}
     \begin{subfigure}[b]{1\textwidth}
         \centering
		\[
		\infer[/E]{\text{who}, \text{ate}, \text{an}, \text{apple} \vdash \textsc{np}\backslash \textsc{np}}{
			\infer[L]{\text{who} \vdash (\textsc{np} \backslash \textsc{np}) / (\textsc{np} \backslash \textsc{s})}{}
			&
			\infer[/E]{\text{ate}, \text{an}, \text{apple} \vdash \textsc{np}\backslash \textsc{s}}{
				\infer[L]{\text{ate} \vdash (\textsc{np}\backslash \textsc{s})/\textsc{np}}{}
				&
				\infer[/E]{\text{an}, \text{apple} \vdash \textsc{np}}{\dots}
			}
		}
		\]
         \caption{Subject-relative transitive verb derivation.}
	\end{subfigure}
	\begin{subfigure}[b]{1\textwidth}
         \centering
		\[
		\infer[/E]{\text{which}, \text{the}, \text{girl}, \text{ate} \vdash \textsc{np}\backslash \textsc{np}}{
			\infer[L]{\text{which} \vdash (\textsc{np} \backslash \textsc{np}) / (\textsc{s} / \textsc{np})}{}
			&
			\infer[/I]{\text{the}, \text{girl}, \text{ate} \vdash \textsc{s}/ \textsc{np}}{
				\infer[\backslash E]{\text{the}, \text{girl}, \text{ate}, \textsc{np} \vdash \textsc{s}}{
					\infer[/E]{\text{the}, \text{girl} \vdash \textsc{np}}{\dots}
					&
					\infer[/E]{\text{ate}, \textsc{np} \vdash \textsc{np} \backslash \textsc{s}}{
						\infer[L]{\text{ate} \vdash (\textsc{np} \backslash \textsc{s})/\textsc{np}}{}
						&
						\infer[Ax.]{\textsc{np} \vdash \textsc{np}}{}
					}
				}
			}
		}
		\]
        \caption{Object-relative transitive verb derivation.}
		\label{fig:lambek_en:obj}
     \end{subfigure}
     \caption[English Lambek Derivations]{Simple english sentences and their derivations using the Lambek Calculus. $L$ is used in place of Ax. for categories identified with lexical items.}
     \label{fig:lambek_en}
\end{figure}

\paragraph{Going Stricter}
In our presentation of the Lambek calculus, we have treated assumptions $\Gamma$, $\Delta$ as sequences, i.e. ordered collections, of formulas.
This can, at times, offer the grammar too much creative liberty, resulting in logically correct but linguistically wrong derivations.
This limitation can be bypassed by enhancing the logic with a notion of structure and set of rules to manipulate it, as originally proposed by Lambek~\cite{lambek1961calculus}.
Under this new regime, called the non-associative Lambek Calculus (NL), the logical rules are only applicable under the condition of appropriately bracketed structures:
\begin{align*}
    \begin{minipage}{0.5\textwidth}
    \begin{align*}
        \infer{(\Gamma \circ \Delta) \vdash B}{
            \Gamma \vdash B / A
            &
            \Delta \vdash A
        }\tag{/E}\\
        \\
        \infer{\Gamma \vdash B/A}{
            (\Gamma \circ A) \vdash B
        }\tag{/I}
    \end{align*}
    \end{minipage}
    \begin{minipage}{0.5\textwidth}
    \begin{align*}
        \infer[\backslash E]{(\Delta \circ \Gamma) \vdash B}{
            \Delta \vdash A
            &
            \Gamma \vdash A\backslash B
        }\tag{$\backslash E$}\\
        \\
        \infer[\backslash I]{\Gamma \vdash A\backslash B}{
            (A \circ \Gamma) \vdash B
        }\tag{$\backslash I$}
    \end{align*}
    \end{minipage}
\end{align*}
where $\circ$ is a binary structure-building operator.

If the Lambek Calculus is a language of \textit{strings}, its non-associative version is the language of \textit{binary branching trees}; it  respects (and requires) the constituent structure of a phrase in providing its derivation.

\paragraph{Exerting Control}
Associativity and commutativity (in the form of permutation) may be added to the logic in the form of structural rules that enable them, thus obtaining L and LP (for Lambek Calculus with permutations)~\cite{van1988semantics}:
\begin{align*}
\infer{\Gamma[(\Delta_1 \circ \Delta_2) \circ \Delta_3] \vdash C}{\Gamma[\Delta_1 \circ (\Delta_2 \circ \Delta_3)]\vdash C} \tag{Associativity}\\
\infer{\Gamma[\Delta_1 \circ \Delta_2] \vdash C}{\Gamma[\Delta_2 \circ \Delta_1] \vdash C} \tag{Commutativity}
\end{align*}

Universally allowing associativity yields a grammar that loses track of constituent structure, while universal commutativity corresponds to a grammar that completely ignores word order. 
Disallowing them altogether, on the on other hand, may be too harsh of a measure.
Notice, for instance, that the example on object-relativisation is no longer derivable without associativity, as seen in Figure~\ref{fig:nl_en:nd}.
To benefit from associativity and/or commutativity while restraining its applicability to only cases when it really is needed for a derivation, the logic may be expanded with unary modal operators that either allow or block structural rules~\cite{kurtonina1997structural}.
Structural control modalities improve the logic's proof-theoretic properties -- maximally, the only choices made available during proof search correspond to actual derivational (i.e. non spurious) ambiguity.
That is, if more than one proof may be devised for a single judgement, they correspond to its different readings.

To briefly illustrate the point, we can consider a pair of unary operators $\diamondsuit$, $\Box$ forming a residual pair such that $\diamondsuit \Box A \vdash A \vdash \Box \diamondsuit A$\footnote{N.D. rules for such modalities will be presented in the next section.}, with associativity only applicable on elements marked with $\diamondsuit$:
\[
\infer[A^{\diamondsuit}]{(A\circ B) \circ \diamondsuit C}{A \circ (B \circ \diamondsuit C)}
\] 

Then adding the type $(\textsc{np} \backslash \textsc{np}) / (\textsc{s}/\diamondsuit \Box \textsc{np})$ in our lexicon for the word ``which'' re-enables the derivability of the running example in Figure~\ref{fig:nl_en:d}.

\begin{figure}[t]
	\begin{subfigure}[b]{1\textwidth}
	\centering
	\[
	\infer[/E]{(\text{which} \circ ((\text{the} \circ \text{girl}) \circ \text{ate}))) \vdash \textsc{np}\backslash \textsc{np}}{
		\infer[L]{\text{which} \vdash (\textsc{np} \backslash \textsc{np}) / (\textsc{s} / \textsc{np})}{}
		&
		\infer[/I]{((\text{the} \circ \text{girl}) \circ \text{ate}) \vdash \textsc{s}/ \textsc{np}}{
			\infer[]{(((\text{the} \circ \text{girl}) \circ \text{ate}) \circ \textsc{np}) \vdash \textsc{s}}{
			\lightning
			}
		}
	}
	\]
	\caption[Non-derivable object-relative clause]{Non-derivable object-relative clause in NL.}		
	\label{fig:nl_en:nd}
	\end{subfigure}
	\begin{subfigure}[b]{1\textwidth}
		\centering
	\[
	\infer[/E]{(\text{which} \circ ((\text{the} \circ \text{girl}) \circ \text{ate}))) \vdash \textsc{np}\backslash \textsc{np}}{
		\infer[L]{\text{which} \vdash (\textsc{np} \backslash \textsc{np}) / (\textsc{s} / \diamondsuit \Box \textsc{np})}{}
		&
		\infer[/I]{((\text{the} \circ \text{girl}) \circ \text{ate}) \vdash \textsc{s}/ \diamondsuit \Box \textsc{np}}{
			\infer[A^{\diamondsuit}]{(((\text{the} \circ \text{girl}) \circ \text{ate}) \circ \diamondsuit \Box \textsc{np}) \vdash \textsc{s}}{
			\infer[/E]{((\text{the} \circ \text{girl}) \circ (\text{ate} \circ \diamondsuit \Box \textsc{np}))\vdash \textsc{s}}{
				\infer[\backslash E]{\text{the}, \text{girl}}{\dots}
				&
				\infer[\backslash E]{(\text{ate} \circ \diamondsuit \Box \textsc{np}) \vdash \textsc{np} \backslash \textsc{s}}{
					\infer[L]{\text{ate} \vdash (\textsc{np}\backslash \textsc{s})/\textsc{np}}{}
					&
					\infer[]{\diamondsuit \Box \textsc{np} \vdash \textsc{np}}{}
				}
			}
			}
		}
	}
	\]
	\caption{..now derivable using a residual pair of control operators.}
	\label{fig:nl_en:d}
	\end{subfigure}
	\caption[Structural Control Example]{Example of a structural control modality as a licensing feature for limited associativity.}
\end{figure}

\paragraph{Lexical Ambiguity}
These proof-theoretic advantages do not come for free, however.
Even though a stricter type system mitigates the difficulty of proof search, the burden is not removed but rather shifted onto the lexicon.
As words are assigned more potential types, the lexicon becomes increasingly ambiguous.
This ambiguity is further exacerbated for languages exhibiting higher degrees of word order freedom, with Dutch being a prime example.

When concerned with relativisation, for instance, which in Dutch is verb-final, we need to either include the second type $\textsc{np}\backslash(\textsc{np}\backslash\textsc{s})$ for transitive verbs, or modal decorations for conditional commutativity and associativity that would allow a hypothetical noun-phrase to move unhindered through the proof until it finds its correct position.
Other permutations are yet possible in the context of inverted clauses, or wh- and yes/no-questions.
The recurrence and variety of these complications can vastly increase the complexity of the type system, if we wish proof search to remain deterministic.
This complexity can become unwieldy when dealing with large-scale corpora, owing mostly to the potentially immense size (and therefore ambiguity) of the lexicon, but also the considerable difficulty of populating such a lexicon in the first place.
Consequently, there is a balance to be sought between the formal well-behavedness of the grammar and its practical applicability.

\section{A TLG for Semantic Compositionality}
\subsection{Intuitionistic Linear Logic}
The above considerations, combined with our goal of constructing a wide-coverage grammar for Dutch, draws our attention towards LP, otherwise known as the Lambek-van Benthem Calculus~\cite{van1988semantics}.
LP coincides with the implication-only fragment of Multiplicative Intuitionistic Linear Logic (MILL)~\cite{girard1987linear} the notation of which is adopted for the purposes of this presentation.
MILL is closely related to the Lambek Calculus shown earlier; the two logical connectives of $/$ and $\backslash$ are collapsed into a single, direction-agnostic implication $\rightarrow$ (an alternative notation is $\multimap$).
Similar approaches to syntactic analysis were foreboded by early proponents, as in~\cite{curry1961some} and their contemporary incarnations of Lambda Grammars~\cite{muskens2001lambda} and Abstract Categorial Grammars ~\cite{de2001towards}; the relation between these and the approach presented here will be briefly examined in Chapter~\ref{chapter:parsing}. 

MILL implication types are inductively defined as:
\[
\textsc{t} := \textsc{a} \ | \ \textsc{t}_1 \to \textsc{t}_2 
\]
where again $\textsc{a}$ is an atomic type and $\textsc{t}_1$, $\textsc{t}_2$ are types.

The implication rules and identity axiom of MILL are presented below:

\begin{center}
\begin{minipage}{0.5\textwidth}
\[
\infer{A \vdash A}{}\tag{Ax.}
\]
\end{minipage}\\
\end{center}
\begin{align*}
    \begin{minipage}{0.5\textwidth}
	\[
        \infer{\Gamma, \Delta \vdash B}{
            \Gamma \vdash A \rightarrow B
            &
            \Delta \vdash A
        }\tag{$\rightarrow$ E}
    \]
    \end{minipage}
    \begin{minipage}{0.5\textwidth}
    \[
        \infer{\Gamma \vdash A \rightarrow B}{
            \Gamma, A \vdash B
        }\tag{$\rightarrow I$}\\
    \]
    \end{minipage}
\end{align*}

Note that rather than the sequences of L or binary branching structures of NL, the assumptions of a judgement in MILL are now \textit{multisets}; $\Gamma$, $\Delta$ is then read as the multiset union of $\Gamma$ with $\Delta$ and is equivalent to $\Delta$, $\Gamma$.
In practical terms, this means that both associativity and commutativity are admitted as holding universally; neither structure nor word order are taken into account when deriving a sentence\footnote{For non-native Dutch speakers, this may at times feel as less of a concession and more of a postulate.}.
In other words, a single type may be used to present multiple versions of functors that would previously differ depending on the variations over their arguments' positions.

\paragraph{The Curry-Howard Correspondence}
The Curry-Howard Correspondence states that logical propositions are in a one-to-one relation with the types of a functional program.
MILL, in particular, is directly equivalent to the simply-typed linear $\lambda$-calculus~\cite{benton1993term, abramsky1993computational}.
A proof then encodes a $\lambda$-term which fully specifies the execution of a functional program, and vice-versa.
Logical connectives are identified with type constructors; specifically, implicational formulas are the type signatures of function spaces. 
Assumptions are free variables, and the rules of introduction and elimination find their computational analogues in function abstraction and function application, respectively, whereas the identity axiom corresponds to variable instantiation.

The ramifications of this insight are far-reaching and their analysis falls beyond the scope of this thesis; we rather want to focus on one particular aspect of the correspondence, namely its significance for semantic compositionality.

Let's begin by inspecting the logical rules decorated with their corresponding $\lambda$-terms, and giving them an intuitive reading.

\begin{center}
\begin{minipage}{0.5\textwidth}
\[
\infer{x: A \vdash x: A}{}\tag{Ax.}
\]
\end{minipage}\\
\end{center}
\begin{align*}
    \begin{minipage}{0.5\textwidth}
	\[
        \infer{\Gamma, \Delta \vdash s(t): B}{
            \Gamma \vdash s: A \rightarrow B
            &
            \Delta \vdash t: A
        }\tag{$\rightarrow$ E}
    \]
    \end{minipage}
    \begin{minipage}{0.5\textwidth}
    \[
        \infer{\Gamma \vdash \lambda x. u : A \rightarrow B}{
            \Gamma, x: A \vdash u: B
        }\tag{$\rightarrow I$}\\
    \]
    \end{minipage}
\end{align*}

The identity axiom simply states that we may instantiate a free variable $x$ of type $A$.
The elimination rule $\rightarrow E$ states that if we have a program that from a typing environment $\Gamma$ (i.e. a set of type-specified variables) can produce a function $s$ of type $A \to B$, and a program that from another set $\Delta$ can produce a variable $t$ of type $A$, then $s$ may be applied to $t$ yielding a variable of type $B$.
Dually, the introduction rule states that from a program that produces a variable $u$ of type $B$ out of a set of variables $\Gamma$ together with a variable $x$ of type $A$, we can construct a program for a function of type $A\to B$ by abstracting $x$ away.

Recalling that our variables are instantiated by a type lexicon, we can easily shift from syntax to semantics via a homomorphic mapping.
Concretely, for each lexical type assignment we need also provide a corresponding semantic assignment.
Then, the process of meaning assembly for a phrase is identical with the execution of the functional program dictated by its syntactic derivation; in other words, we may use the semantic values of our lexicon, applying function-words to their arguments in a hierarchical manner, guided by the $\lambda$-term that encodes the proof structure.

As semantic compositionality is one of the planned applications of our grammar, we will stress this point by providing an abstract example.
First, let's return to the prior examples, now using an MILL-adapted lexicon\footnote{Throughout the remainder of this thesis, we will use a right-implicit parentheses notation for MILL types; that is, $A\to B \to C$ is read as $A\to (B \to C)$ and is distinct from $(A\to B) \to C$}:

 \[
 \begin{array}{cc}
 \text{Word} & \text{Category} \\
 \hline
 \text{meisje}, \ \text{appel} & \textsc{n} \\
 \text{het}, \ \text{een}  & \textsc{n} \to \textsc{np}\\
 \text{at}  & \textsc{np} \to \textsc{np} \to \textsc{s}\\ 
 \text{dat}, \ \text{die} & (\textsc{np}\to \textsc{s})\to \textsc{np} \to \textsc{np} \\
 \end{array}
 \]

Figure~\ref{fig:ill_dutch} presents derivations for a transitive verb phrase in primary and embedded clauses.
The corresponding $\lambda$-terms are obtained by following the proof constructions top-down.
Note that the embedded clause example that gave us trouble earlier is now trivial to derive, while maintaining the subject-relative reading.

\begin{figure}
	\begin{subfigure}[b]{1\textwidth}
	\centering
	\small
		\[
		\infer[\rightarrow E]{\text{het}, \text{meisje}, \text{at}, \text{een}, \text{appel} \vdash \textsc{s}}{
			\infer[\rightarrow E]{\text{at}, \text{een}, \text{appel} \vdash \textsc{np} \to \textsc{s}}{
				\infer[L]{\text{at} \vdash \textsc{np} \to \textsc{np} \to \textsc{s}}{}
				&
				\infer[\rightarrow E]{\text{een}, \text{appel} \vdash \textsc{np}}{
					\infer[L]{\text{een} \vdash \textsc{n} \to \textsc{np}}{}
					&
					\infer[L]{\text{appel} \vdash \textsc{n}}{}
				}
			}
			&		
			\infer[\rightarrow E]{\text{het}, \text{meisje} \vdash \textsc{np}}{
				\infer[L]{\text{het} \vdash \textsc{n} \to \textsc{np}}{}
				&
				\infer[L]{\text{meisje} \vdash \textsc{n}}{}
			}
		}
		\]
		\caption{Simple transitive verb derivation, with $\lambda$-term (at(een appel))(het meisje)}
		\end{subfigure}
		\begin{subfigure}[b]{1\textwidth}
		\centering
		\scriptsize
		\[
		\infer[\rightarrow E]{\text{dat}, \text{een}, \text{appel}, \text{at} \vdash \textsc{np} \to \textsc{np}}{
			\infer[L]{\text{dat} \vdash (\textsc{np} \to \textsc{s}) \to \textsc{np} \to \textsc{np}}{}
			&
			\infer[\rightarrow E]{\text{een}, \text{appel}, \text{at} \vdash \textsc{np} \to \textsc{s}}{
				\infer[L]{\text{at} \vdash \textsc{np} \to \textsc{np} \to \textsc{s}}{}
				&
				\infer[\rightarrow E]{\text{een}, \text{appel} \vdash \textsc{np}}{
					\infer[L]{\text{een} \vdash \textsc{n} \to \textsc{np}}{}
					&
					\infer[L]{\text{appel} \vdash \textsc{n}}{}
				}
			}
		}
		\]
		\caption{Subject-relative transitive verb derivation, with $\lambda$-term dat(at (een appel))}
		\label{subfig:ill_dutch:sub}
		\end{subfigure}
		\begin{subfigure}[b]{1\textwidth}
		\centering
		\scriptsize
		\[
		\infer[\rightarrow E]{\text{die}, \text{het}, \text{meisje}, \text{at} \vdash \textsc{np} \to \textsc{np}}{
			\infer[L]{\text{die} \vdash (\textsc{np} \to \textsc{s})\to \textsc{np} \to \textsc{np}}{}
			&
			\infer[\rightarrow I]{\text{het}, \text{meisje}, \text{at} \vdash \textsc{np} \to \textsc{s}}{
				\infer[\rightarrow E]{\text{het}, \text{meisje}, \text{at}, \textsc{np} \vdash \textsc{s}}{
					\infer[\rightarrow E]{\text{at}, \textsc{np}\vdash \textsc{np} \to \textsc{s}}{
						\infer[L]{\text{at} \vdash \textsc{np} \to \textsc{np} \to \textsc{s}}{}
						&
						\infer[Ax.]{\textsc{np} \to \textsc{np}}{}
					}
					&
					\infer[\rightarrow E]{\text{het}, \text{meisje} \vdash \textsc{np}}{
						\infer[L]{\text{het} \vdash \textsc{n} \to \textsc{np}}{}
						&
						\infer[L]{\text{meisje} \vdash \textsc{n}}{}					
					}
				}
			}
		}
		\]
		\caption{Object-relative transitive verb derivation, with $\lambda$-term die($\lambda x$.((at $x$)(het meisje)))}
		\label{subfig:ill_dutch:obj}
		\end{subfigure}
\caption[Example MILL Derivations]{Example derivations in MILL}
\label{fig:ill_dutch}
\end{figure}

Now, given a mapping $\lceil .\rceil$ from words to semantic objects, we may obtain a compositionally driven semantic interpretation over larger linguistic units by recursively applying the mapping on the proofs' $\lambda$-terms, e.g.:
\[
\lceil \text{die het meisje at} \rceil = \lceil \text{die} \rceil (\lambda x. ((\lceil \text{at} \rceil \ x)(\lceil \text{het} \rceil \lceil \text{meisje} \rceil))
\]

The exact semantic spaces operated on are still open to our creative libery; specifying those escapes the context of this thesis, but the grammar's ability to accommodate a multitude of such spaces is still an important point to consider.

\subsection{Dependency Refinement}
Earlier, we saw how MILL simplifies the derivation process for cases that would otherwise require involved structural reasoning.
The observant reader will, however, have noticed that this laxity can result in erroneous analyses.
Returning, for instance, to the relativisation examples of Figure~\ref{fig:ill_dutch}, there is no restriction enforcing us to derive the particular proofs presented.
The proof structure of~\ref{subfig:ill_dutch:sub} could be used in~\ref{subfig:ill_dutch:obj} (and vice-versa), resulting in linguistically inaccurate readings.
Even though associativity and commutativity mitigate the problem of resolving long-distance or crossing dependencies, they vastly increase the proof-search space and permit derivations that are completely off-point.

This is a compromise consciously made; even though the types will not suffice for deductive parsing, they may be used as an auxiliary information source on top of the words themselves.
The assumption made here is that the combination of lexical-level preferences and type-level information will prove adequate in the selection of the most plausible reading and its proof; for example, when concerned with the sentence ``het meisje at een appel'' (\textit{the girl ate an apple}), we know with a degree of certainty that apples are much more plausible as objects of eating compared to girls.

The crucial insight is that even though the position of phrasal dependants (and thus functor arguments) may be variable, their syntactic role remains constant.
These roles are left implicit for non-permuting calculi (for example the inner-most argument of the $\textsc{np}\backslash(\textsc{s}/\textsc{np})$ refers to the object of the transitive verb), but can be explicitly defined in the current setting.
To imlement this refinement, we subclass the implication arrow into several named variants, each (roughly) corresponding to a particular dependency label.
Functor types now also specify the syntactic slot occupied by their arguments, denoted by the name decoration of their corresponding implication.
A transitive verb then gets typed as $\textsc{np}\myrightarrow{su}\textsc{np}\myrightarrow{obj}\textsc{s}$; a curried function that consumes first a noun-phrase in subject position and then a noun-phrase in object position before producing a sentence.

Such an enrichment of the type system has multifold benefits; functor types gain a more intuitive reading, which also greatly increases their informational content.
They no longer encode just the local phrase structure, but also the dependencies enacted by the structure; functors that share the same arguments but have them occupy different syntactic slots are now distinguishable from one another.
Further, lexical preferences can now be canonically shared across types rather than being tied to argument positions over individual types.
Last but not least, novel usecases for dependency decorations are likely to arise in the context of semantic interpretations.

Of course, this addition is expected to significantly increase the lexicon size.
However, differently decorated types arising from a single MILL type will really be functionally distinct (both syntactically and semantically), as opposed to simply an artifact of permutation (as would be the case in a directional system).
Consider, for instance, the case of the relativiser ``die'', which would get the two different types\footnote{The dependency label body refers to relative clause body.} $(\textsc{np}\myrightarrow{su}\textsc{s})\myrightarrow{body}\textsc{np}\myrightarrow{mod}\textsc{np}$ and $(\textsc{np}\myrightarrow{obj}\textsc{s})\myrightarrow{body}\textsc{np}\myrightarrow{mod}\textsc{np}$, the first for subject- and the second for object-relativisation.
The size increase is anyway modest compared to what a directional grammar would yield; therefore, the decorations are cost-efficient, in the sense that they offer the most advantages for the least added complexity and lexical ambiguity.

It is noteworthy to point out that not all implication instances need be decorated; higher-order types might involve hypothetical reasoning over objects which do not project dependency information.

\paragraph{A Formal Treatment}
Throughout the rest of the thesis, we will be using the decsribed dependency decorations in a informal manner, as a meta-logical notation that is simply used to convey useful auxiliary information.
However, such decorations can be properly included in the logic in the form of modal operators.
First, consider that the arrow decoration is a characterization of the means of argument consumption; as such, it can be shifted onto the argument itself.
With this in mind, we can insert a unary logical connective $\diamondsuit^d$ for each dependency label $d$, together with its corresponding structural counterpart $\langle \rangle ^d$.

The logic then requires two extra rules for $\diamondsuit^d$ Introduction and Elimination:

\begin{align*}
    \begin{minipage}{0.5\textwidth}
	\[
        \infer{\langle \Gamma \rangle^d \vdash \diamondsuit^d A}{\Gamma \vdash A}\tag{$\diamondsuit^d I$}
    \]
    \end{minipage}
    \begin{minipage}{0.5\textwidth}
    \[
        \infer{\Gamma [\Delta] \vdash B}{
        \Delta \vdash \diamondsuit^d A
        &
        \Gamma[\langle A \rangle^d] \vdash B
        }\tag{$\diamondsuit^d E$}
    \]
    \end{minipage}
\end{align*}

The first states that a proof of $A$ from assumptions $\Gamma$ can be converted into a proof a $d$-decorated $A$ from a $d$-bracketed $\Gamma$.
The second says that given a proof for $\diamondsuit^d A$ from structure $\Delta$ and a proof of $B$ from a structure $\Delta$ that contains a $d$-bracketed $A$ as a sub-structure, the latter may be replaced by a $\Gamma$ and still yield a $B$.
Figure~\ref{fig:ill_dutch_modal} presents another relativisation example, showcasing the effect of dependency decorations in the proof.
Note that the modalities do not alter proof-search when performed in a forward manner, but require the dependency bracketing when done in a backward manner.
A two-step backward-forward strategy may be used to first construct a standard MILL proof, then insert the modal decorations.

\begin{figure}
	\begin{sideways}
	\begin{subfigure}[lt]{1\textheight}
	\centering
	\scriptsize
	\[
	\infer{\langle \text{eieren} \rangle^\text{mod}, \text{die}, \langle \langle \text{kippen} \rangle^\text{obj}, \text{leggen} \rangle^\text{body} \vdash \textsc{np}}{
		\infer[\rightarrow E]{\text{die}, \langle \langle \text{kippen} \rangle^\text{obj}, \text{leggen}\rangle^\text{body} \vdash \diamondsuit^\text{mod}\textsc{np} \rightarrow \textsc{np}}{
			\infer[L]{\text{die} \vdash \diamondsuit^\text{body}(\diamondsuit^\text{su}\textsc{np} \to \textsc{s})\to(\diamondsuit^\text{mod} \textsc{np} \to \textsc{np})}{}
			&
			\hspace{-20pt}
			\infer[\diamondsuit^\text{body}I]{\langle \langle \text{kippen} \rangle^\text{obj}, \text{leggen} \rangle^\text{body} \vdash \diamondsuit^\text{body}(\diamondsuit^\text{su}\textsc{np} \to \textsc{s})}{
				\infer[\rightarrow I]{\langle \text{kippen} \rangle^\text{obj}, \text{leggen}  \vdash \diamondsuit^\text{su} \textsc{np}\to \textsc{s}}{
					\infer[\diamondsuit^\text{su} E]{\langle \text{kippen} \rangle^\text{obj}, \text{leggen}, \diamondsuit^\text{su} \textsc{np} \vdash \textsc{s}}{
						\infer[Ax.]{\diamondsuit^\text{su} \textsc{np} \vdash \diamondsuit^\text{su} \textsc{np}}{}
						&
						\infer[\rightarrow E]{\langle \text{kippen} \rangle^\text{obj}, \text{leggen}, \langle \textsc{np} \rangle^\text{su} \vdash \textsc{s}}{
							\infer[\rightarrow E]{\text{leggen}, \diamondsuit^\text{su} \vdash \diamondsuit^\text{obj} \textsc{np} \to \textsc{s}}{
								\infer[L]{\text{leggen} \vdash \diamondsuit^\text{su}\textsc{np}\to\diamondsuit^\text{obj} \textsc{np} \to \textsc{s}}{}
								&
								\infer[Ax.]{\diamondsuit^\text{su}\textsc{np} \vdash \diamondsuit^\text{su}\textsc{np}}{}
							}
							&
							\infer[\diamondsuit^\text{obj} I]{\langle \text{kippen} \rangle^\text{obj}\vdash \diamondsuit^\text{obj}\textsc{np}}{
							\infer[L]{\text{kippen} \vdash \textsc{np}}{}
							}
						}
					}
				}
			}
		}
		&
		\hspace{-90pt}
		\infer[\diamond^\text{mod}I]{\langle \text{eieren} \rangle^\text{mod} \vdash \diamondsuit^\text{mod} \textsc{np}}{
			\infer[L]{\text{eieren} \vdash \textsc{np}}{}
		}
	}
	\]
	\caption{Subject-relative derivation.}
	\label{subfig:modal_su}
	\end{subfigure}
	\end{sideways}
		\begin{sideways}
	\begin{subfigure}[lb]{1\textheight}
	\centering
	\scriptsize
	\[
	\infer{\langle \text{eieren} \rangle^\text{mod}, \text{die}, \langle \langle \text{kippen} \rangle^\text{su}, \text{leggen} \rangle^\text{body} \vdash \textsc{np}}{
		\infer[\rightarrow E]{\text{die}, \langle \langle \text{kippen} \rangle^\text{su}, \text{leggen}\rangle^\text{body} \vdash \diamondsuit^\text{mod}\textsc{np} \rightarrow \textsc{np}}{
			\infer[L]{\text{die} \vdash \diamondsuit^\text{body}(\diamondsuit^\text{obj}\textsc{np} \to \textsc{s})\to(\diamondsuit^\text{mod} \textsc{np} \to \textsc{np})}{}
			&
			\infer[\diamondsuit^\text{body}I]{\langle \langle \text{kippen} \rangle^\text{su}, \text{leggen} \rangle^\text{body} \vdash \diamondsuit^\text{body}(\diamondsuit^\text{obj}\textsc{np} \to \textsc{s})}{
				\infer[\rightarrow E]{\langle \text{kippen} \rangle^\text{su}, \text{leggen} \vdash \diamondsuit^\text{obj}\textsc{np} \to \textsc{s}}{
					\infer[L]{\text{leggen} \vdash \diamondsuit^\text{su} \textsc{np} \to \diamondsuit^\text{obj} \textsc{np} \to \textsc{s}}{}
					&
					\infer[\diamondsuit^\text{su}I]{\langle \text{kippen} \rangle^\text{su} \vdash \diamondsuit^\text{su} \textsc{np}}{
						\infer[L]{\text{kippen} \vdash \textsc{np}}{}
					}
				}
			}
		}
		&
		\infer[\diamond^\text{mod}I]{\langle \text{eieren} \rangle^\text{mod} \vdash \diamondsuit^\text{mod} \textsc{np}}{
			\infer[L]{\text{eieren} \vdash \textsc{np}}{}
		}
	}
	\]
	\caption{Object-relative derivation.}
	\label{subfig:modal_obj}
	\end{subfigure}
	\end{sideways}
\caption[Example MILL Derivations with Dependency Modalities]{Example derivations in MILL, using dependency modalities, for the sentence ``eieren die kippen leggen''. Subfigure~\ref{subfig:modal_su} presents the subject-relative reading (\textit{eggs that lay chickens}), while subfigure~\ref{subfig:modal_obj} presents the object-relative reading (\textit{eggs that chickens lay}). The combination of semantic preferences and dependency annotations should select the second derivation as the most plausible one.}
\label{fig:ill_dutch_modal}
\end{figure}

\section{Summary}
This chapter examined Type-Logical Grammars and some of the major steps throughout their evolution.
We saw how parsing may be understood as a formal deduction process, and how stricter logics can regulate this process so as to keep it linguistically grammatical.
Using a few examples, we recognized that stricter logics come at the cost of increased lexical ambiguity, further emphasized for a language with large word order variety like Dutch. 
With this in mind, we considered a laxer grammar based on Intuitionistic Linear Logic.

A MILL-based grammar boasts simplicity, ease of dealing with discontinuous or long-range dependencies and a clear and direct correspondence with the simply typed linear $\lambda$-calculus, making it an ideal driving force for semantic compositionality.
On the other hand, the axiomatic treatment of associativity and commutativity permits more proofs than desired and increases parsing complexity.
To address this issue, we enriched the grammar by subclassing the linear implication to a set of named versions, each one suggesting a unique ``means of consumption'' as specified by a corresponding syntactic dependency.
The novel contribution of this chapter lies in this enrichment, which can facilitate a preferential lexical bias, thus aiding in the reduction of the search space ambiguity.

\chapter{Extraction}
\label{chapter:extraction}
%todo: more examples
%todo: modifier copying
%todo: multi-arg cnj

\section{Background}
Categorial grammars rely on lexical categories, in our case types, to provide meaningful syntactic derivations for phrases and sentences.
They thus differ from standard dependency or phrase-structure formalisms in the sense that most of the information necessary for parsing is located within the words and their types rather than the syntactic structure built on top of them.
This makes it evident that the most crucial component of a categorial grammar treebank are the type assignments themselves.
Constructing such a treebank from scratch is impractical; the annotation process is a costly and time-demanding endeavour. 
Instead, it is easier to utilize pre-annotated corpora, converting them into a categorial format.
The conversion could be performed manually, in the case of smaller corpora, but a degree of automation becomes imperative for larger ones, which are a prerequisite for building statistical models.
The process of automatically transforming a syntactically annotated corpus into a categorial grammar treebank is hereon referred to as grammar extraction.
It involves the design and application of an algorithm which manipulates the sentential syntactic structure and projects it into categories for each of its parts, as well as a number of preprocessing steps.
Depending on the grammar and its type system, these categories and their rules of interaction encode the collapsed structure, allowing its full or partial reconstruction.

No universal algorithm exists for grammar extraction; the process may vary significantly depending on factors such as the source corpus to be converted, the syntactic formalism it abides by, the language and its particularities, and the target grammar and its specification.
Despite their distinguishing points, all works on grammar extraction share a common core in the underlying algorithmic process.
Phrases are analyzed as binary branching structures, with one branch corresponding to a word that acts as the phrasal head and a branch that corresponds to a phrase or word that plays the role of the phrasal dependent. 
The distinction between the two is made on the basis of the dependencies and/or syntactic tags provided by the original annotation.
The head is then treated as a functor from the category of the dependant to the category of their common root, where these are provided by means of some translation from the annotation's tags and dependencies to a closed set of atomic formulae.

Significant work has been carried out for a wide range of corpora of different languages in the context of various categorial frameworks. 
Pioneering research provided the groundwork for converting constituency-based corpora to CCG~\cite{steedman2011combinatory} derivations, in languages like English~\cite{hockenmaier2007ccgbank} and German~\cite{hockenmaier2006creating}, with later expansions for languages like Hindi~\cite{ambati2018hindi} and Japanese~\cite{uematsu2015integrating} adapted to dependency-based corpora.

The main point of reference for this chapter is Moortgat and Moot's extraction algorithm~\cite{moortgat2002using} (for an updated version, refer to~\cite{moot2010extraction}), proposed for the extraction of type-logical supertags from the spoken dutch corpus.
It is in many ways similar to the current endeavor, owing to the affinities between the source corpora and the common language.
Albeit being the work most related to ours, the proposed algorithm still had to be adapted in many ways.
Although the annotations of CGN~\cite{hoekstra2001syntactic}, the spoken dutch corpus, are largely the same as Lassy, the corpus employed in the current work (see Section~\ref{section:corpus}), a few key differences occur between the two.
As many of these are incompatible with the originally proposed algorithm, considerable extra effort is necessitated in terms of preprocessing steps.
The algorithm itself also differs, as our target type-logical grammar is largely divergent from established practices, as described in Chapter~\ref{chapter:tlg}.

The rest of this chapter is aimed towards providing a functional overview of the extraction process, including abstract algorithmic descriptions, linguistic justifications for the design decisions made and motivating examples.
As the extraction is closely tied to the input corpus and its annotations, we will begin by providing a brief overview of it first, using it as a constant point of reference afterwards.

\section{Corpus}
\label{section:corpus}
Our core corpus for this endeavor is Lassy-Small~\cite{Lassy}.
Lassy consists of 65\,200 sentences of written Dutch, originating from various sources such as newspaper articles, wikipedia crawls, books and magazines, etc.
In total, it contains approximately 1.1 million words and 78\,570 unique tokens.

The sentences have been automatically parsed by the Alpino parser~\cite{bouma2001alpino}, producing annotations in the form of directed acyclic graphs (DAGs)\footnote{A detailed overview of Lassy's annotations is provided by the corresponding manual at~\url{https://www.let.rug.nl/vannoord/Lassy/sa-man_lassy.pdf}.}.
The graphs' nodes represent complete words and phrases, the former in the case of leaves (terminal nodes) and the latter otherwise, associated with a part-of-speech tag or phrasal category respectively.
Nodes are connected with one-another through labeled, directed edges, connecting phrasal nodes to their constituents, with the labels denoting the dependency relation between two items. 
Reentrancy is used to model phenomena such as embedded clauses, ellipses and abstract semantic relations.
It is implemented via the insertion of ``phantom'' nodes, i.e. nodes which are not lexically grounded, but share a mutual index with their material counterparts allowing them to be identified.

The tree structures, dependency labels, and part-of-speech and phrasal tags within Lassy-Small have been manually verified and corrected.
The resulting corpus boasts both high quality and adequate size, making it an ideal test bed for our experimentation.

Aside from Lassy-Small, an additional resource that is available is Lassy-Large.
Lassy-Large is a silver-standard corpus of considerable size (almost 700 times that of Lassy-Small), also automatically annotated by the Alpino parser but not manually verified.
As the annotation format and conventions followed by Lassy-Small is in near perfect agreement with that of Lassy-Large, the extraction process is also applicable on the latter.
Even though preliminary tests have been carried out successfully, we refrain from utilizing it in the current work, as its noisy nature makes it a lower-grade resource for grammar induction.

\subsection{Corpus Statistics}
\paragraph{Tag Sets}
Tables~\ref{table:lex} and~
\ref{table:colors} depict the sets of part-of-speech tags and phrasal categories and dependency relations, respectively, as used in the corpus.
A short description is provided for each item, as well as its relative frequency within its domain.
Auxiliary items serving meta-annotation purposes have been excluded.

\begin{table}
\begin{center}
\begin{tabularx}{1\linewidth}{usss}
      \textbf{Tag} & \textbf{Description} & \textbf{Frequency} (\%) & \textbf{Assigned Type}\\
      \toprule
      \multicolumn{3}{c}{Short POS Tags}\\
      \midrule[0.005pt]
            
      \textit{adj} & Adjective & 7.3 & \textsc{adj}\\
      \textit{bw} & Adverb & 4.5 & \textsc{bw}\\
      \textit{let} & Punctuation & 11.2 & \textsc{let}\\
      \textit{lid} & Article & 10.7 & \textsc{lid}\\
      \textit{n} & Noun & 22.5 & \textsc{n}\\
      \textit{spec} & Special Token & 3.5 & \textsc{spec}\\
      \textit{tsw} & Interjection & $<$0.1 & \textsc{tsw}\\
      \textit{tw} & Numeral & 2.4 & \textsc{tw}\\
      \textit{vg} & Conjunction & 4.2 & \textsc{vg}\\
      \textit{vnw} & Pronoun & 6.5 & \textsc{vnw}\\
      \textit{vz} & Preposition & 13.7 & \textsc{vz}\\
      \textit{ww} & Verb & 13.2 & \textsc{ww}\\
      \midrule[0.005pt]
      \multicolumn{3}{c}{Phrasal Category Tags}\\
      \midrule[0.005pt]
      \textit{advp} & Adverbial Phrase & 0.6 & \textsc{adv}\\
      \textit{ahi} & Aan-Het Infinitive & $<$0.1 & \textsc{ahi}\\
      \textit{ap} & Adjectival Phrase & 2.1 & \textsc{ap}\\
      \textit{cp} & Complementizer Phrase & 3.3 & \textsc{cp}\\
      \textit{detp} & Determiner Phrase & 0.2 & \textsc{detp}\\
      \textit{inf} & Bare Infinitival Phrase & 4.7 & \textsc{inf}\\
      \textit{np} & Noun Phrase & 36.7 & \textsc{np}\\
      \textit{oti} & Om-Te Infinitive & 0.8 & \textsc{oti}\\
      \textit{pp} & Prepositional Phrase & 23.2 & \textsc{pp}\\
      \textit{ppart} & Past Participial Phrase & 4.2 & \textsc{ppart}\\
      \text{ppres} & Present Participial Phrase & 0.1 & \textsc{ppres}\\
      \textit{rel} & Relative Clause & 1.9 & \textsc{rel}\\
      \textit{smain} & SVO Clause & 4.7 & \textsc{smain}\\
      \textit{ssub} & SOV Clause & 0.8 & \textsc{ssub}\\
      \textit{sv1} & VSO Clause & $<$0.1& \textsc{sv1}\\
      \textit{svan} & Van Clause & $<$0.1 &\textsc{svan}\\
      \textit{ti} & Te Infinitive & 1.8 &\textsc{ti}\\
      \textit{whq} & Main WH-Q & 0.1 & \textsc{whq}\\
      \textit{whrel} & Free Relative & 0.2 &\textsc{whrel}\\
      \textit{whsub} & Subordinate WH-Q & 0.2 & \textsc{whsub}\\
      \textit{du} & Discourse Unit & 2.6 & N/A*\\
      \textit{mwu} & Multi-Word Unit & 5.9 & N/A*\\
      \textit{conj} & Conjunct & 5.7 & N/A*\\
      \midrule[0.005pt]
\end{tabularx}
\end{center}
\caption[Extracted Atomic Types]{Part-of-speech tags and phrasal categories, and their corresponding type translations.\\ (*): Not used as a type.}
\label{table:lex}
\end{table}

\begin{table}
\begin{tabularx}{1\linewidth}{ussU}
      \textbf{Dep} & \textbf{Description} & \textbf{Frequency} (\%) & \textbf{Implication Label}\\
      \toprule
      \textit{app} & Apposition & 0.8 & \textit{app}\\
      \textit{whd-body} & WH-question body* & 0.1 & \textit{whd\_body}\\
      \textit{rhd-body} & Relative clause body* &0.1& \textit{rhd\_body}\\
      \textit{body} & Complementizer body* &2& \textit{body}\\
      \textit{cmp} & Complementizer & 2& \textit{cmp}\\
      \textit{cnj} & Conjunct &4.3& \textit{cnj}\\
      \textit{crd} & Coordinator &1.9& \textit{crd}\\
      \textit{invdet} & Syntactic head of a noun phrase &9.7& \textit{invdet}\\
      \textit{dlink} & Discourse link & 0.2 & N/A**\\
      \textit{dp} & Discourse part & 0.8 & N/A**\\
      \textit{hd} & Phrasal Head & 27.8 & N/A*** \\
      \textit{hdf} & Final part of circumposition & $<$0.1& \textit{hdf}\\
      \textit{ld} & Locative Complement &0.5& \textit{ld}\\
      \textit{me} & Measure Complement &0.1& \textit{me}\\
      \textit{mod} & Modifier &16.4& \textit{mod}\\
      \textit{mwu} & Multi-word part &5.1& N/A**\\
      \textit{nucl} & Nuclear Clause &0.5& N/A**\\
      \textit{obcomp} & Comparison Complement &0.1& \textit{obcomp}\\      
      \textit{obj1} & Direct Object &10.8& \textit{obj1}\\
      \textit{obj2} & Secondary Object &0.2& \textit{obj2}\\
      \textit{pc} & Prepositional Complement &10.6& \textit{pc}\\      
      \textit{pobj1} & Preliminary Direct Object &$<$0.1& \textit{pobj1}\\      
      \textit{predc} & Predicative Complement &1.3& \textit{predc}\\      
      \textit{predm} & Predicative Modifier &0.1& \textit{predm}\\      
      \textit{sat} & Satellite &0.2& N/A**\\
      \textit{se} & Obligatory Reflexive Object &0.7& \textit{se}\\      
      \textit{su} & Subject &6.9& \textit{su}\\
      \textit{sup} & Preliminary Subject &$<$0.1& \textit{sup}\\
      \textit{svp} & Separable Verbal Participle &0.7& \textit{svp}\\   
      \textit{vc} & Verbal Complement &2.8& \textit{vc}\\
      \textit{tag} & Appendix &0.1& \textsc{tag}\\
      \textit{whd} & WH-question head &0.1&N/A***\\
      \textit{rhd} & Relative clause head &0.1& N/A***\\
\end{tabularx}
\caption[Extracted Dependency Labels]{Dependency relations and their corresponding implication labels.\\
(*): No distinction between the three subtypes of \textit{body} is made in the original annotation.\\
(**): Not used as an implication or not covered by the extraction.\\
(***): Head types not projecting an implication label.}\label{table:colors}
\end{table}

\begin{figure}
    \centering
    \includegraphics[scale=0.29]{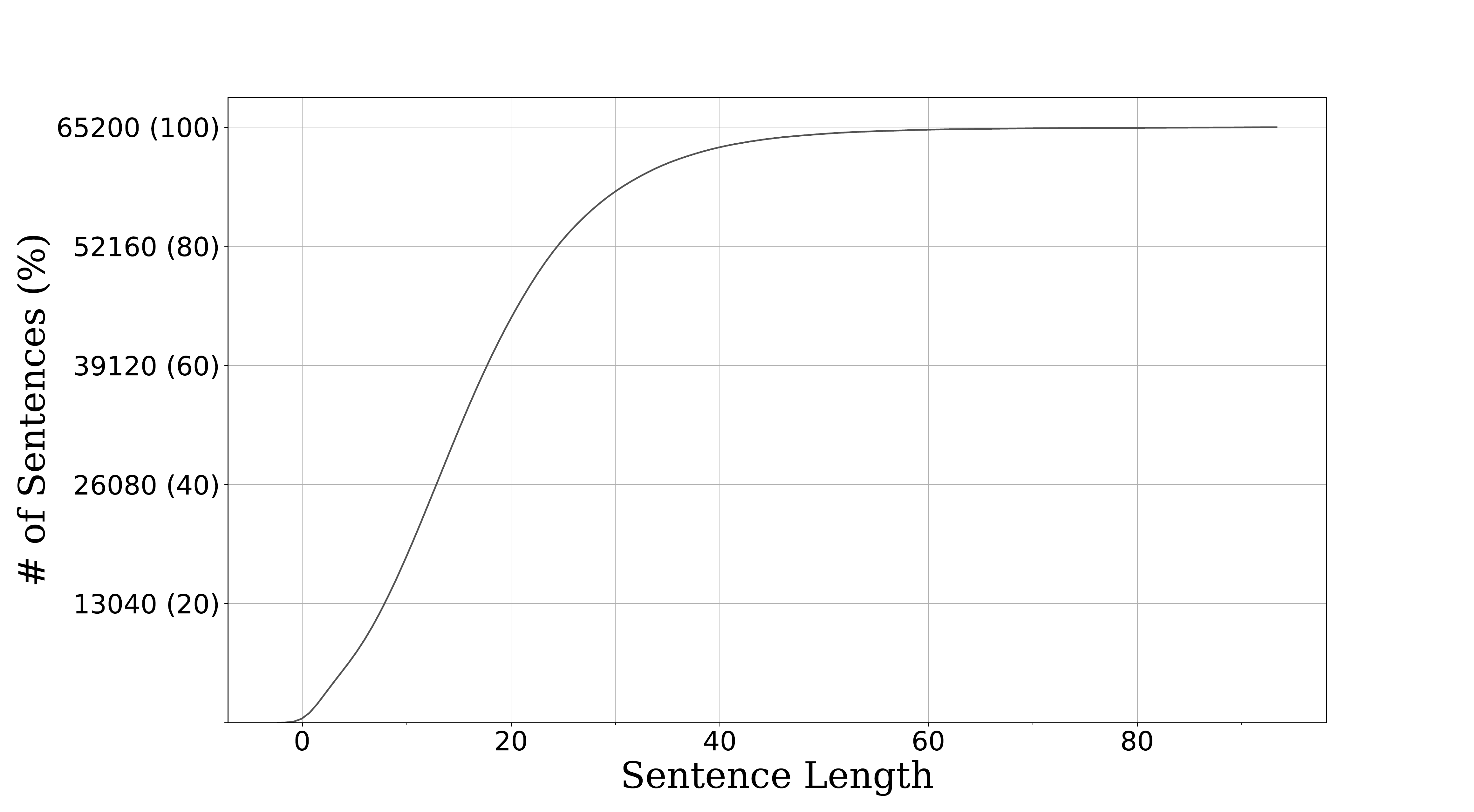}
    \caption[Lassy-Small Sentence Lengths]{Cumulative sentence lengths in the original corpus, including punctuation.}
    \label{fig:lassy_sentence_lens}
\end{figure}

\paragraph{Sentence Lengths}
Figure~\ref{fig:lassy_sentence_lens} displays the cumulative distribution of sentence lengths in the corpus.

\section{Extraction Process}
\subsection{Extraction Algorithm}
\label{subsec:extraction_alg}
Algorithms~\ref{alg:utils} and~\ref{alg:ra} summarize the type-assignment process, subdivided into components. 
The first revolves around utility functions that implement contextual assignment of phrasal arguments, root nodes and modifiers, as well as a method that concerns the binarization of multi-argument functors.
The second presents a method for iterating over a DAG.
During the iteration, the DAG's nodes are decorated with types by calling the prior functions as needed.
The goal here is to draw a rough sketch of the overall process rather than delve into a detailed description.
In that sense, the presentation abstracts away from the minutiae of the implementation, but also makes a few simplifying assumptions, namely the full compatibility between the system's routines and its input, as well as an absence of exceptional cases.
These simplifications will be easier to address after the extraction's core has been exposed.

\begin{algorithm}{
\caption{Type Assignment Utilities}\label{alg:utils}
\begin{algorithmic}[1]
\Procedure{Trans}{node $N$}\label{alg:ra:trans:start}\\
\Comment{Translates independent nodes to atomic types}
\If{is\_terminal($N$)}
    \State $\textbf{return}$ pos\_table[$N$.part\_of\_speech]
\Else
    \State $\textbf{return}$ cat\_table[$N$.syntactic\_category]
\EndIf
\EndProcedure\label{alg:ra:trans:end}
\\
\Procedure{TypeAssign}{node $N$, dependency $dep$, parent\_type $\textsc{p}$}\\
\Comment{Assigns contextually-informed types to modifiers and arguments}
\label{alg:ra:typeassign:start}
\If{$dep$ $\in$ mod\_labels}
    \State $d$ $\gets$ dep\_table[dep]
    \State $\textbf{return}$ $\textsc{p}\myrightarrow{d}\textsc{p}$
\Else
    \State $\textbf{return}$ $\textsc{Trans}(N)$
\EndIf
\EndProcedure
\label{alg:ra:typeassign:end}
\\
\Procedure{MakeComplex}{Arguments $A$, Result \textsc{r}}\\
\Comment{Converts a list of type \& dependency pairs and a result into a binarized functor}
\label{alg:ra:makecomplex:start}
    \State $A$ $\gets$ sort($A$)
    \State $D$ $\gets$ map(snd, $A$)
    \State $d_1, \dots, d_n$ $\gets$ map(dep\_table, $D$)
    \State \textbf{return} $\textsc{a}_1$ $\myrightarrow{$d_1$}$ $\textsc{a}_2$ \dots $\textsc{a}_N$ $\myrightarrow{$d_N$}$ \textsc{r}
    \label{alg:ra:makecomplex:fold}
\EndProcedure
\label{alg:ra:makecomplex:end}
\end{algorithmic}
}
\end{algorithm}

\paragraph{Arguments and Modifiers}
The simplest, primitive component is a function \textsc{Trans} (\ref{alg:ra:trans:start}-\ref{alg:ra:trans:end}).
It simply distinguishes between leaves and non-terminals, extracts the part-of-speech  or syntactic category tag respectively, and uses the corresponding translation table to map it to an atomic type.
The instantiation of the type translation tables is presented in Table~\ref{table:lex}.
\textsc{Trans} is a context-agnostic type-assignment function; it is used to assign direct phrasal dependents and root nodes, the typing of which depends neither on their ancestors nor their descendants.

Modifiers require different treatment; their types are instances of the parametrically polymorphic scheme:
\[
\textsc{t} \myrightarrow{d} \textsc{t}
\]
where \textsc{t} the type of the phrase being modified and $d$ the dependency label enacted by the modifier (an element of closed set).
In the current implementation, aside from plain modifiers (\textit{mod}), appositions (\textit{app}) and predicate modifiers (\textit{predm}) are also treated as modifying labels.
To correctly instantiate modifier types, we implement a minimally context-aware typing function, \textsc{TypeAssign} (\ref{alg:ra:typeassign:start}-\ref{alg:ra:typeassign:end}).
Given a node, the type of the node's parent and the dependency linking the two, it either assigns a type according to the above scheme, if the dependency is modifying, or defaults to \textsc{Trans} otherwise.
In the former case, the dependency is translated into an implication name through a translation table; its running instantiation is presented in Table~\ref{table:colors}.

\paragraph{Head Types}
At this point we need to recall that our complex types are unary functors, i.e. they have a single argument and produce a single result.
Yet the DAGs' branchings are not binary, and the pervasiveness of word-order freedom in the language means that a large degree of argument permutation is to be anticipated within multi-argument types.
Acknowledging these permutations is meaningful for parsing, but less so for semantic tasks; the order of words matters little if their dependencies are already specified.
Additionally, permitting multiple different representations for types that are functionally equal has the negative side-effect of enlarging the size of our extracted type lexicon.
As a counter-measure, we may enforce unary functor types by setting up an order over types, dictating their proximity to the end-result when they participate as arguments in the construction of a complex type.
Function \textsc{MakeComplex} (\ref{alg:ra:makecomplex:start}-\ref{alg:ra:makecomplex:end}) utilizes such an order, implemented as a sorting of a list of pairs of types \& dependencies.
The concrete implementation may remain unspecified for now; we will return to it at a later stage.
The main functionality is generally independent of the sorting criterion; the type-dependency pairs are sorted, and their dependencies are translated into implication names.
Then, a unary functor is built up by folding over the sequence in reverse order, using the result type as the initial argument and the next sequence item as the new outer argument (\ref{alg:ra:makecomplex:fold}).

\paragraph{Recursive Type Assignment}
Finally, the bulk of the work is carried out by \textsc{RecursiveAssignment} (\ref{alg:ra:recursiveassignment:start}-\ref{alg:ra:recursiveassignment:end}).
Given a parent node, its type, and a partially filled dictionary mapping nodes to types, it is responsible for carrying out the type-assignment for the part of the DAG that lies below.
First, it inspects whether the parent node is a leaf, in which case the current call terminates.
Otherwise, it inspects the node's daughters and their corresponding edges, and selects for the head (\ref{alg:ra:recursiveassignment:select_head}). 
Note that for this to work, a deterministic process for selecting a head is necessary; in other words, all structures are considered headed. 
Head-daughters can be told apart by their incoming dependency label, which is either of head (\textit{hd}), relative-clause head (\textit{rhd}), wh-clause head (\textit{whd}), comparative (\textit{comp}) or coordinator (\textit{crd}).
An empty list is then instantiated for arguments to be filled by type \& dependency pairs.
Next, the algorithm iterates over the node's daughters, deciding the type of each via \textsc{TypeAssign}.
For each daughter, an empty list of embedded arguments, also being type \& dependency pairs, is also instantiated.
A full downwards iteration of the DAG, starting from that daughter is carried out.
If at any time a descendant node is identified with the current call's head, its type, as inferred by \textsc{TypeAssign} called with the new-found inner dependency and the current daughter as arguments, is added to the embedded argument list.
After this iteration is complete, the dependency between the daughter and the current parent node is inspected; if it bears no secondary marking, the type dictionary is updated and \textsc{RecursiveAssigment} is called anew on the daughter using its inferred type.
Recursive calls of this function essentially fill in types for the DAGs' elements in a depth-first, left-biased manner.

When the control returns to the caller function, the daughter's type is updated by applying \textsc{MakeComplex} using the embedded argument list and the previous daughter type.
If the embedded argument list is populated, one of the daughter's (not necessarily immediate) descendants is also its sibling.
The updated type then reflects the incomplete nature of the DAG rooted at this daughter, and the need for hypothetical reasoning to resolve the lexically immaterial arguments.
Finally, if the main dependency between the parent node and the daughter currently examined is not a modifier, the daughter's type is added into the argument list.

After all daughters have been processed, iterated through and their types assigned, the next step is to construct the head's type.
Since the embedded arguments already account for gaps, the head is simply the functor from the argument types (as collected in the arguments list) to the parent's type.

It is now straightforward to annotate a full DAG by simply selecting its root node, finding its type via a direct translation, initializing an empty dictionary, and filling it via \textsc{RecursiveAssignment} which is simply called with the prior arguments (\ref{alg:ra:annotatedag:start}-\ref{alg:ra:annotatedag:end}).

\begin{algorithm}{
\caption{Type Assignment Process}\label{alg:ra}
\begin{algorithmic}[1]
\Procedure{RecursiveAssignment}{node $N$, node\_type \textsc{p}, type\_dict $\mathcal{D}$}\\
\Comment{Depth-first recursion over a partial DAG}
\label{alg:ra:recursiveassignment:start}
    \If{is\_terminal($N$)}
        \State $\textbf{exit}$
    \EndIf
    \State $head \gets$ select\_head($N$.daughters)
    \label{alg:ra:recursiveassignment:select_head}
    \State $arguments \gets []$
    \For{(daughter $d$, dependency $dep$) \textbf{in} $N$.daughters $\wedge$ \{head\}}
        \State $daughter\_type \gets$ \textsc{TypeAssign}($d$, $dep$, \textsc{p})
        \State $embedded\_args \gets$ []
        \label{alg:ra:recursiveassignment:findhot:start}
        \For{(descendant $f$, dependency $inner\_dep$) \textbf{in} $d$.descendants}
            \If{$f$ == head}
                \State $inner\_type \gets$ \textsc{TypeAssign}($f$, $inner\_dep$, daughter\_type)
                \State $embedded\_args$.append([$inner\_type$, $inner\_dep$])
            \EndIf
        \EndFor
        \label{alg:ra:recursiveassignment:findhot:end}
        \If{is\_primary($dep$)}
            \State $\mathcal{D}[d] \gets daughter\_type$
            \State \textsc{RecursiveAssignment}($d$, $daughter\_type$, $\mathcal{D}$)
        \EndIf
        \State $daughter\_type \gets$ \textsc{MakeComplex}($embedded\_args$, $daughter\_type$)
        \If{$dep$ $\notin$ mod\_labels}
            \State $arguments$.append([$daughter\_type$ $\otimes$ $dep$])
        \EndIf
    \EndFor
    \State $\mathcal{D}$[$head$] $\gets$ \textsc{MakeComplex}($arguments$, \textsc{p})
\EndProcedure
\label{alg:ra:recursiveassignment:end}
\\
\Procedure{AnnotateDAG}{DAG $G$}\\
\Comment{Type annotates a full parse DAG}
\label{alg:ra:annotatedag:start}
    \State $r$ $\gets$ $G$.root
    \State $\mathcal{D}$ $\gets$ \{\}
    \State \textsc{t} $\gets$ \textsc{Trans}($r$)
    \State \textsc{RecursiveAssignment}($r$, \textsc{t}, $\mathcal{D}$)
    \State \textbf{return} $\mathcal{D}$
\EndProcedure
\label{alg:ra:annotatedag:end}
\end{algorithmic}
}
\end{algorithm}

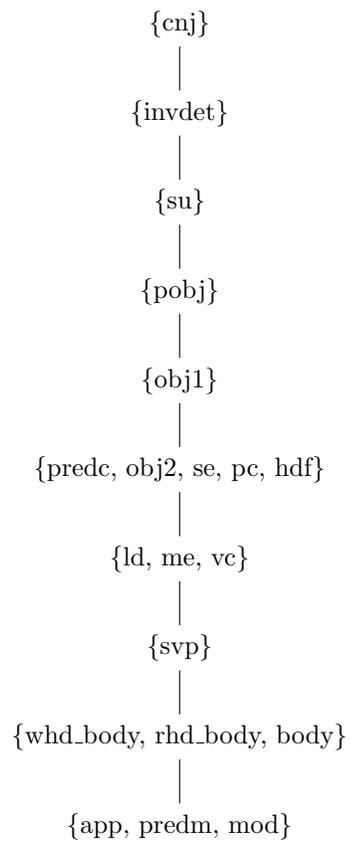
\begin{figure}
\centering
\begin{tikzpicture}
\matrix (a) [matrix of nodes, column sep=0.5cm, row sep=0.6cm]{
\{cnj\} \\
\{invdet\} \\
\{su\} \\
\{pobj\} \\
\{obj1\} \\
\{predc, obj2, se, pc, hdf\} \\
\{ld, me, vc\} \\
\{svp\} \\
\{whd\_body, rhd\_body, body\}\\
\{app, predm, mod\}\\
};

\draw (a-1-1) -- (a-2-1) ;
\draw (a-2-1) -- (a-3-1);
\draw (a-3-1) -- (a-4-1);
\draw (a-4-1) -- (a-5-1);
\draw (a-5-1) -- (a-6-1);
\draw (a-6-1) -- (a-7-1);
\draw (a-7-1) -- (a-8-1);
\draw (a-8-1) -- (a-9-1);
\draw (a-9-1) -- (a-10-1);

\end{tikzpicture}
    \caption[Dependency Role Obliqueness Ordering]{Dependency role obliqueness ordering. Lower roles take precedence in priority over higher ones (appear closer to the result)}
    \label{fig:lattice}
\end{figure}

\paragraph{Argument Ordering}
To consistently binarize multi-argument complex types, we establish a strict partial order over the set of dependency relations.
The order is motivated by the obliqueness hierarchy present within verbal arguments, loosely inspired by~\cite{dowty}\footnote{Even though this ordering is normally established relative to the result category, we can simplify the process by considering that incompatible arguments will not compete for hierarchy within a single functor --- thus, no more than one partial order is necessary.}.
The sorting algorithm then arranges a sequence of type \& dependency pairs based solely on the latter.
Figure~\ref{fig:lattice} depicts the Hasse diagram of the poset, where incomparable items are collapsed into a single set.
These either never co-exist as arguments to the same functor due to being subclasses of the same dependency (such as the various instances of clause bodies) or occupying mutually exclusive positions (as in the complemental verbal arguments), or when they do their order is irrelevant (e.g. modifiers).
In the case of incomparable items appearing simultaneously, an alphabetical default is applied.
Even though conjuncts and modifiers do not directly belong in the obliqueness hierarchy, they are positioned in the poset in a way that ensures type readability and sanity; for instance, modifiers are always placed last, so if a functor corresponds to a modifier construction, the resulting modifier type will always appear last.
Similarly, coordinators are not derived but may still partake in higher-order types; as such, conjunction dependencies always appear first.

\subsection{Transformations and Exceptions}
\label{subsec:transformations}

The extraction algorithm, as specified earlier, makes a few assumptions pertaining to the DAGs' structure, which are not always valid.
In order to bring the corpus to a format that is maximally compatible with the algorithm, a series of preprocessing steps, in the form of conversions of the original annotation, become necessary.
In addition, on several occasions correct type inference requires a few alterations to the general algorithm flow, implemented in the form of conditional case management.
The paragraphs below detail these transformations and exceptions.
Most transformations are accompanied by an example of their application on an actual corpus sample (identified by its filename), visualizing their effect on the parse DAG.

\paragraph{From Trees to DAGs}
To begin with, ``phantom'' nodes are useful for maintaining a tree-like view of the DAG, but are otherwise hard to utilize.
In only being linked to a subset of the lexical item's ancestors, they obfuscate the overall parse structure, necessitating multiple iterations to distinguish and type-assign.
For that reason, we remove node copies by allowing multiple incoming edges onto a single node.
To distinguish between edges provided by the original treebank and edges inserted by the transformation, we refer to the former as primary and the latter as secondary.
Any non-root node is always associated with exactly one incoming primary edge, and zero or more secondary edges.
This invertible transformation serves mostly representational and algorithmic needs, in the sense that it permits us to refer to a node as a single unit regardless of the multiplicity of roles it assumes within the parse structure. 

\begin{figure}[t]
    \begin{subfigure}[t]{0.49\textwidth}
        \centering
        \includegraphics[scale=0.49]{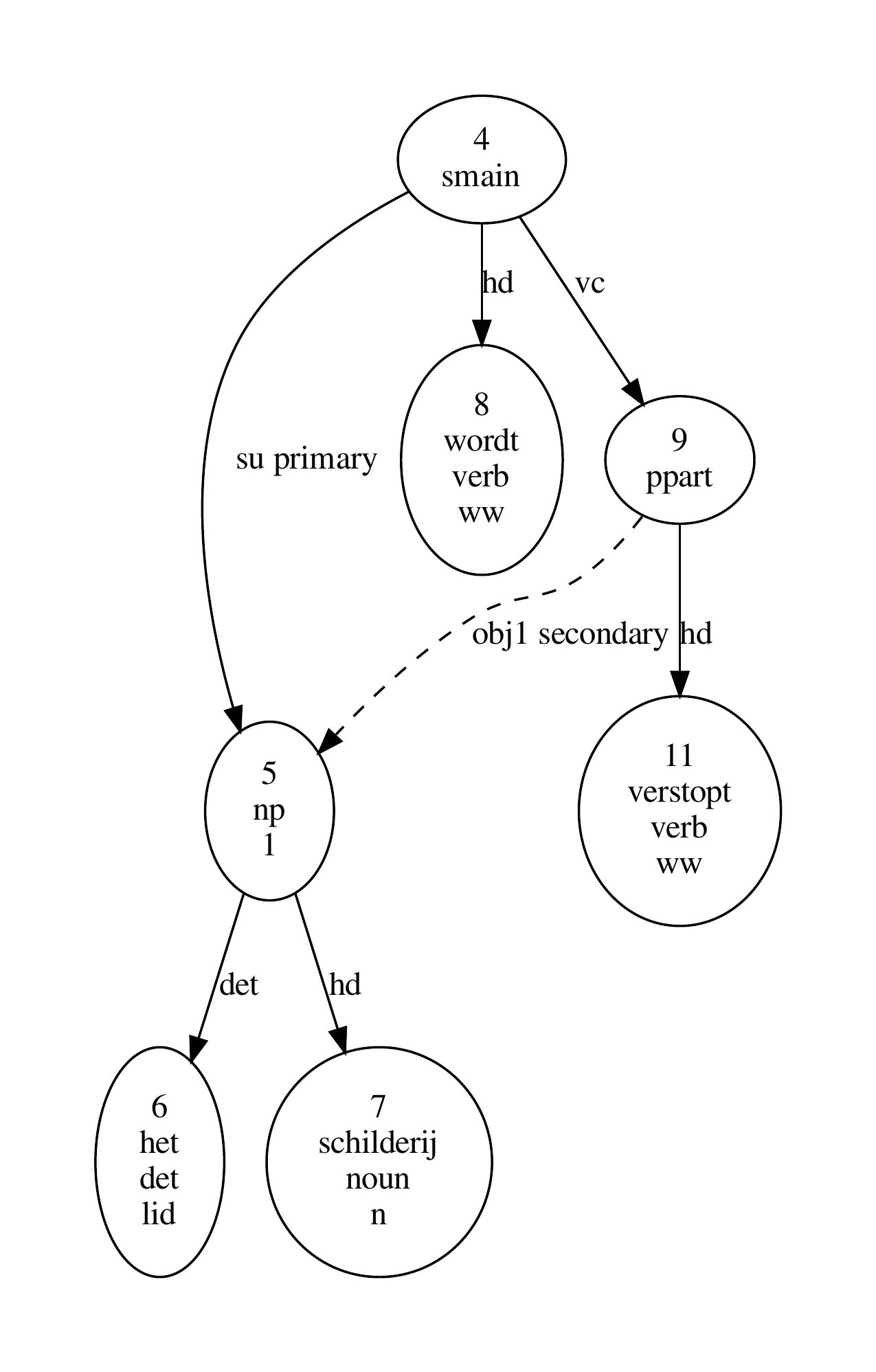}
        \caption{Before removal}
        \end{subfigure}
    \begin{subfigure}[t]{0.49\textwidth}
        \centering
        \includegraphics[scale=0.49]{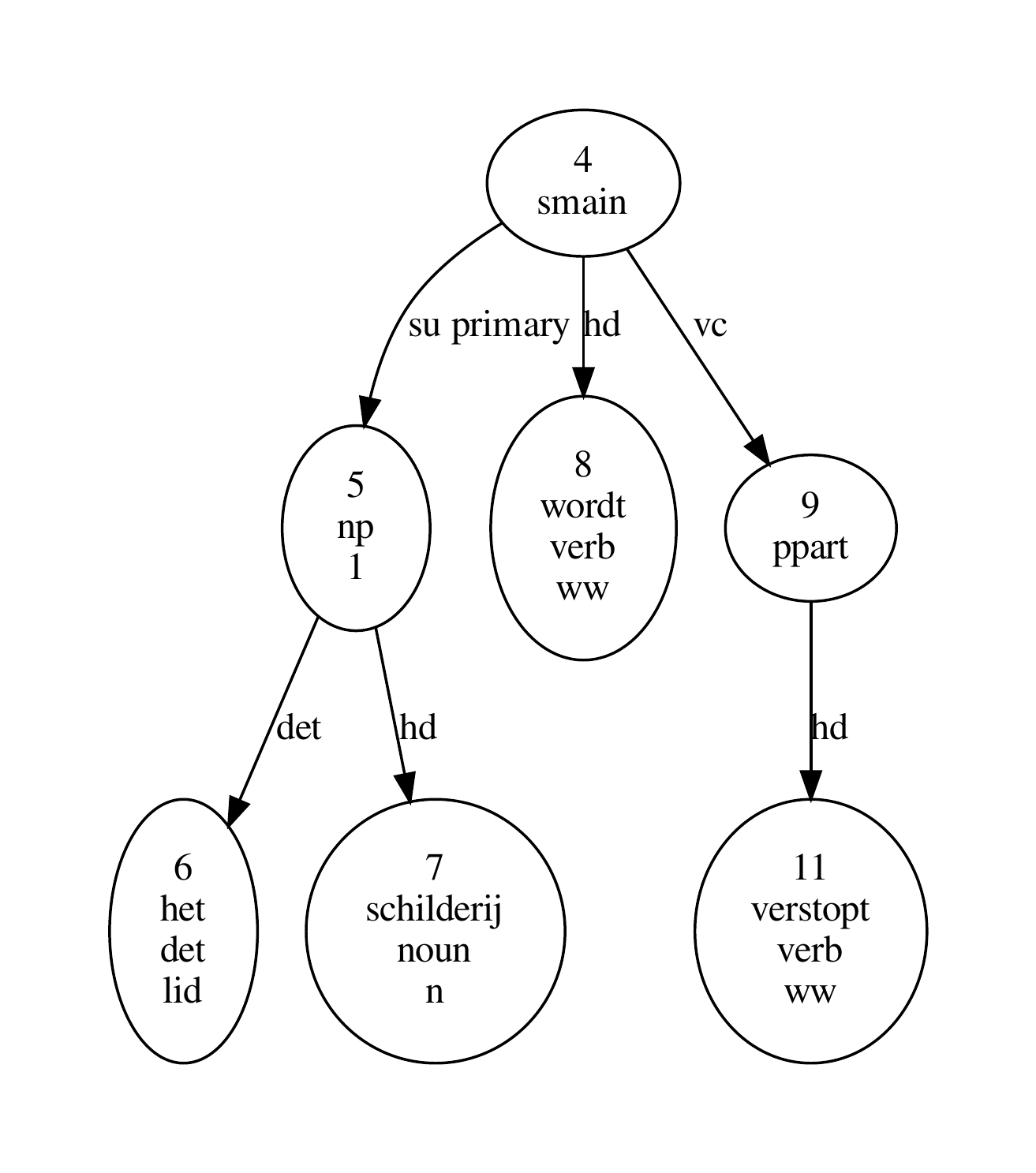}
        \caption{After removal}
    \end{subfigure}\vfill
    \caption[Abstract Semantic Argument Removal]{Derivation structure before (a) and after (b) abstract semantic argument removal for the sentence ``het schilderij wordt verstopt'' (\textit{The painting was hidden}) [\texttt{wiki-9720.p.23.s.1}]. The phrase ``het schilderij'' acts as both a subject argument for the main sentence verb ``is'' and the understood object for the participial head ``verstopt''. After the transformation, the second dependency is erased. }
    \label{fig:abstract_arg}
\end{figure}

\paragraph{Abstact Semantic Arguments}
Lassy's annotation scheme includes secondary edges for verbs' abstract semantic arguments embedded under past participial phrases and infinitival clauses.
These dependencies account for semantic flows within sentences, but in doing so escape from the boundaries imposed by the phrase-local syntax.
Furthermore, the distinction between primary and secondary edge labeling in the case of abstract arguments is not structurally consistent, injecting a source of unnecessary ambiguity for the type extraction.
In light of the above, we first homogenize the labeling by setting the node that resides at the higher level of the graph as the primary one in all ambiguous constructions.
Since embedded clauses are systematically dependents of primary clauses, this may be viewed as equating between sentence-primary roles and primary edges.
We can then consistently remove all abstract arguments by simply filtering out secondary links with a subject or object label that occur between past participles/infinitives and their daughters, if the latter have a primary subject or object link with an (immediate or otherwise) ancestor of their direct parents.
An example of the transformation is presented in Figure~\ref{fig:abstract_arg}.

\paragraph{Noun-Phrase Headedness}
When determiners and nouns are participating in the formation of a phrase, nouns are assigned the head-role to which determiners act as dependents.
Without making any argument for determiner-headed phrases, we acknowledge two problems with treating nouns as functors.

First, for the type grammar itself, it would mean that nouns are contextually-typed --- a noun in isolation might act as an independent noun-phrase, whereas a noun paired with a determiner is a functor type from determiners to noun-phrases.
Secondarily, anticipating a future semantic interpretation of our type system, we would like nouns to be viewed as atomic, individual objects rather than functions, for reasons both practical and conceptual.
To avoid the linguistic repercussions of considering determiners as heads, we opt to make a distinction between headedness and functoriality in this particular case.
Treating determiners as the phrasal functors resolves both of the above problems, as they consistently appear paired to a nominal (which may be seen as their argument), and can be trivially interpreted into identity maps in the semantic space (thus preserving the content of their arguments).

\begin{figure}[t]
    \begin{subfigure}{0.49\textwidth}
        \centering
        \includegraphics[scale=0.48]{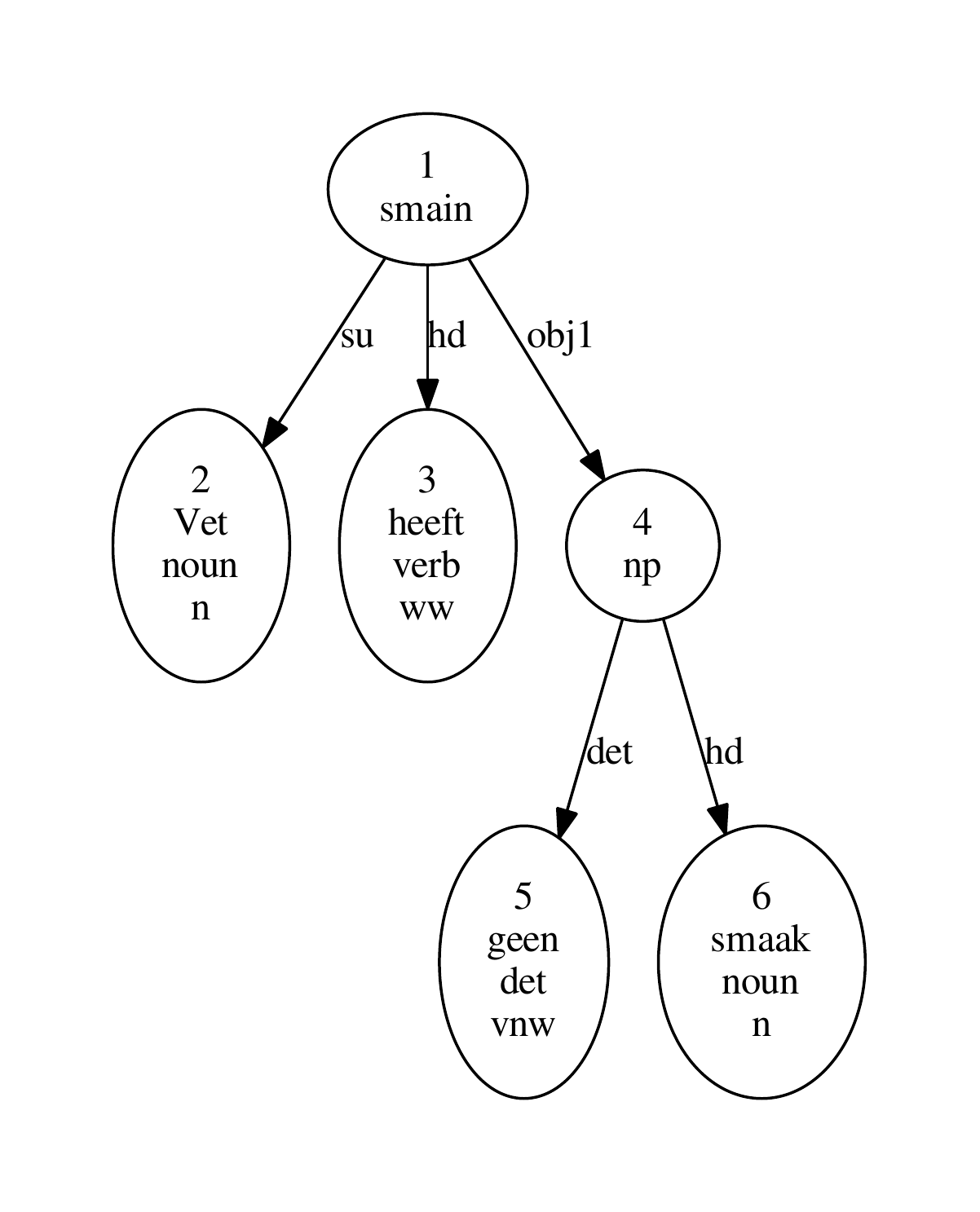}
        \caption{Before swapping}
    \end{subfigure}
    \begin{subfigure}{0.49\textwidth}
        \centering
        \includegraphics[scale=0.48]{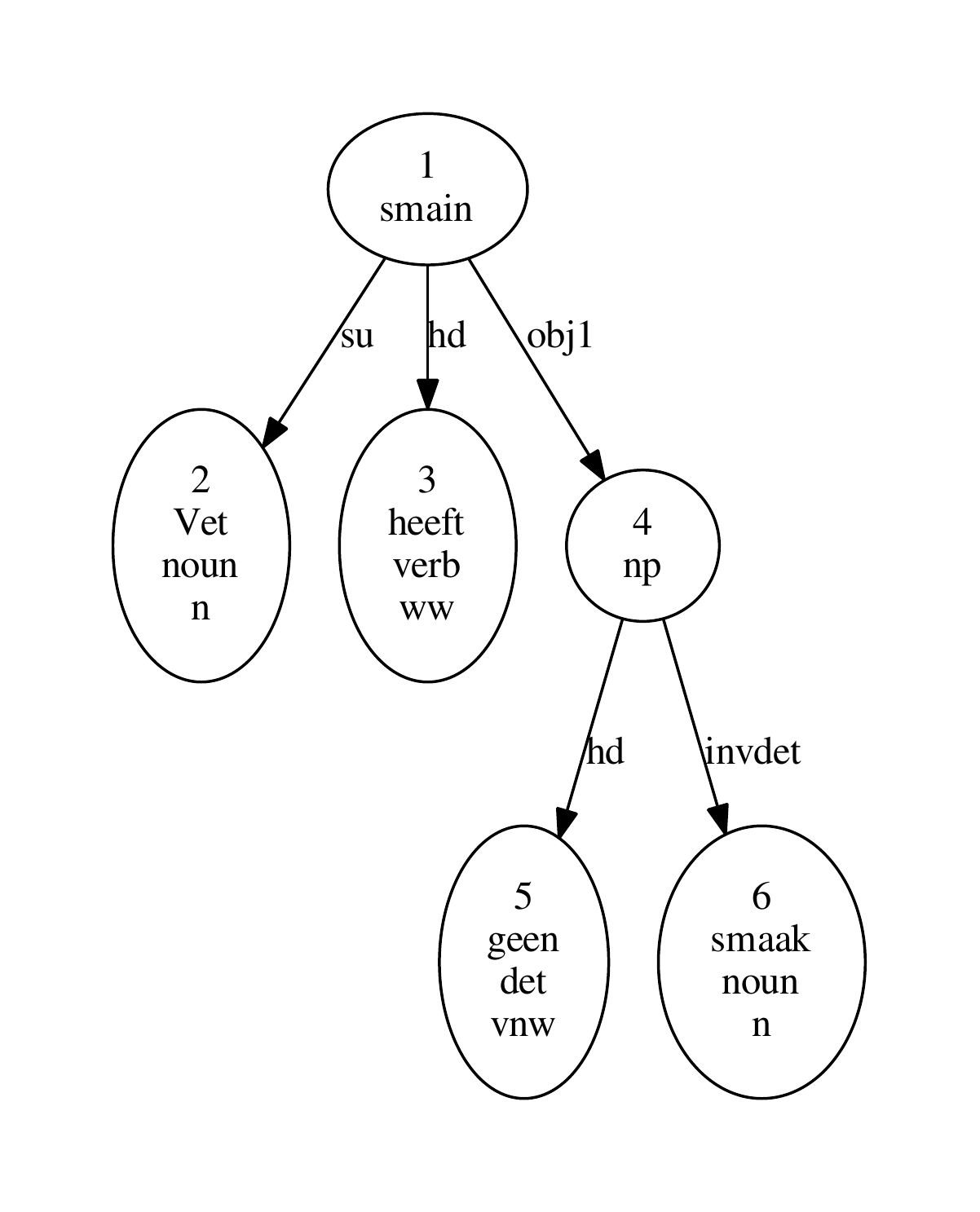}
        \caption{After swapping}
    \end{subfigure}
    \caption[Noun-Phrase Head Swapping]{Dependency relation naming before (a) and after (b) swapping noun-phrase heads for the example sentence ``Vet heeft geen smaak' (\textit{Fat has no taste}) [\texttt{WR-P-P-I-0000000130.p.1.s.1}]. The pronoun ``geen'' acts as a determiner to the noun-phrase ``geen smaak''. After the transformation, it is assigned the head role, with the noun-phrase being assigned the dual label \textit{invdet}.}
    \label{fig:invdet}
\end{figure}

For these reasons, we choose to alter the dependency labels internal to noun-phrases as follows; determiner labels are set to head labels (for compatibility purposes) and head labels occupied by nouns are set to a dual of the determiner, which we name \textit{invdet}.
An example of the transformation is shown in Figure~\ref{fig:invdet}.

\begin{figure}[t]
    \begin{subfigure}{0.49\textwidth}
        \centering
        \includegraphics[scale=0.48]{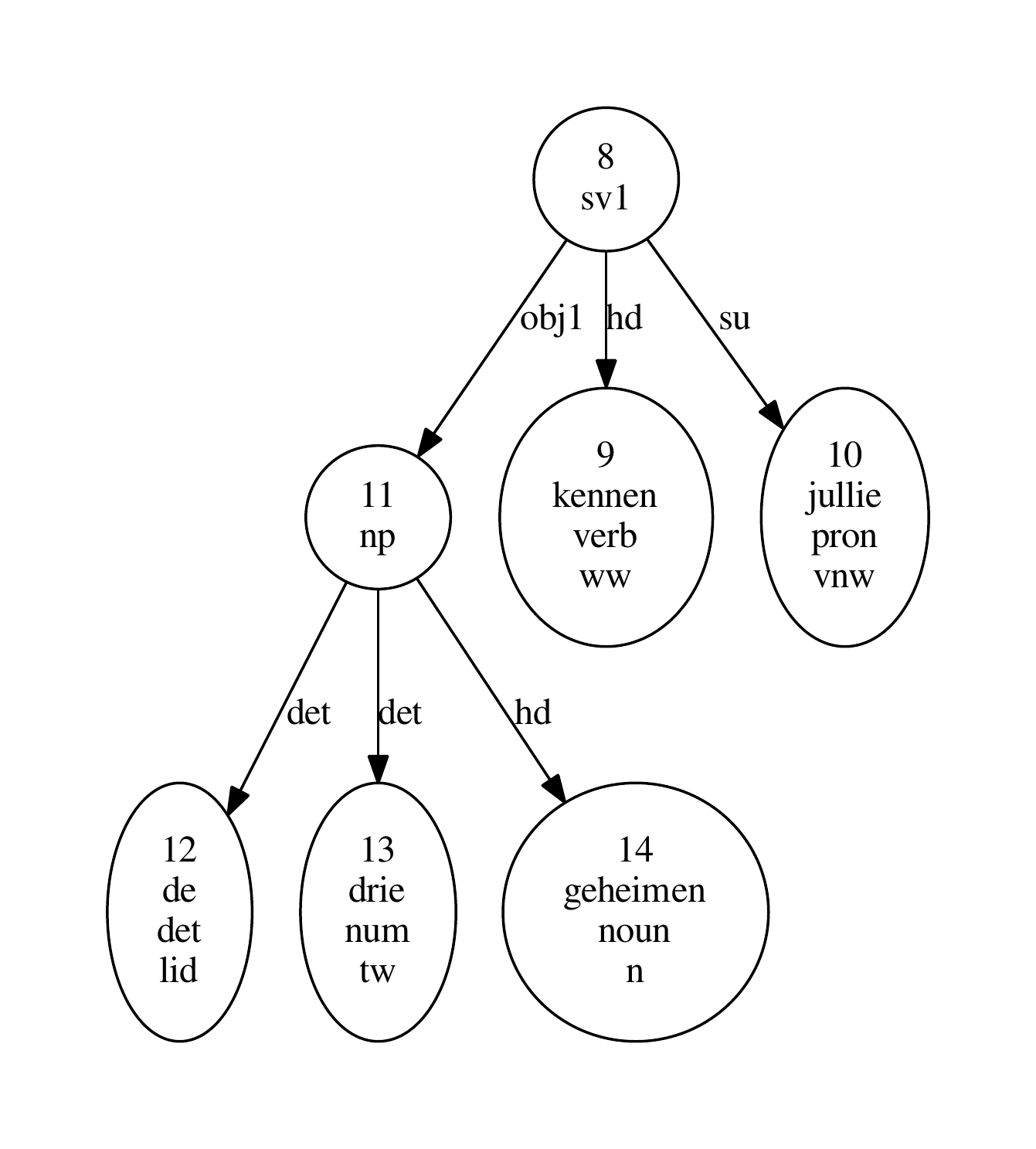}
        \caption{Before renaming}
    \end{subfigure}
    \begin{subfigure}{0.49\textwidth}
        \centering
        \includegraphics[scale=0.48]{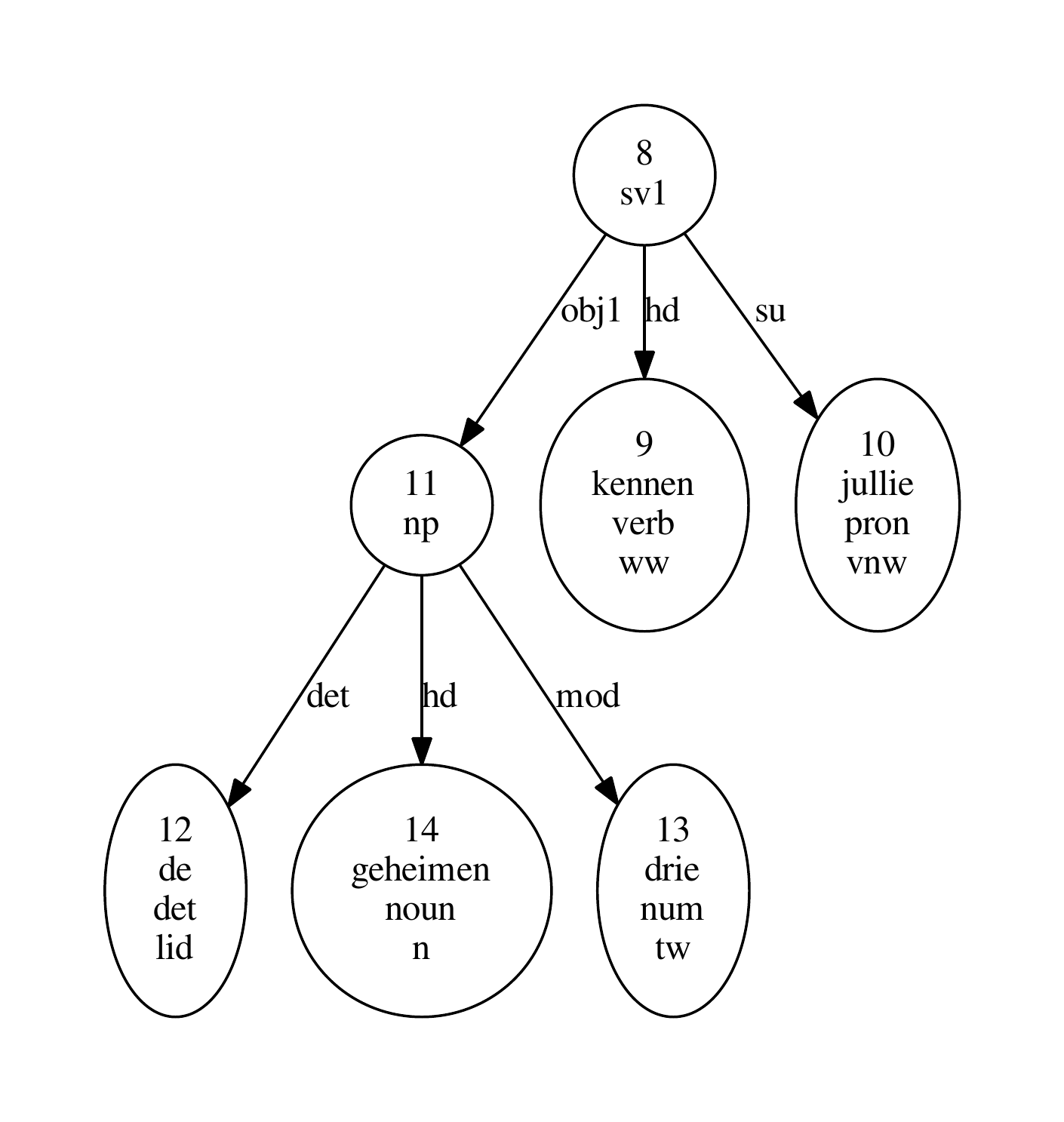}
        \caption{After renaming}
    \end{subfigure}
    \caption[Numeral Determiner Relabeling]{Dependency relation naming before (a) and after (b) the transformation of numeral determiners into modifiers for the example sentence ``kennen jullie de drie geheimen'' (\textit{do you know the three secrets}) [\texttt{WR-P-P-B-0000000001.p.90.s.1}]. Before the renaming, the numeral ``drie'' acts as an additional determiner for the noun-phrase ``de drie geheimen''. After the transformation, its dependency is altered to a modifier.}
    \label{fig:tw_to_mod}
\end{figure}

\paragraph{Numerals and Multiple Determiners}
The transformation described above is not always as straightforward.
Although it is the case that each noun-phrase has a single head-noun, it is not the case that it has a single determiner.
The most common instance of a noun-phrase with more than one determiner are phrases that contain numerals and count-words, the dependencies of which are also labeled as determiners.
Examples of such phrases include ``de twaalf profeten'' (\textit{the twelve prophets}), ``de beide zijluiken'' (\textit{the two side-panels}) etc. 
These cases are easy to solve by simply adjusting the numerals' dependency label into modifier (\textit{mod}), when they co-exist with a real determiner.
If a numeral is the sole companion of a nominal, we allow it to retain its determiner role, meaning it later gets converted into the phrasal functor.
An example of the transformation is depicted in Figure~\ref{fig:tw_to_mod}.

A less frequent but more problematic instance is that of determiner pairs, i.e. words that jointly form a determiner when occurring one next to the other, such as for example ``geen enkele'' (\textit{not a single one}), ``een enkele'' (\textit{one}), etc.
As hierarchy is not present on such expressions (no word dominates the phrase over the other), it is ill-advised to insert makeshift structure in the form of a binary branching determiner phrase.
To bypass this issue, we set the first item's label to head, and assign a placeholder label \textit{\_det} to the second one. 
During the type assignment process, leaves carrying the special label default to the \textsc{\_det} type, to be read as \textit{second item of a determiner pair}. 
Functionally, it is as a marking above the logical language, denoting that this word needs to be collapsed with its immediately proximal neighbour to the left, forming a multi-word phrase that inherits the neighbour's type.

\paragraph{Head-Body Pairs}
A common construction in Lassy-Small is that of head-body pairs, which is used to annotate relative clauses, wh-questions and phrases, complementizers and comparisons.
To distinguish between the above, the head label is subclassed into three different instances; \textit{rhd} (for relative clause heads), \textit{whd} (for wh-question- and phrase-heads) and \textit{hd} (for all remaining scenarios).
Seeing as head labels do not come out in the type system (being that these are indicators of functoriality rather than refinements of implications), we transfer the refinement from the heads onto their respective body labels, replacing \textit{body} with \textit{rhd\_body} or \textit{whd\_body} according to the context.

\paragraph{Unheaded and Multi-Headed Structures}
An operational assumption for the algorithm is that all complex phrases have exactly one head.
This assumption is not a product of the algorithm's design, but rather the distinguishing feature of phrase structure grammars, and the driving force behind compositionality in the case of categorial grammars.
Lassy's annotation scheme abides by a dependency-based formalism; that means that not all of the analyses satisfy this assumption.
The problem is manifested in two distinct ways; in unheaded structures, i.e. branchings where no clear distinction between functor and argument(s) can be made between a node's daughters, and multi-headed structures, i.e. branchings where more than one head candidates co-exist.

Examples of the first include multi-word phrases, conjunctions and discourse-level annotations.
Multi-headed structures occur indirectly in the context of determiner pairs as described earlier, but also directly in coordinator pairing and discourse-level annotations.
As no universal, algorithmic solution can be designed to account for all incompatible analyses, the following paragraphs are concerned with specific solutions targeted at particular problematic instances.

\subparagraph{Multi-Word Phrases}
The original annotation usually refrains from specifying the structure of multi-word phrases, but not consistently.
Multi-word phrases are for the most part depicted as a node with the syntactic category tag \textit{mwu} (for multi-word unit), all daughters of which are leaves with their edges marked as \textit{mwp}.
This approach detracts from the quality of the types that could be assigned in two ways; first and foremost, the absence of specified functor-argument relations in combination with the lack of an algorithmic way to construct them disallows us from providing precise types at the word level.
Furthermore, the category label of \textsc{mwp} corrupts type assignments for words and phrases outside the multi-word phrase, whenever it partakes as an argument to the formation of wider phrases.

To treat the first problem, we opt for the least drastic solution; multi-word phrases are collapsed into a single node, their outgoing edges are cut-off, and their corresponding lexical item becomes a contiguous span of the sentence rather than discrete items of it. 
The purpose of this transformation is to simply hide the uninformative edges.
In practical terms, no real information is erased (seeing as the edges may be re-introduced at any point with no extra knowledge).
The impact of this transformation on later applications (e.g. supertagging) is also relatively minor; all it requires is an extra component that identifies multi-word phrases and collapses them into a single element to be later used by the type-assignment process, which can be implemented as a chunking algorithm.

\begin{figure}[t]
    \begin{subfigure}[t]{0.49\textwidth}
        \centering
        \includegraphics[scale=0.43]{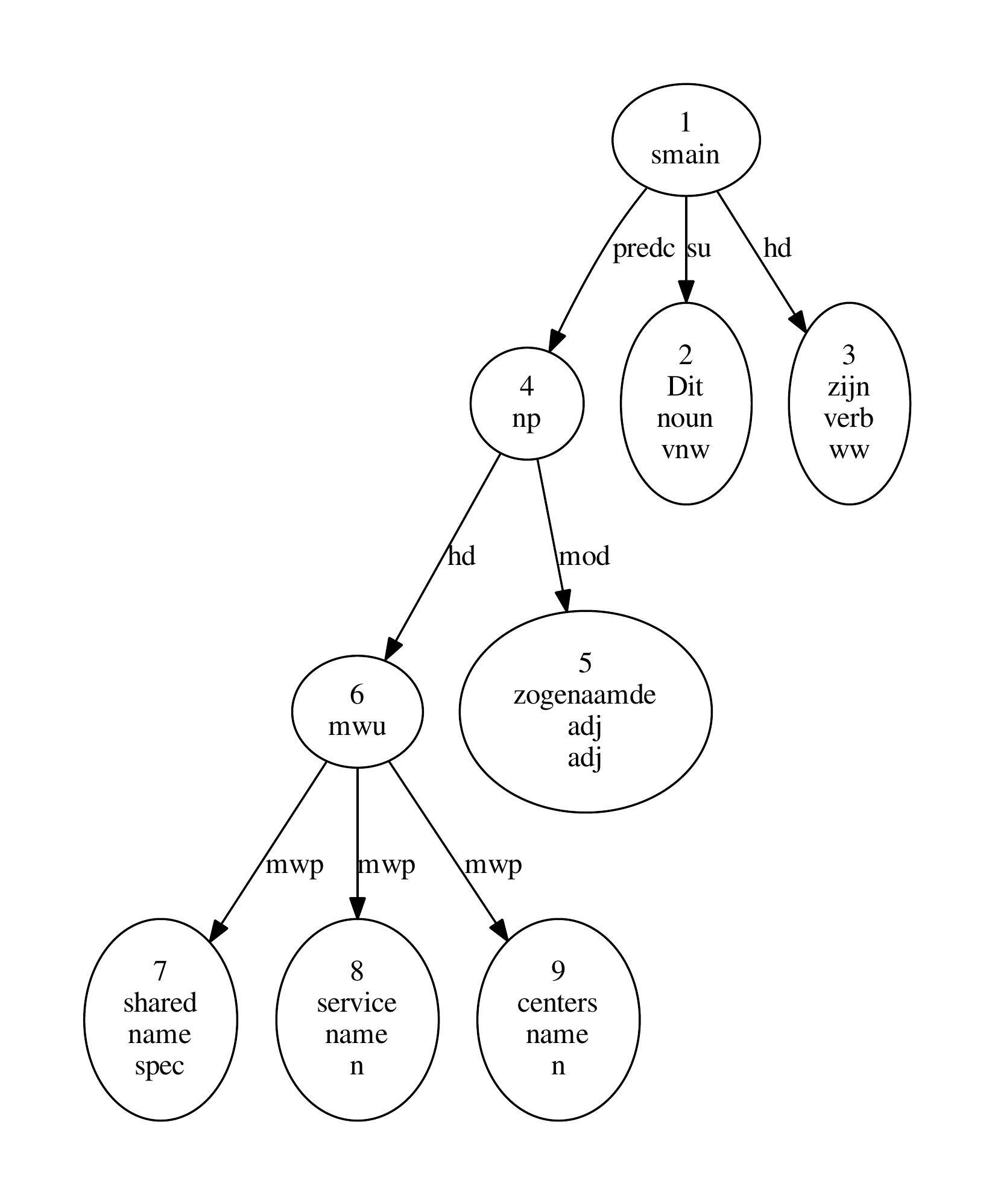}
        \caption{Before collapse}
    \end{subfigure}
    \begin{subfigure}[t]{0.43\textwidth}
        \centering
        \includegraphics[scale=0.48]{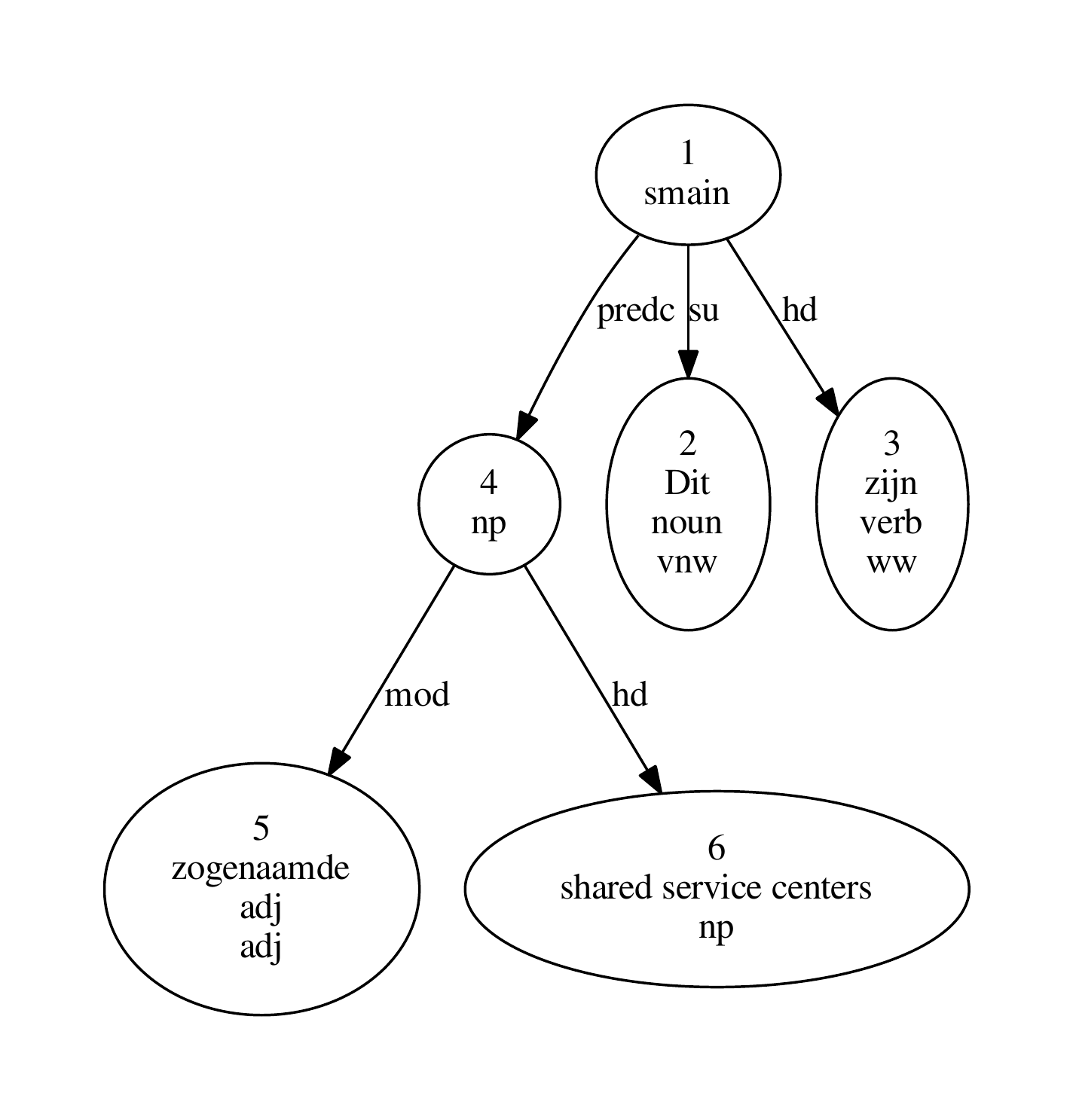}
        \caption{After collapse}
    \end{subfigure}
    \caption[Multi-Word Unit Collapse]{Derivation structure before (a) and after (b) the collapse of a mwu sub-DAG into a single phrase for the example sentence ``Dit zijn zogenaamde shared service centers'' (\textit{These are so-called shared service centers}) [\texttt{dpc-ibm-001315-nl-sen.p.85.s.1}]. Prior to the transformation, the phrase ``shared service centers'' is annotated as a sequence of nodes, rooted to a common ancestor. After the transformation, the three nodes are chunked into a single node that spans all three lexical items. The phrasal category of the new node is inferred by majority voting over its original children.}
    \label{fig:mwu}
\end{figure}

To avoid explicitly defining a type for multi-word phrases (thus also the consequences of such a type), we implement a \textit{majority voting scheme} with biased priorities that consistently converts the \textit{mwu} tag onto other tags.
The scheme relies on a simple but intuitive and general rule; the syntactic category of the multi-word phrase may be almost perfectly inferred by the categories of its daughters. 
Therefore, we simply sum the occurrences of each part-of-speech tag exhibited by its daughters. 
Most of the time, the set of these tags is singular, containing either the noun (\textit{n}) or the special (\textit{spec}) tag as its sole item, the latter being used as a catch-all tag.
If at least one of the above tags is present (possibly accompanied by a number of adjective sisters), we simply promote the multi-word unit to the noun phrase category.
Otherwise, if there is a significant difference between the occurrence ratios between tag pairs, we promote the major tag to its corresponding phrasal category. 
An example of the transformation is presented in Figure~\ref{fig:mwu}.

\subparagraph{Conjunction Phrases}
Conjunction annotations may pose problems in three different ways.

Quite like multi-word phrases, conjunction nodes are always assigned the syntactic category tag \textit{conj}, which is uninformative with respect to the conjunction's role or its contents.
The solution prescribed earlier in the form of a majority voting scheme is applicable in this case as well, modulo a minor difference.
Conjunctions are more general than multi-word units, in the sense that their constituents may carry a wider variety of tags.
When multiple daughter categories exist and their occurrence counts match, the decision on what tag to assign is not clear-cut; to account for conjunctions of uneven parts, we therefore expand upon the scheme by adding further biases.
The most informed decision we can make is to insert a preferential treatment towards sentential types, followed by nouns (assuming that all non-noun daughters are nominalized), followed by adjectives (assuming that participial phrases are used as adjectival phrases).
If neither of these preferences are satisfied, we default to the first item of the most highly occurring tag in the phrase.
We find this heuristic to be adequately performing; the uncertainty introduced, when it is not, is a minimal compromise.

Just like determiner pairs, conjunctions are sometimes coordinated not by a single head, but rather a pair of coordinators; for instance ``zowel.. als'' (\textit{such.. as}), ``respectievelijk ..en'' (\textit{respectively.. and}), etc. 
These pairs are again treated via a special type, \textsc{\_crd}, read as \textit{second item of a coordinator pair}, which looks for a coordinator type to its left (not necessarily adjacent) and collapses with it inheriting its type.

Finally, a major issue arises when a conjunction is unheaded.
This is a by-product of the annotation philosophy, which abstains from analyzing punctuation symbols, pushing them to the top of the parse with a default dependency under the root node.
Then conjunctions coordinated by comma(s) appear as simply an arrangement of sisters with no common head.
This lack of internal structure is not as easy to circumvent as in multi-word units.
Conjunction phrases can not be collapsed into a single item, as their daughters are not  necessarily leaves; they may be phrasal nodes, themselves containing a lot of important structure that cannot be ignored.
A considered solution would be to arbitrarily assign the head role to the first conjunct; this, however, would have the effect of potentially introducing exceedingly high-order functors to describe words further down the DAG, which bare little semblance to their actual phrase-local syntactic functionality.
The hard decision to make then is whether to enforce well-typedness at the sentence-level, or constrain the type scope to the local, conjunct-internal context, favoring corpus-wide uniformity and type clarity.
For the time being, we opt for the latter, leaving these constructions unheaded.
A temporary but minimally pervasive treatment for such analyses is detailed in the paragraph to follow. 
A future solution would be to attempt to push punctuation into the DAG and allow it to partake in or interact with the type assignment, filling in the gaps left by the missing coordinators.

\begin{figure}
    \centering
    \includegraphics[scale=0.29]{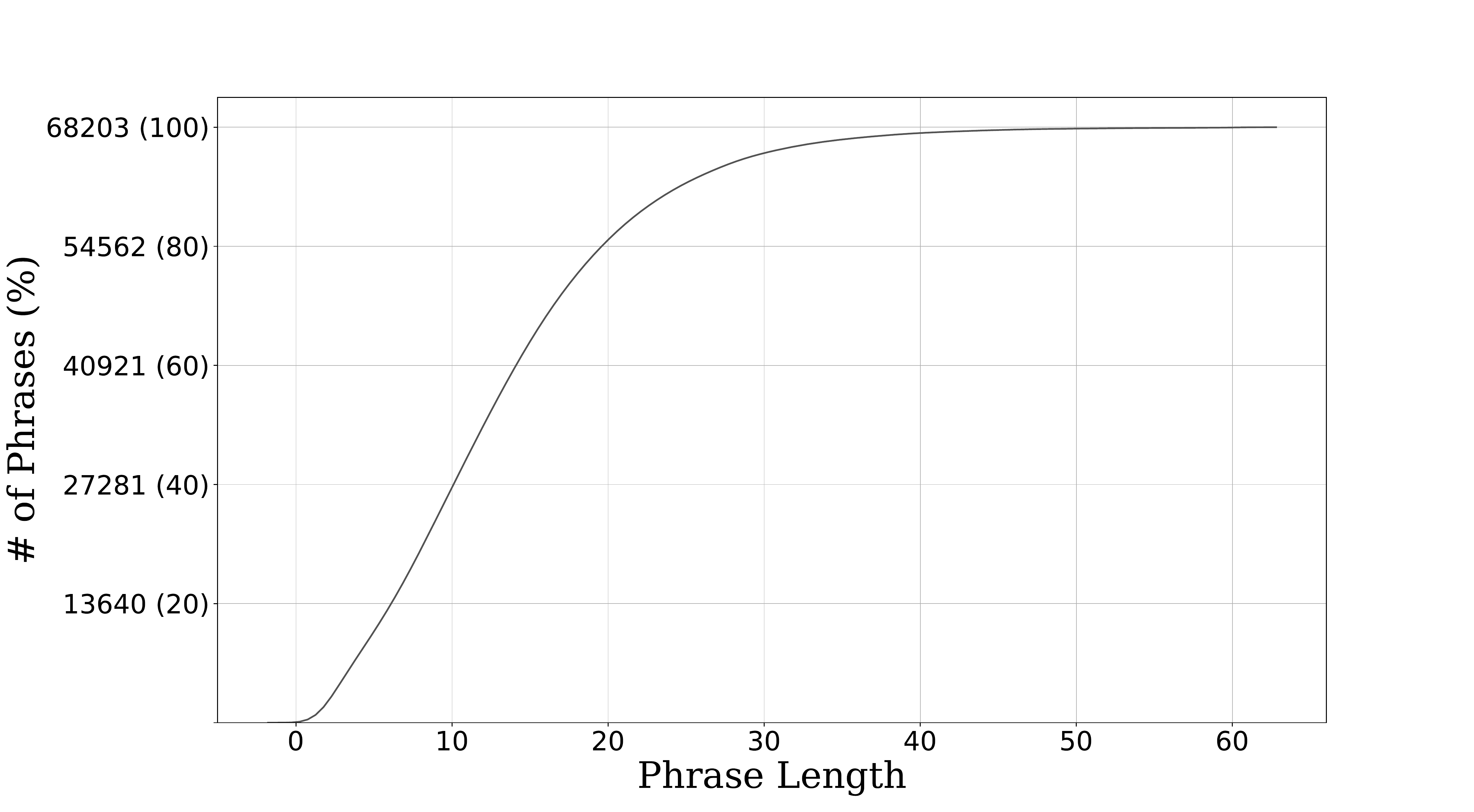}
    \caption[Processed Phrase Sentence Lengths]{Cumulative phrase length in the processed corpus, excluding punctuation.}
    \label{fig:dataset_lens}
\end{figure}

\subparagraph{Discourse Labels and Headless Conjuncts}
Excluding multi-word phrases (for which an adequate solution has been realized), unheaded analyses amount to about 3\% of the overall branchings in the corpus.
These mostly include discourse-level annotations (nuclei, satellites, discourse links and parts), but also the aforementioned conjunctions with no coordinator.
The overall percentage is minor, but cannot be taken lightly; sentences containing at least one such branching occupy a non-negligible part of the treebank, and ignoring them would incur a significant loss of samples for the extraction and its later applications.
Given our inability to perform type inference in such scenarios, a solution that minimizes the sample loss is to instead split each unheaded branch, discarding problematic dependencies and all components above them, and treating each disjoint sub-DAG underneath as a new, independent phrase, rooted at the cut-off daughter node.
This way, we may still utilize proper annotations that are enclosed within unheaded structures under broader phrases.
The partial nature of the resulting DAGs remains in line with the rest of the treebank, as a number of unprocessed samples already portray phrases that do not stand as independent linguistic units.
The result of the splitting process is a new treebank which contains more but smaller samples; Figure~\ref{fig:dataset_lens} depicts the size and length distribution of the altered treebank.
The impact of this transformation is a decrease in the average length of a phrase of about 3 words (also including the removal of punctuations), and an increase in the corpus size by 3\,000 samples.

\paragraph{Nominalization}
No explicit action is taken towards nominalization, with the annotation's tags taken at face value.
This increases the plurality of verb types, as new instances of them will arise from non-standard argument combinations.
Effectively, these new types may be used as an indication for the different kinds of possible noun promotions, which may be performed a posteriori as an informed post-processing step, if deemed necessary.
Majority voting ensures that, even in the case of uneven parts, conjuncts are properly typed, thus eliminating any ambiguity that might have arisen.

\paragraph{Coordinator Types}
Coordinators, when present in a conjunction, are instances of the polymorphic inductive scheme:
\[
C :=  \left \{ C_\textsc{t} \ | \ \forall \ \textsc{t} \in \textsc{Type}   \right \}
\]
where
\[
C_\textsc{t} := \textsc{t} \myrightarrow{cnj} \textsc{t} \ | \ \textsc{t} \myrightarrow{cnj} C_\textsc{t}
\]

Intuitively, the scheme says that the type of coordinators is parametric to the type of their arguments (i.e. there exists a scheme of coordinators for every type\footnote{In practice, not all types can be conjoined, e.g. there is no impredicativity (conjunction of coordinators)}), but also to the number of arguments they are applied on (i.e. for each coordinator scheme, there exists a different instance that specifies the number of conjuncts). 

The multitude of potential types occurring under conjunction, in combination with the great variety in the number of possibly conjoined sisters yields an alarming number of different instantiations of the above scheme, dramatically increasing the ambiguity of coordinator types.
To remedy this, we replace the induction by the meta-type:
\[
C_\textsc{t} := \star \textsc{t} \myrightarrow{cnj} \textsc{t}
\]
where $\star$ is a meta-logical unary connective denoting a sequence of types $\textsc{t}$ with a minimal length of two.
This change is implemented via a conditional branch in the head-type construction method.
Table~\ref{table:coords} presents the most common instantiations of the polymorphic scheme, which together account for approximately 75\% of the total occurrences of coordinator types.

The absence of strict nominalization implies that (rarely) some conjunction cases will not follow the polymorphic scheme; for example, a nominalized adjective (typed as \textsc{adj}) or a pronoun (\textsc{vnw}) might be conjoined with one or more nouns.
When such situations arise, we are forced default to a mixed conjunction type:
\[
\star \textsc{x}_1 \myrightarrow{cnj} \star \textsc{x}_2 \dots \myrightarrow{cnj} \textsc{y}
\]
where $X$=\{$\textsc{x}_1, \dots $\} a set of distinct types, and \textsc{y} $\in$ $X$ the conjunction type, as inferred by majority voting.

\begin{table}
    \centering
    \newcommand{\ra}[1]{\renewcommand{\arraystretch}{#1}}
    \ra{1.1}
    \newcolumntype{Y}{>{\centering\arraybackslash}X}
    \begin{tabularx}{0.65\linewidth}{YY}
         \textbf{Type} & \textbf{Count}  \\
         \toprule
         $\textsc{np}$ & 12308 \\
         $\textsc{s}_\text{main}$ & 3485 \\
         $\textsc{ap}$ & 1990 \\
         $(\textsc{n} \myrightarrow{invdet} \textsc{np}) \rightarrow \textsc{np}$ & 1373 \\
         $\textsc{pp}$ & 1322 \\
         $\textsc{np} \myrightarrow{su} \textsc{s}_\text{main}$ & 1154 \\
    \end{tabularx}
    \caption[Common Coordinator Types]{Most common types and occurrence counts for the polymorphic coordinator scheme $\star \textsc{t} \myrightarrow{cnj} \textsc{t}$. The fourth type is due to conjunction of noun-phrases with a shared determiner. The sixth type is due to conjunction of sentences with a shared noun-phrase subject.}
    \label{table:coords}
\end{table}

\paragraph{Ellipses}
So far, we have universally regarded secondary edges as embedded clause arguments.
This is, however, not always the case; secondary edges may also signal ``copied'' arguments, i.e. nodes that have the same functional role repeated multiple times under a phrase-wide conjunction.
Such secondary edges need to be distinguished from embedded arguments, as they require different treatment.
This distinction is made easy by the collapse of ``phantom'' and lexical nodes; to tell a copied argument from an embedded argument, it suffices to inspect the set of incoming edges it is associated with.
If they are all identical (modulo the primary/secondary marking), the node is classified as the former. 
Otherwise, if at least one of the dependencies is not in agreement with the rest, it is classified as the latter.

Several different elliptical structures appear in the treebank, each necessitating a different approach.

\subparagraph{Modifier Copying}
If all daughters of a conjunction node are associated with the same modifier, the modifying node is separated from them and linked to the conjunction node itself.
This serves two purposes. 
First, it suppresses the need for a logical treatment of the copying on the type level.
Further, it indirectly enforces the polymorphic nature of the modifying node, even in the case of non-polymorphic conjunctions.
Essentially, even if not all conjuncts have equal types, the majority voting scheme creates the most plausible type at the conjunction level. 
Since the modifier is now applied on the higher level, there is no added ambiguity for its type.

\begin{figure}[t]
    \begin{subfigure}[t]{0.49\textwidth}
        \centering
        \includegraphics[scale=0.46]{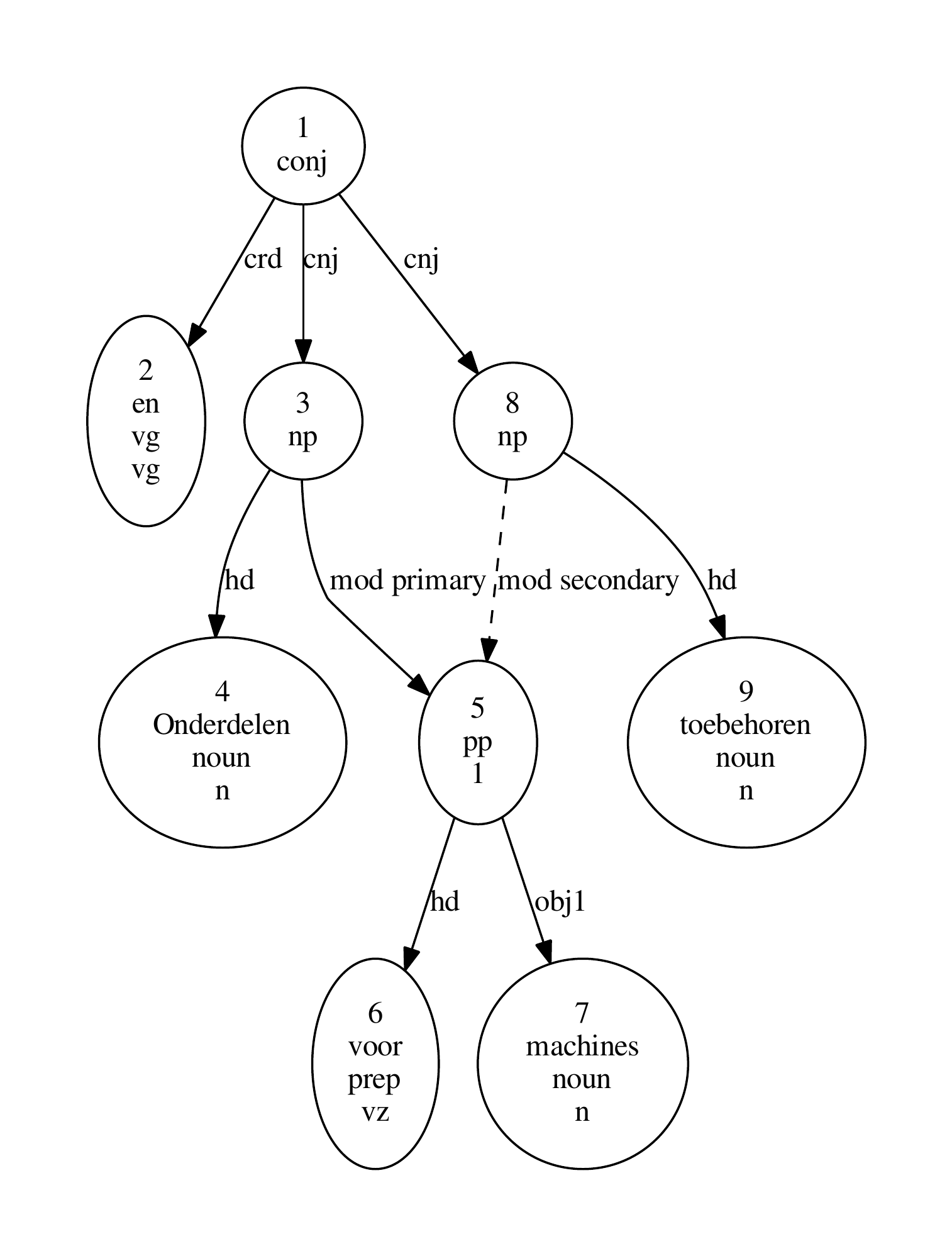}
        \caption{Before reattachment}
    \end{subfigure}
    \begin{subfigure}[t]{0.49\textwidth}
        \centering
        \includegraphics[scale=0.46]{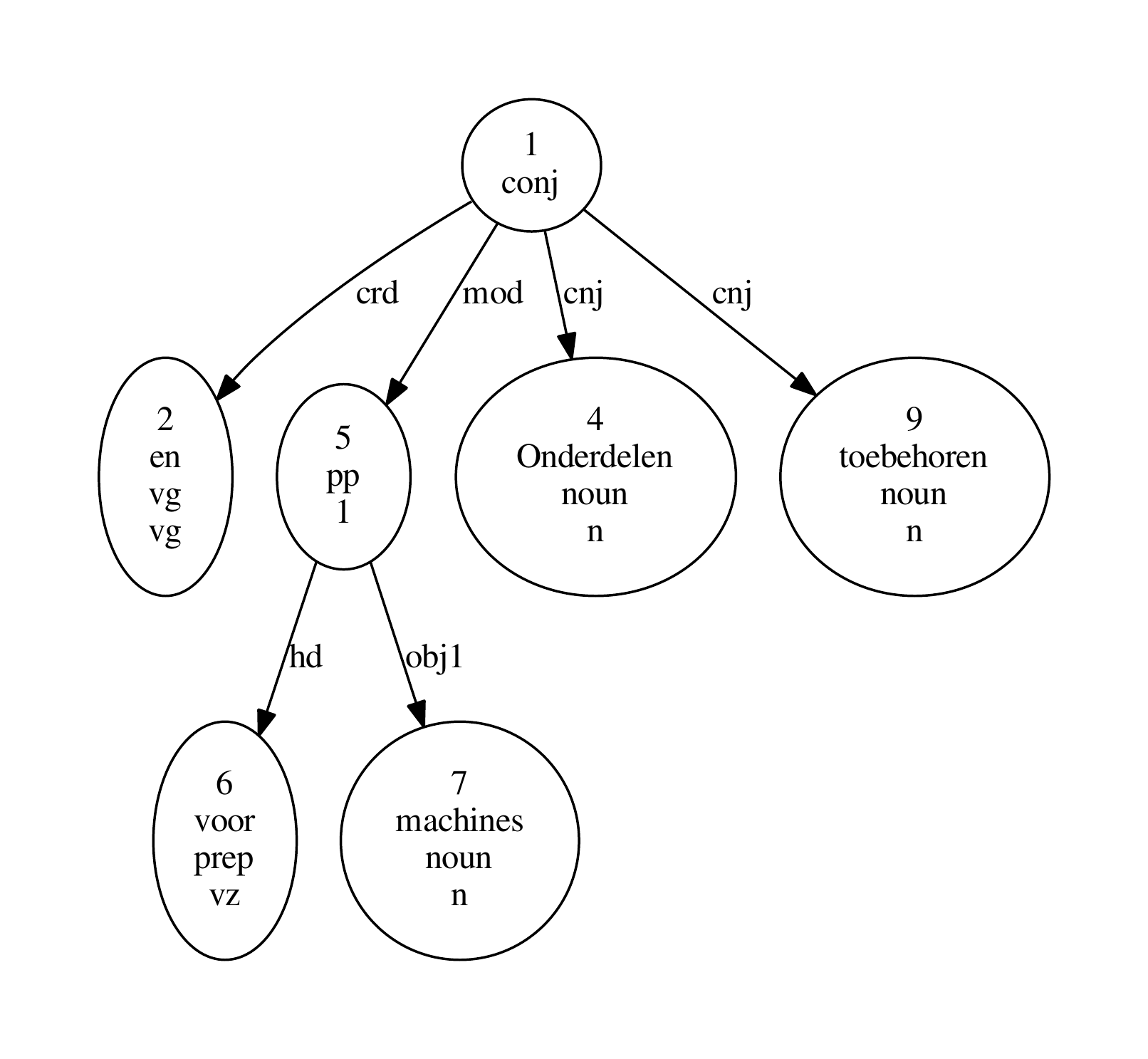}
        \caption{After reattachment}
    \end{subfigure}
    \caption[Conjunction Modifier Reattachment]{Derivation structure before (a) and after (b) reattachment of shared modifiers for the conjunction ``onderdelen en toebehoren voor machines'' (\textit{parts and accessories for machines}) [\texttt{WR-P-E-H-0000000052.p.689.s.1}]. The transformation detaches the prepositional phrase ``voor machines'' from the two conjuncts it modifies, and connects it once as a daughter of the conjunction node.}
\end{figure}

\subparagraph{Argument Copying} 
If a non-modifying argument is applied on more than one descendants of a conjunction, the type inference process gets slightly more involved.
A minimal solution would be to simply allow copying of the missing material via a unary operator such as ILL's \textit{bang} (!).
Although this trivializes the extraction, it significantly increases the grammar's complexity.
The alternative we pursue instead seeks to perform the type inference process at a local level, deriving the conjunct's type as the result of a partial argument application.
The head functor is allowed to consume its sisters' types, minus the ones that are shared across the conjunction.
The incomplete result, a shorter functor, is used to update the local conjunct's type in a reverse-recursive manner.
The updated conjuncts are then used as the instantiating type for the polymorphic scheme used by the coordinator.
After the coordinator has been applied to all daughters, the remainder is a single instance of this partially reduced functor, which is then allowed to consume the shared arguments.

\subparagraph{Head Copying}
Another common pattern is head copying, where each conjunct has its own individual arguments, but shares the head with its sisters.
Again, a bang operator on the functor would resolve this trivially.
Aside from that, ILL provides another direct solution in virtue of the tensor ($\otimes$); a pair of argument sequences could be converted into a sequence of pairs, to be then consumed by the functor.
Rather than enhancing the logic with the pair type and altering the functor type, we may instead utilize an equivalent implicational type.
That is:
\[
A_1, \dots, A_N \vdash (A_1 \rightarrow \dots \rightarrow A_N \rightarrow C) \rightarrow C
\]
which states that from a sequence of assumptions $A_1$,\dots $A_N$, one can derive a higher-order function that accepts a curried function from the sequence to $\Gamma$ to produce a $\Gamma$.
This higher-order proposition at the right hand side of the turnstile is now the instantiating type for the polymorphic coordinator.

To drive the point across, Figure~\ref{fig:head_crd} presents an abstract derivation, with shorthands $Y \equiv A \rightarrow B \rightarrow C$ and $X \equiv Y \rightarrow C$.
A concrete example of the symbolic scheme would be a sentence such as ``logicians like proofs and linguists derivations'', with type assignments: 
\[
\begin{array}{cc}
\text{Word} & \text{Type} \\
\hline
\text{logicians} & A \equiv \textsc{np} \\
\text{proofs} & B \equiv \textsc{np} \\
\text{linguists} & A \\
\text{derivations} & B \\
\text{like} & \quad Y \equiv \textsc{np}\myrightarrow{su}\textsc{np}\myrightarrow{obj1}\textsc{s}_\text{main} \\
\text{and} & \quad \star(Y\rightarrow \textsc{s}_\text{main})\myrightarrow{cnj}Y\rightarrow \textsc{s}_\text{main}

\end{array}
\]

\begin{figure}
    \centering
    \scriptsize
    \[
    \infer[\rightarrow E]{A, Y, B, \star X \rightarrow X, A, B \vdash C}
    {
       \infer[id]{Y \vdash Y}{}
       &
       \infer[\rightarrow E]{A, B, \star X \rightarrow X, A, B \vdash Y \rightarrow C}{
            \infer[\rightarrow I]{A, B \vdash Y \rightarrow C}{
                \infer[\rightarrow E]{A, B, A \rightarrow B \rightarrow C \vdash C}{\dots}
            }
            &
            \infer[\rightarrow E]{A, B, \star X \rightarrow X \vdash X \rightarrow X}{
                \infer[\rightarrow I]{A, B \vdash Y \rightarrow C}{
                    \infer[\rightarrow E]{A, B, A \rightarrow B \rightarrow C \vdash C}{\dots}
                }
                &
                \infer[id]{\star X \rightarrow X \vdash X \rightarrow X \rightarrow X}{}
            }
       }
    }
    \]
    \caption[Head Coordination Schema]{Abstract schema for the derivation of a sentence under head coordination.}
    \label{fig:head_crd}
\end{figure}

\subparagraph{Mixture} 
The last kind of ellipses we need to consider are those where both the head and one or more arguments are shared across the conjuncts, as in the case of a shared phrasal verb.
The treatment is essentially a combination of the two prior cases. 
The type schema remains identical to the simple head coordination, except for the fact that the argument sequence $A_1$,\dots $A_N$ now consists only of the arguments that are uniquely instantiated per daughter, and $\Gamma$ is no longer a simple type, but rather a function type from the common arguments to the conjunction type, as in the simple argument sharing case.

\subparagraph{Non-Polymorphic Ellipses}
Non-polymorphic ellipses occur when there is shared material between two or more conjuncts, but their internal structure does not match, thus rendering our designed solutions unusable.
In total, 400 such annotations occur over the corpus; they are either annotation discrepancies or edge cases necessitating unique modeling. 
A temporary solution is in the works, inspired by ILL's \textit{with} connective (\&).
Its applicability and utility can be evaluated only after a manual inspection of each of these cases, and a correction of those that are indeed erroneous, and is left as future work.

\paragraph{Single Daughter Non-Terminals}
The displacements and detachments performed by the transformations described often result in non-terminal nodes with a single daughter. As the nodes may potentially carry different category tags to their daughters (by-product of some now obsolete phrasal promotion), this can result in inconsistencies among identical or very similar structures. To void this, we collapse non-terminal nodes with a single outgoing edge in an iterative, bottom-up manner, with the resulting node inherits its attributes from its immediate descendant.

\section{Implementation Notes}
Although a full technical overview of the extraction escapes the purposes of this document, this section seeks to provide some key insights on the implementation, in order to make future use and modification easier for any interested party.
The implementation language is Python 3.6; although this choice comes at a cost in elegance compared to a typed language that would naturally accommodate the type system, it is driven by the need for homogeneity and compatibility with the extraction's output's later uses, which will involve statistical learning, for which Python is adopted as the go-to choice.
The extensive number of cases requiring unique treatment demand an iterative approach to the problem, with code parts altered and added as the need arises.
This constantly evolving nature of the codebase makes it hard to design a static software structure that is guaranteed to capture any future grammar alteration or unforeseen edge case.
As such, this section is more of a snapshot of the design at the time of the writing and its overall principles rather than a conclusive technical report.
The extraction code is publicly available at \url{https://github.com/konstantinosKokos/Lassy-TLG-Extraction} and the latest commit at the time of writing is 3d10be8646dfa08d7ead9589297bc264ddd298bb.

\paragraph{System Decomposition}
On the most abstract level, the extraction algorithm may be viewed as a parameterizable map that accepts as input a DAG representing the derivation of a sentence in Alpino-style format as its input, applies a number of modifications on it producing one or more DAGs as an intermediate step, which are then decorated with a type for each of their nodes and returned as the output.

The system consists of two core, interdependent components --- the type component and the extraction component, each enclosed within a corresponding file.
The first implements the particulars of the extracted type system; it defines the classes that implement the type variations present in the grammar, their inductive construction and their means of interaction. 
The second revolves around the treebank transformations and the extraction algorithm itself, as well as providing a wrapper for managing a dataset (i.e. a collection of samples) and a minimal visualization tool.

\subsection{Type System}
The type system is implemented by a hierarchy of classes, each class corresponding to a family of types exhibiting common structure.

\paragraph{WordType} The top class is \textit{WordType}, an abstract class inherited by all other classes which defines the minimal attributes and functions that all other classes must implement.
A listing of class functions shared between all types is given below.
The first is \textit{equality}, a binary function that compares two types based on both their structure and the singular elements they are composed of and returns True if they are in agreement, or False otherwise.
Equality is implemented as a non-commutative function; a type $\textsc{t}_1$ might be left-equal to a type $\textsc{t}_2$ if the latter is a subclass of the former with identical common attributes, but this does not imply the inverse.
The second is \textit{order}, a function that returns a non-negative integer expressing the order of the functor that a type represents.
\textit{Hashing} is a utility function that uniquely maps types to integers, allowing the arrangement of types into sets and dictionaries.
Finally, \textit{string} is a utility function that converts types into a unique and visually informative string format that allows types to be directly printed.

\paragraph{AtomicType}
The simplest class of types and the building block for all other types is \textit{AtomicType}.
These are zero-ary functors, i.e. constants, identified by the strings they were constructed from.
The set of AtomicTypes is provided by the co-domain of the type lexicon, which is a user-specified, many-to-one mapping from part-of-speech tags and phrasal categories to Atomic Types, as seen in Table~\ref{table:lex}.

\paragraph{ComplexType}
The \textit{ComplexType} class implements the family of inductively defined functor types $\textsc{t}_\text{A} \to \textsc{t}_\text{B}$, a unary functor from the argument type $\textsc{t}_\text{A}$ into the result type $\textsc{t}_\text{B}$.
Multi-argument functors are folded into their unary representations; that is, a functor from a sequence of arguments $\textsc{a}_1, \dots \textsc{a}_N$ into a result type $\textsc{r}$ may be written as the $\textsc{a}_1 \to \textsc{a}_2 \to \dots \textsc{r}$, where the ordering can be specified by a preference scheme.
Otherwise, commutativity may be re-introduced by overloading the comparison operator, allowing for an equivalence relation between multi-argument functors to be easily established (useful in cases where no argument ordering is forced).

\paragraph{ColoredType}
\textit{ColoredType} inherits \textit{ComplexType} and implements the family of functor types $\textsc{t}_\text{A} \myrightarrow{c} \textsc{t}_\text{B}$, where $\myrightarrow{c}$ is a subtyped variant of the implication arrow (a ``colored'' arrow), identified by the label c.
Colored types refine the notion of function and allow distinguishing between functors that share the same arguments and result but differ in the means of conversion from the argument space to the result space.
Colors are strings, elements of a closed set defined by a mapping from the dependency relations present in the corpus, as seen in Table~\ref{table:colors}.
Similarly to complex types, multi-argument functors can be restructured into their chained unary counterparts.
In this case, an extra opportunity for binarization arises, as the ordering can be informed by the implication labels rather than the argument types, as described in~\ref{subsec:extraction_alg}.

\paragraph{ModalType}
The \textit{ModalType} class implements a special family of types which are decorated with some unary operator $[.]$.
The class lacks a universal semantics interpretation; they are instead defined on a per-instance basis.
An example of this is the \textit{StarType}, a case of \textit{ModalType} where the operator in question is the $\star$, denoting a sequence of at least 2 repetitions of the argument type.
Other usecases could involve intuitionistic (i.e. non-linear) types, or expanding the grammar with structural control modalities for limiting the application of copying/erasing, exchange and associativity.

\paragraph{Other Types}
A few extra classes of types are implemented, which are not yet used by the extraction process.
They are mentioned here for the sake of completeness, and to showcase how drastic alterations to the extraction may potentially be achieved with only minor changes in the code.

\begin{itemize}
    \item \textbf{DirectedComplexType} \\
    Inherits \textit{ComplexType}, but also includes an implication direction, allowing for the construction of implicational types more akin to the Lambek Calculus.
    \item \textbf{DirectedColoredType}
    Inherits both \textit{DirectedComplexType} and \textit{ColoredType}. Can be used to create types that are simultaneously direction- and dependency-aware.
\end{itemize}

\paragraph{Type Transformations}
The hierarchical arrangement of type-implementing classes is of major practical importance. 
Any subclass of types may be trivially transformed to any higher subclass by simple erasure of the extra information.
Essentially, as long as this design principle is respected, the extraction algorithm and the type system may be further expanded upon, producing more sophisticated type structures while still allowing an immediate interpretation into a less refined variant.
In other words, given an extracted grammar, simpler versions of it can be obtained trivially with neither a do-over of the extraction code or even a repeated run of the algorithm over the corpus.

The inverse process, i.e. refining the extraction, is also largely simplified; one could always simply overload the type constructors being called throughout the DAG iteration, adding any extra required arguments, thus adapting the extracted grammar with only minimal changes over the type inference algorithm.

\subsection{Processing}
As mentioned earlier, the development process of the extraction component had to follow an iterative approach that incorporated solutions for particular cases in a gradual fashion, either as exceptions or general rules and transformations depending on their regularity.
Even though this poses a difficulty in pre-defining a concrete, end-to-end design pattern, extra effort was put into the adaptability and extensibility of the code in order to accommodate future additions and changes.
To that end, the extraction component relies on a simple but robust procedure; a decomposition module that accepts an input sample, applies a series of transformations onto it, and finally calls the type inference algorithm on the result.

This yields a number of significant benefits.
First off, the overall transformation is malleable, being the composition of arbitrary independent functions.
Each transformation is designed individually; as long as its output is compatible with the expected input of another, the two may be seamlessly composed with no added complexity required to model their interaction.
New transformations may be inserted, and existing ones removed or altered, adapting the input DAG to the extraction's needs while keeping the type inference algorithm unchanged.
More practically, as the first transformation is the one responsible of converting the input file into an extraction-compatible format, replacing it with another could admit processing of different input structures.
At the same time, if a new grammar is specified, it suffices to change the last transformation, it being the inference algorithm, virtually allowing switching between implemented grammars at a whim.
Additional post-processing steps may be chained after the inference to convert the type-annotated DAGs into MILL proofs, pairs of word and type sequences, or simply lexicons mapping words to empirical distributions over types. 

Further, the overall process is stateless, letting us capitalize on the independence of a sample's output from inputs other than the current one. 
As such, the extraction lends itself nicely to the MapReduce paradigm~\cite{dean2008mapreduce}, enabling the use of tunable multi-threading and parallel I/O processing.
By performing the corpus transformations dynamically on the input data, rather than sequentially processing and storing intermediate results, storage and processing overhead is minimized. 
The per-sample processing scheme, in combination with lazy evaluation and on-demand reading ensures smooth scaling to arbitrarily large datasets, allowing a perfectly seamless transition to Lassy-Large.

Finally, the extraction boasts minimal system requirements but near-perfect utilization of additional computational resources, despite its complex nature; the whole of Lassy-Small can fully processed in under 2 minutes using a commercial grade computer.

\section{Summary}
This chapter presented the means through which our type-logical grammar, as specified in Chapter~\ref{chapter:tlg}, can be experimentally aligned with real-world data via the means of an extraction algorithm.
We reviewed the algorithm that accomplishes this alignment, the difficulties that may arise in such a process and the transformations required to adopt Lassy's dependency annotations into grammar-compatible structures.
The resulting treebank combines the high-quality of the Lassy-Small with a highly refined type-system, and offers itself for many practical and theoretical applications, ranging from supertagging and constituency parsing to semantic analysis and statistical models of the language as a whole.
It is our hope that as the extraction gets further refined, its output will arise as a useful, publicly available linguistic resource that can act as the groundwork for further research endeavours.

\chapter{Supertagging}
\label{chapter:sup}
This chapter is an extended version of~\cite{constructive}, to be presented in the fourth workshop on Representation Learning for NLP (REPL4NLP).

\section{Background}
The extraction algorithm, as described in Chapter~\ref{chapter:extraction}, produces a type-annotated DAG.
Projecting a DAG's yield we obtain a sentence where each word is decorated with its corresponding type, which is the minimal information required to begin a proof-theoretic analysis of the sentence.
Obviously, a system only capable of analyzing sentences that it has already been exposed to is of little practical use.
The question then naturally arises of how to enable analyses for sentences not contained in the original corpus.
The answer is supertagging~\cite{supertagging}; a process that employs statistical learning techniques to probabilistically model the type assignment process.
In what follows, we will inspect supertagging, as initially introduced, gradually expanding the original formulation with additional components that broaden its applicability, alongside the literature that introduced them. 
For each such component, a paragraph will be devoted to motivating our reasons for incorporating such additions with reference to our data and the current problem specification.
Regardless of the particular implementation, the common point lies in the treatment of the type-annotated corpus as a training dataset. 
The corpus is treated as a collection of information-carrying samples, which may be used for inferring the parameters of a statistical model (of varying complexity).
The hypothesis is that, after parameter tuning, the trained model can be general enough to correctly analyze new sentences.

\subsection{Original Formulation}
To set things off, we may define $\mathcal{T}$ as the set of types assigned by the extraction process.
First, recall that each word occurrence is assigned a corresponding type.
However, not all occurrences of the same word are necessarily assigned the same type.
A sensible approach is to then view the extraction's output as a type lexicon $\mathcal L$, a mapping that sends each word of the input vocabulary $\mathcal{V}$ to a probability distribution over types:
\[
\mathcal{L}: \mathcal{V} \to \mathcal{S}^{|\mathcal{T}|}
\]
where $\mathcal{S}^{|\mathcal{T}|}$ refers to the standard $|\mathcal{T}|$-simplex, such that:
\[
\mathcal{L}(w) = \left \{ \frac{o(w, \tau)}{\sum_{t \in \mathcal{T}}{o(w, t)}} \ \forall \ \tau \in \mathcal{T} \right \} 
\]
where $o(w, \tau)$ simply counts the number of times word $w$ has occurred with type $\tau$.

This definition of a type lexicon is completely faithful to its original formulation by Bangalore and Joshi.
Although illuminating as a starting point, it suffers from a number of limitations.
Several of those have already been addressed by prior work, and the next subsections are devoted to explaining how.

\subsection{Unbounded Domain}
Our treatment of the lexicon as a mapping from a prespecified vocabulary inherently fails to address type-assignment for words not present in the training corpus.

\paragraph{Word Embeddings}
At this point, we will need to take a short detour to briefly introduce word vectors.
Word embeddings are dense, finite-dimensional vectorial representations that capture word semantic content. 
They inherit their functionality in virtue of the distributional hypothesis, which states that words with similar meanings exhibit similar use (i.e. they tend to appear under similar contexts).
The contextual surroundings of words may be statistically modelled using large-scale corpora in an unsupervised manner; the resulting distributions are high-dimensional and sparse.
Word vectorization techniques are applied on top of these, producing low-dimensional, dense representations.
This may be accomplished either in a predictive manner (e.g. fit the parametric representations of words so as to predict a missing word given its context) or as simple factorizations of the input (e.g. perform singular-value decomposition on the co-occurrence matrix).
Generally speaking, word vectorization techniques accept large, unstructured textual input and produce a vector space with implicit but rich topological structure. 
Words are assigned vectors within that space, and linguistic notions are inherited in the form of numeric operations, with the prime example being semantic similarity between words materializing as angular distance between vectors.
With the recent advent of neural networks, word vectorization techniques have risen in popularity and efficiency, achieving impressive results in encoding both syntactic and semantic information.

\paragraph{Domain Expansion}
Given that word embedding algorithms require no structured input, they may be trained on corpora scales of magnitude larger than the extraction algorithm, therefore containing many more words.
Let $\mathcal{E}$ be a word embedding system, trained on a corpus with vocabulary $\mathcal{V}'$, where $\mathcal{V}' \supset \mathcal{V}$, that produces vectors in a $d$-dimensional space:
\[
\mathcal{E}: \mathcal{V}' \to \mathbb{R}^d
\]
Then, we may use our lexicon $\mathcal{L}$ to parametrically fit a continuous function $f_\mathcal{L}: \mathbb{R}^d \to \mathcal{S}^{|\mathcal{T}|}$, which by function composition yields a probabilistic lexicon over an expanded domain:
\[
 f_\mathcal{L} \circ \mathcal{E}: \mathcal{V}' \to \mathbb{R}^d
\]

\paragraph{Domain Unbounding}
The above addition allows expanding the supertagger's domain to words not present in the type-annotated corpus.
Language is not a closed construct, however; there still is a possibility of the supertagger encountering a word $w$ that is not present in the expanded corpus.
Rather than backing to default behavior, a further and final expansion of the domain may be achieved by incorporating sub-word information (e.g. morpheme- and character-level content).
In fact, modern vectorization techniques do utilize such information in order to account for languages that feature word generation by compounding or rich but systematic morphology~\cite{fasttext}.
Then, $\mathcal{E}$ becomes a mapping from any word in the language $L$ onto an object of the vector space:
\[
\mathcal{E}: L \to \mathbb{R}^d
\]
and its composition with the continuous type-assignment function is an unbounded domain supertagger:
\[
f_\mathcal{L} \circ \mathcal{E}: L \to \mathcal{S}^{|\mathcal{T}|}
\]

\paragraph{Case In Point}
The practical importance of achieving the maximal possible expansion of the supertagger's domain is substantiated by our corpus' statistics; the extraction output contains little over 73\,000 unique words, whereas the Woordenboek der Nederlandsche Taal (Dictionary of the Dutch Language)\cite{woordenboek} contains approximately 400\,000.

\subsection{Type Disambiguation}
The next thing to address is the ambiguity of the type assignment process.
As pointed out in the original supertagging proposal, the fact that lexical items are assigned different types for each syntactic context they appear in comes at the cost of local ambiguity that needs to be resolved for accurate parsing.

In order to reduce it (or even completely eliminate it), we may inform the type assignment function with surrounding context $\mathcal{C}$:
\[
f_\mathcal{L}: \mathcal{L} \ \otimes \ \mathcal{C} \to \mathcal{S}^{|\mathcal{T}|}
\]
Our notion of context, and the means of representing it, can for now remain abstract as it varies between implementations.
Generally, the syntactic and semantic content of words is largely disambiguated by the surrounding words, so these are valid candidates for inclusion.
Similarly, the proof-theoretic behavior of types imposes certain constraints on the types they may interact with; therefore a formulation that iteratively assigns types while taking prior assignments into consideration is bound to benefit overall performance.

The first approaches to reducing ambiguity involved simple heuristics pertaining to the satisfiability of minimal structural constraints (e.g. the span of the annotation must fall within the sentence boundaries), or modeling joint probabilities over word-type pair spans.
During the later half of the last decade, recurrent networks rose to prominence in supertagging literature, owing to their general adoption as the de-facto computational machinery for sequential processing, as well as the increased availability of high-quality word vectors.
~\cite{xu-etal-2015-ccg} first applied a simple recurrent network~\cite{elman1990finding} for CCG supertagging, whereas~\cite{lewis-etal-2016-lstm} and~\cite{vaswani-etal-2016-supertagging} utilized bi-directional Long Short-Term Networks~\cite{hochreiter1997long}, each successively achieving higher performance benchmarks.
In their general formulation, such systems accept as their input a sequence of vectors representing a sentence, and apply a recurrent function on them, tasked with producing contextually aware feature vectors.
The latter are then used to model each word's class membership over supertags, turning the problem into an instance of sequence labeling, a common case of supervised learning~\cite{graves2012supervised}.

\paragraph{Case In Point}
The fine-grained nature of our type system results in a large number of unique supertags, each characterizing a very specific phrasal structure.
On one hand, this means that words that uniformly assume a single syntactic role are highly likely to be assigned a single type.
On the other hand, words that exist in a broader range of contexts are increasingly ambiguous, even when these contexts are similar.

\begin{figure}
    \centering
    \includegraphics[scale=0.29]{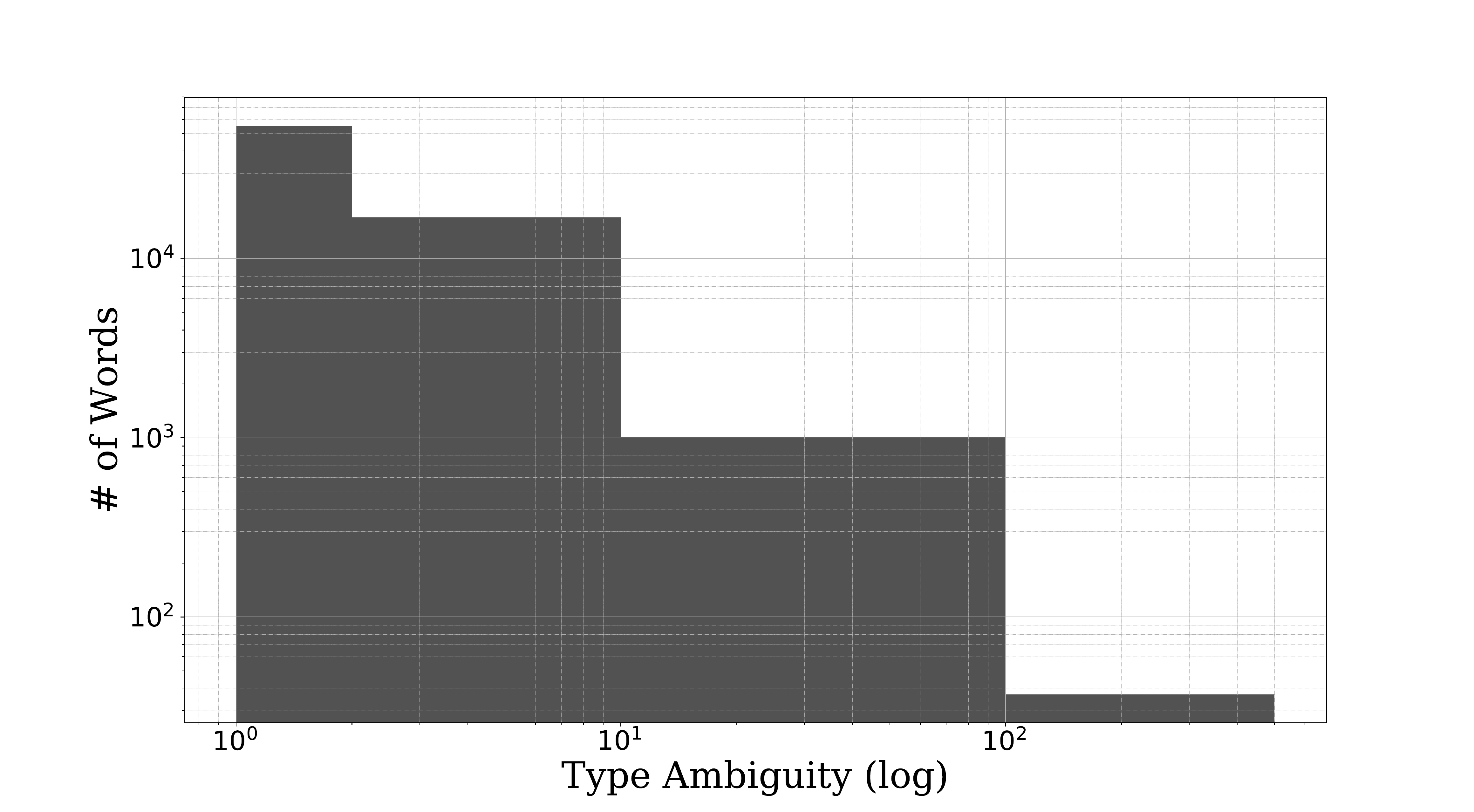}
    \caption[Extracted Type Ambiguity]{Log-log bins of type ambiguity. The horizontal axis bins ranges of type ambiguity, and the vertical axis counts the number of unique lexical items that belong in each bin.}
    \label{fig:type_ambiguity}
\end{figure}

In practical terms, Figure~\ref{fig:type_ambiguity} displays a log-log plot of type ambiguity. 
Words are, on average, assigned 1.8 unique types.
The majority of words (55\,000 and 75\% of the total) are unambiguous throughout the corpus.
The next most likely bin counts 17\,000 words (23\%) that have up to 10 different possible types. 
1\,000 words (1.4\%) are highly ambiguous, with up to 100 different types.
Finally, there are almost 30 lexical items (mostly function words, e.g. coordinators) which are extremely ambiguous, reaching up to 500 potential assignments.

\section{Unbounded Codomain}
The previous section presented a brief overview of supertagging and its progress through the years.
It concluded with the current state of the art, which treats the problem as a case of sequence labeling, modeled by recurrent neural networks.
This perspective is quite natural, and allows supertagging systems to directly inherit the constant and ongoing progress of sequence labeling architectures.
These benefits, however, come at the expense of two significant side-effects, both inherent to classification algorithms.

The first pertains to the issue that class imbalance poses to supervised learners.
Across all categorial grammars, some categories have disproportionately low frequencies of occurrence compared to the more common ones, leading to severe sparsity issues, which are further pronounced the more refined the grammar is.
Under-represented categories are very hard to learn, especially in the context of sequence labeling, where under- and over-sampling are not immediately applicable.
As a result, modern supertagging models are commonly evaluated and tested against only small subset of the full categories present in a grammar, the elements of which have occurrence counts lying above a predetermined treshold.

Moreover, they operate under the strong assumption that the set of types that may be assigned is bounded, and also fully represented by the training corpus, i.e. known in advance.
Practically, the implicit hypothesis is that sentences that would require previously unseen supertags to be correctly analyzed do exist, but are sufficiently rare to be safely ignored.
Such a compromise might be realistic, but it still sets a fixed upper bound on the associated parser's strength.

Jointly, the above concessions have a far-reaching consequence; they place an implicit constraint on the nature of the grammars that may be learned.
Essentially, the grammar must be sufficiently coarse, allocating most of its probability mass on a small number of unique categories.
Grammars enjoying a higher level of analytical sophistication are virtually unusable, as to train an adequate supertagger for them would require prohibitive amounts of annotated data in order to overcome their sparsity.

Our grammar is one such, necessitating an alternative perspective.
This section is devoted to pointing out how a subtle reformulation of the problem bypasses the aforementioned limitations, allowing accurate supertagging with an unbounded codomain.

\subsection{The Language of Types}
Classification over an open set is a difficult problem undergoing active research.
Fortunately, even though our type vocabulary is open, its inhabitants are characterized by an important regularity.
They are all the made out of the same atomic components, the union of the sets of atomic types and $n$-ary logical connectives, each of which is itself closed.
Further, not all sequences of these components constitute valid types. 
Rather, all types are produced by a underlying inductive scheme:
\begin{align}
\textsc{t} ::= \ & \textsc{a} \ | \  \textsc{t}_1 \myrightarrow{\text{d}} \textsc{t}_2
\label{eqn:induction}
\end{align}
where $\textsc{t}$, $\textsc{t}_1$, $\textsc{t}_2$ are types, $\textsc{a}$ is an atomic type and $\myrightarrow{d}$ an implication arrow, decorated with the label $d$.

An examination of the above scheme reveals a simple context-free grammar (CFG).
Given the fact that our connectives are of fixed arity, the polish notation~\cite{hamblin1962translation} can be adopted, abolishing the need for parentheses and reducing the representation length of types, while also encoding their order in a succinct, up-front manner.
Then, the grammar materializes using just two meta-rules for productions:
\begin{align}
\{ (S & \implies A) \  \forall \ A \in \mathcal{A} \}
\\
\{(S & \implies d \ S \ S) \ \forall \ d \in \mathcal{D} \}
\label{eqn:cfg}
\end{align}
where $S$ is the initial symbol and the only non-terminal, $\implies$ is the CFG production arrow, $\mathcal{A}$ is the set of terminals corresponding to atomic types, and $\mathcal{D}$ is the set of terminals corresponding to named implication arrows.
In this light, types are no more than words of the type-forming language.

Of course, the CFG specified above is particular to our type system.
However, any logically grounded categorial grammar is associated with one; consequently, the methodology to be described next is applicable to other formalisms as well. 

\subsection{Supertagging as Sequence Transduction}
This intuition exposes a range of alternative perspectives.
To begin with, it has been shown that neural networks are capable to implicitly acquire context-free languages, both as recognizers and generators~\cite{noPhysics}.
Additionally, state of the art supertagging architectures can already perform the context-sensitive type-assignment process.
It is therefore reasonable to expect that, given enough representational capacity and a robust training process, a network should be able to jointly learn the two tasks; namely, a context-free grammar embedded within a broader labeling task.
A joint acquisition of the two amounts to learning a) how to produce types (including novel ones), and b) which types to produce under different contexts.
Successfully doing both provides the means for an unrestricted codomain supertagger.

A number of data representations and network structures are suitable for meeting the above specifications.
The simplest and most natural one is to simply encode a type as a sequence of atomic symbols $S = s_1 s_2\dots s_n$, where $s_i \in \mathcal{A}\cup\mathcal{D}$.
A sequence of types may then be represented $S_1 \# S_2 \# \dots S_M$, where \# is an auxiliary symbol used to mark type boundaries.

The problem then boils down to learning how a sequence of words can be transduced into a (longer) sequence of unfolded symbols denoting types.
Neural machine translation architectures naturally lend themselves to such a problem specification; this becomes clearer if we imagine the problem as a case of translation operating on word-level input and producing character-level output, where the source language is the natural language trained on and the target is the language mutually defined by the type-logical grammar and the underlying grammar of types.

\begin{figure}[t]
    \centering
    \includegraphics[scale=0.29]{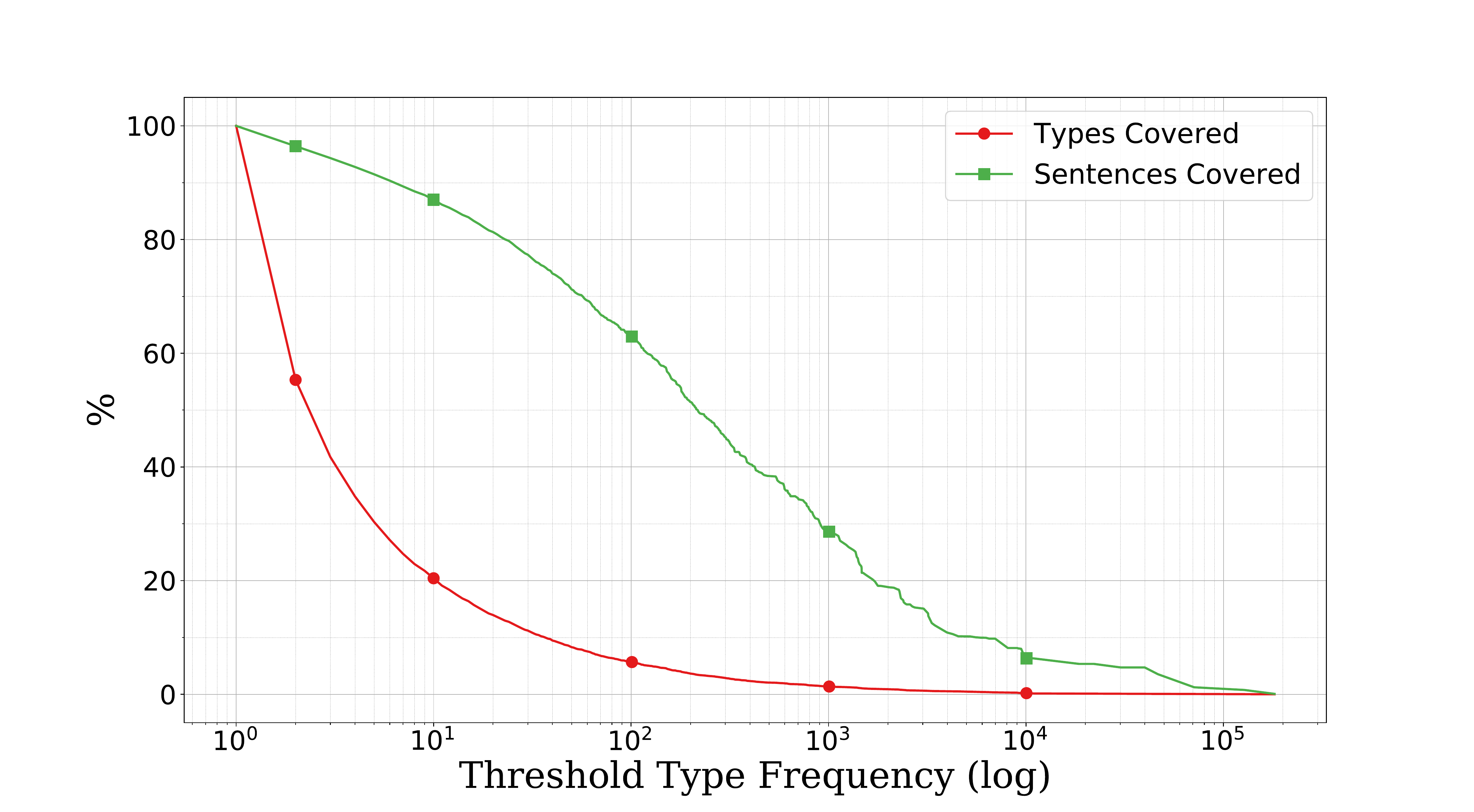}
    \caption[Extracted Type Sparsity]{Cumulative distribution of types and sentences covered as a function of type frequency. The horizontal axis depicts the logarithm of type occurrences in the corpus. The vertical axis depicts the percentage of types (red line)and sentences (green line) kept, if all types below an occurrence rate (or sentences containing them) are discarded.}
    \label{fig:type_sparsity}
\end{figure}

\paragraph{Case In Point}
Our grammar boasts a rich type system, enumerating about 5\,700 unique types.
Its fine-grained nature has the side-effect of a high degree of sparsity; the distribution of types frequencies has a steep exponential curve.
As Figure~\ref{fig:type_sparsity} shows, the vast majority of types (80\%) are rare, i.e. have less than 10 occurrences, and at least one such type exists in a non-negligible portion of the overall sentences (12\%).
Under this regime, recognizing rare types as first-class citizens of the grammar becomes imperative.

Additionally, a significant portion of types (45\%) appear only once throughout the corpus.
Such types would be completely unusable under a predictive setting.
Yet worse, their presence is suggestive of the existence of many more types than those extracted, necessitating an approach that can dynamically construct new types as needed.
With the above in mind, we set out to design a system that can exploit the above observation towards efficient unbounded codomain supertagging.

\section{Implementation}
Our setup imposes a number of design specifications that we must satisfy.
First, given the relatively free word order of the language, the type assignment process needs to be informed by the entirety of the input sentence.
In other words, our system must posses a global receptive field over the input.
Moreover, given the co-dependence between types, owing to their proof-theoretic properties, the output generation process has to be informed by prior outputs, i.e. the architecture needs to follow an auto-regressive formulation.
Finally, the presence of both short- and long-range dependencies, the first dictated by the type-forming grammar and the latter by the sentence-wide grammar, our system would benefit from representations that are global and sensitive to dynamic contexts, rather than local and iterative.

One recent proposal, originally intended for machine translation, perfectly fits the above requirements, namely the Transformer~\cite{vaswani2017attention}, which is an instance of a self-attention network.
To understand why that is the case and provide the necessary background for the rest of the subsection, we will begin by briefly going over the theoretical basis for self-attention networks and the mathematical formulation of the Transformer.

\subsection{Self-Attention}
Self-attention is a recently emerging idea that has gained traction in neural computation.
It is loosely inspired by cognitive attention, i.e. the ability to discriminate between sensory signals that are relevant to the current task and those that are not.
Neural attention was originally intended for use in image processing, but it has been adopted in the natural language processing domain as well, following its successful use in translation~\cite{bahdanau2014neural}.
Neural attention may be thought of as a means of content-based addressing, allowing a network to selectively shift its focus among arbitrary, non-contiguous indices over a set of dimensions of high-order tensors.
The general formulation requires a notion of similarity for the vector space occupied by some input and some context, as established by a matching function $s$:
\[
s: \mathbb{R}^d \otimes \mathbb{R}^d \to \mathbb{R}
\]
where $s(i, c)$  represents the real-valued weight of agreement between the input $i$ and the context $c$ (where $i, c \in \mathbb{R}^d$).
The output of this matching function may be used in a number of possible ways. 
To begin with, it allows the same vector $i$ to enact different roles under different scenarios, enabling context-sensitive processing.
Alternatively, $s(i,c)$ can exert a multiplicative factor over the value of $i$, deciding to what influence the latter will participate in further computation.
Attention then acts as a means of addressing that essentially allows highly precise information retrieval out of arbitrarily large tensors, superseding the need to lossily encode them in a fixed vector space.
Computationally, this is analogous to a program which refers only to variables invoked by its currently run instruction, where the variables are sought ought via partial or approximate reconstructions thereof using autoassociative memory.

If the function $s$ is continuous and differentiable, $s$ may be implemented as a neural function (fixed or parametric), through which gradients can flow during the backward pass, allowing either of its arguments to be optimized (soft attention).
Self-attention is a particular case of soft attention, in which the vectors of both $i$ and $c$ are not provided as direct arguments, but are in themselves productions of the network.
Crucially, the network is then tasked with finding the appropriate transformations such that its inner representations can be efficiently exploited by the matching function $s$.

\subsection{Transformers}
The first applications of attention involved the utilization of attentional interfaces on top of a recurrent network's hidden representations.
Transformers divert from that norm, instead modeling sequential processing purely via self-attention and position-local feed-forward connections.
This approach does away with the computational cost of recurrence (i.e. temporal iteration) and the conceptual constraints it imposes.
In finding an batch-efficient variation of attention, sequential processing may be done in a single step, owing to the lack of temporal dependencies.
For reasons of completeness, this section will briefly cover the main building blocks of a standard Transformer.

\paragraph{Scaled Dot-Product}
Given matrices $Q \in \mathbb{R}^{M\times d}$, $K, V \in \mathbb{R}^{N\times d}$, denoting queries, keys and values respectively, scaled dot-product attention $A$ is modeled as:
\[
A(Q,K,V) = \text{softmax}(\frac{QK^T}{\sqrt{d}})V
\]
The term $\frac{QK^T}{\sqrt{d}}$ is simply the dot-product of the queries against the keys scaled by the inverse of the square root of the vector space dimensionality.
It computes the weighted matching between $M$ queries and $N$ keys (both of dimensionality $d$), in the form of an $M$ by $N$ matrix.
The latter is normalized across its second dimension by application of the softmax function, thus converted into $M$ $N$-simplexes.
The result is matrix multiplied against $V$, producing a $M$ by $d$ matrix, each row of which is a weighted sum of the original rows of $V$.

\paragraph{Multi-Head Attention} A $k$-headed attention block is a neural function consisting of three sets of linear maps $f_{i,q}, f_{i, k}, f_{i, v}: \mathbb{R}^D \to \mathbb{R}^d$ and a linear map $f_o: \mathbb{R}^{kd} \to \mathbb{R}^D$.
It accepts three inputs $\hat{Q} \in \mathbb{R}^{M\times D}$ and $\hat{K}, \hat{V} \in \mathbb{R}^{N\times D}$ and produces an output $\text{MHA}(\hat{Q}, \hat{K}, \hat{V}) \in \mathbb{R}^{M \times D}$ as follows:
\[
\text{MHA}(\hat{Q}, \hat{K}, \hat{V}) = f_o \left ( 
    \text{concat} \left (
        \left [
            A \left (
                f_{i,q}(\hat{Q}),
                f_{i,k}(\hat{K}),
                f_{i,v}(\hat{V})
            \right )
        \
        \forall \ i \in 1,\dots, k
        \right ]
    \right )
\right )
\]
Practically, each $f_{i,x}$ is a parametric projection of its input to a lower-dimensional space.
The scaled dot-product of each of the $k$ projection triplets is computed, and they are concatenated together forming a matrix in $\mathbb{R}^{M\times kd}$.
The result is projected back into the original space by application of $f_o$.

\paragraph{Encoder Block} A transformer encoder block is a neural component that models the composition of a multi-headed attention block and a two-layer, position-local feed-forward network (FFN).
It receives as input a sequence of $M$ vectors, $X \in \mathbb{R}^{M\times D}$, to which it applies the multi-headed attention block producing an intermediate representation $H = \text{MHA}(X, X, X) \in \mathbb{R}^{M \times D}$.
The position-local network is then applied independently on each vector over the $M$ index of $H$.
Optionally, regularization in the form of layer normalization~\cite{lei2016layer} and dropout layers~\cite{hinton2012improving} as well as highway connections~\cite{srivastava2015highway} may be inserted before, between and/or after the encoder block's subcomponents.

\paragraph{Decoder Block} A transformer decoder block differs from the encoder block in two ways: it accepts two inputs $X \in \mathbb{R}^{M \times D}$ and $Y \in \mathbb{R}^{N\times D}$, and it contains two multi-head attention blocks.
The first computes the intra-attention of $Y$ over itself to produce a sequence of factored queries $Q_Y = \text{MHA}(Y, Y, Y)$.
The second computes the attention between the input $X$ and $Q_Y$: $H = \text{MHA}(Q_Y, X, X)$.
Finally, the position-wise transformation is applied on $H$ just as before.

\paragraph{Putting Things Together}
The overall architecture consists of an encoder and a decoder, each being a stack of their respective blocks.
The encoder is first applied over the full input sequence, producing contextually informed vectors over it.
The absence of recurrence necessitates the modulation of the input to distinguish between repetitions of the same item in a sequence; this is accomplished by adding a positional encoding over both input and output embeddings, implemented as a multi-dimensional sinusoidal wave.
The decoder then composes the decoder blocks as follows:
\begin{itemize}
    \item each block's $Y_i$ input is produced by the immediately prior block (where $Y_0$ is the vectorial representation of the output sequence produced up until the current timestep, right shifted once)
    \item each block's $X$ input is the encoder's output
    \item the output of the last decoder block is passed through a standard classifier to produce class labels over each output item
\end{itemize}

\begin{figure}
    \centering
    \begin{tikzpicture}[every text node part/.style={align=center},
                        every node/.style={transform shape},
                        scale=1,
                        miniblock/.style={
                            rectangle, 
                            inner sep=0pt,
                            rounded corners,
                            thick,
                            minimum width=50pt,
                            minimum height=25pt,
                            draw=gray!90},
                        encblock/.style={
                            rectangle, 
                            inner sep=0pt,
                            rounded corners,
                            very thick,
                            minimum width=70pt,
                            minimum height=100pt,
                            fill=gray!20,
                            draw=gray!90},
                        decblock/.style={
                            rectangle, 
                            inner sep=0pt,
                            rounded corners,
                            very thick,
                            minimum width=70pt,
                            minimum height=150pt,
                            fill=gray!20,
                            draw=gray!90},
                        textblock/.style={
                            rectangle,
                            inner sep=0pt,
                            outer sep=7.5pt,
                            minimum width=50pt,
                            minimum height=0pt},
                        dotblock/.style={
                            circle,
                            fill=black,
                            draw=black,
                            minimum size=3pt,
                            inner sep=0pt}
                        ]
        \definecolor{mha}{RGB}{253,184,99};
        \definecolor{ffn}{RGB}{178,171,210};
        \definecolor{linear}{RGB}{166,219,160};
        \definecolor{embedding}{RGB}{255,255,178};
        \node[textblock,align=left] (input) at (0, 0) {Input\\ Embeddings};
        
        \node[encblock] (enc1) at (0,3) {};
        \node[dotblock] (enc1prior) at (0,1.5) {};
        \node[miniblock, fill=mha] 
            (enc1mha) at (0,2.25) {MHA};
        \node[miniblock,fill=ffn] 
            (enc1ffn) at (0,3.75) {FFN};
            
        \node[encblock] (enc2) at (0,8) {};
        \node[dotblock] (enc2prior) at (0,6.5) {};
        \node[miniblock, fill=mha] 
            (enc2mha) at (0,7.25) {MHA};
        \node[miniblock,fill=ffn] 
            (enc2ffn) at (0,8.75) {FFN};
            
        \node[textblock] (enc_interm) at (0, 5.5) {\dots};

        \draw (input) edge[very thick,->] (enc1prior);
        \draw (enc1prior.center) edge[bend right, very thick] (enc1mha);
        \draw (enc1prior.center) edge[bend left, very thick] (enc1mha);
        \draw (enc1prior) edge[very thick] (enc1mha);
        \draw (enc1mha) edge[very thick,->] (enc1ffn);
        
        \draw (enc1ffn) edge[very thick] (enc_interm);
        \draw (enc_interm) edge[very thick,->] (enc2prior);
        \draw (enc2prior.center) edge[bend right, very thick] (enc2mha);
        \draw (enc2prior.center) edge[bend left, very thick] (enc2mha);
        \draw (enc2prior) edge[very thick] (enc2mha);
        \draw (enc2mha) edge[very thick,->] (enc2ffn);

        \node[textblock,align=left] (output) at (4, 0) {Output\\ Embeddings\\ (right-shifted)};
        
        \node[decblock] (dec1) at (4,3.8) {};
        \node[dotblock] (dec1prior) at (4,1.5) {};
        \node[miniblock, fill=mha] 
            (dec1mha) at (4,2.25) {MHA};
        \node[miniblock, fill=mha] 
            (dec1mmha) at (4,3.85) {MHA};
        \node[miniblock, fill=ffn] 
            (dec1ffn) at (4,5.6) {FFN};
        \node[dotblock] (dec1posterior) at (4, 3) {};
    
        \node[decblock] (dec2) at (4,10.8) {};
        \node[dotblock] (dec2prior) at (4,8.5) {};
        \node[miniblock, fill=mha] 
            (dec2mha) at (4,9.25) {MHA};
        \node[miniblock, fill=mha] 
            (dec2mmha) at (4,10.85) {MHA};
        \node[miniblock, fill=ffn] 
            (dec2ffn) at (4,12.6) {FFN};
        \node[dotblock] (dec2posterior) at (4, 10) {};
        
        \draw (output) edge[very thick,->] (dec1prior);
        \draw (dec1prior.center) edge[bend right, very thick] (dec1mha);
        \draw (dec1prior.center) edge[bend left, very thick] (dec1mha);
        \draw (dec1prior) edge[very thick] (dec1mha);
        \draw ($(dec1mha.north) + (-0.25, 0)$) edge[very thick,->] ($(dec1mmha.south) + (-0.25, 0)$);
        \draw (dec1posterior.center) edge[bend right, very thick] (dec1mmha);
        \draw (dec1posterior.center) edge[very thick] (dec1mmha);
        \draw (dec1mmha) edge[very thick,->] (dec1ffn);
        
        \node[textblock] (dec_interm) at (4, 7.3) {\dots};

        \node[miniblock, fill=linear] (classifier) at (8, 12.6) {Classifier};
        \draw (dec1ffn) edge[very thick,->] (dec_interm);
        \draw (dec_interm) edge[very thick] (dec2prior);
        \draw (dec2prior.center) edge[bend right, very thick] (dec2mha);
        \draw (dec2prior.center) edge[bend left, very thick] (dec2mha);
        \draw (dec2prior) edge[very thick] (dec2mha);
        \draw ($(dec2mha.north) + (-0.25, 0)$) edge[very thick,->] ($(dec2mmha.south) + (-0.25, 0)$);
        \draw (dec2posterior.center) edge[bend right, very thick] (dec2mmha);
        \draw (dec2posterior.center) edge[very thick] (dec2mmha);
        \draw (dec2mmha) edge[very thick,->] (dec2ffn);

        \node[textblock] (probabilities) at (8, 10.6) {Output Probabilities};
        \draw (classifier) edge[very thick,->] (probabilities);
        \draw   [decorate,
                very thick, 
                decoration={brace,amplitude=10pt}]
                (-1.5,1) -- (-1.5,10) node[black,midway,xshift=-1.3cm] 
{Encoder};
        \draw   [decorate,
                decoration={brace, amplitude=10pt,mirror},
                yshift=0pt,
                very thick]
                (5.5,1) -- (5.5,13.5) node[black,midway,xshift=1.3cm] {Decoder};
        \draw (dec2ffn) edge[very thick,->] (classifier);
        
        \draw [rounded corners=5pt, very thick,->] (enc2ffn)--(0,10)--(dec2posterior);
        \draw [rounded corners=5pt, very thick,->] (2,10)--(2,3)--(dec1posterior);
        \node[inner sep=0] at (2, 10) {};
        
        \node[miniblock, align=left, fill=embedding] (emb) at (8,0) {\small Embedding \\ Layer};
        \draw (probabilities) edge[->, dotted, very thick] node[right] {$argmax$} (emb);
        \draw (emb) edge[->, dotted, very thick] (output);

    \end{tikzpicture}
    \caption[Transformer Architecture]{Abstract view of the Transformer architecture, with unspecified number of Encoder and Decoder layers. During training, the output embeddings are precomputed. During inference, the embedding of the most probable output class of each timestep is iteratively computed and appended to the previous output embeddings, as depicted by the dotted line.}
    \label{fig:transformer}
\end{figure}
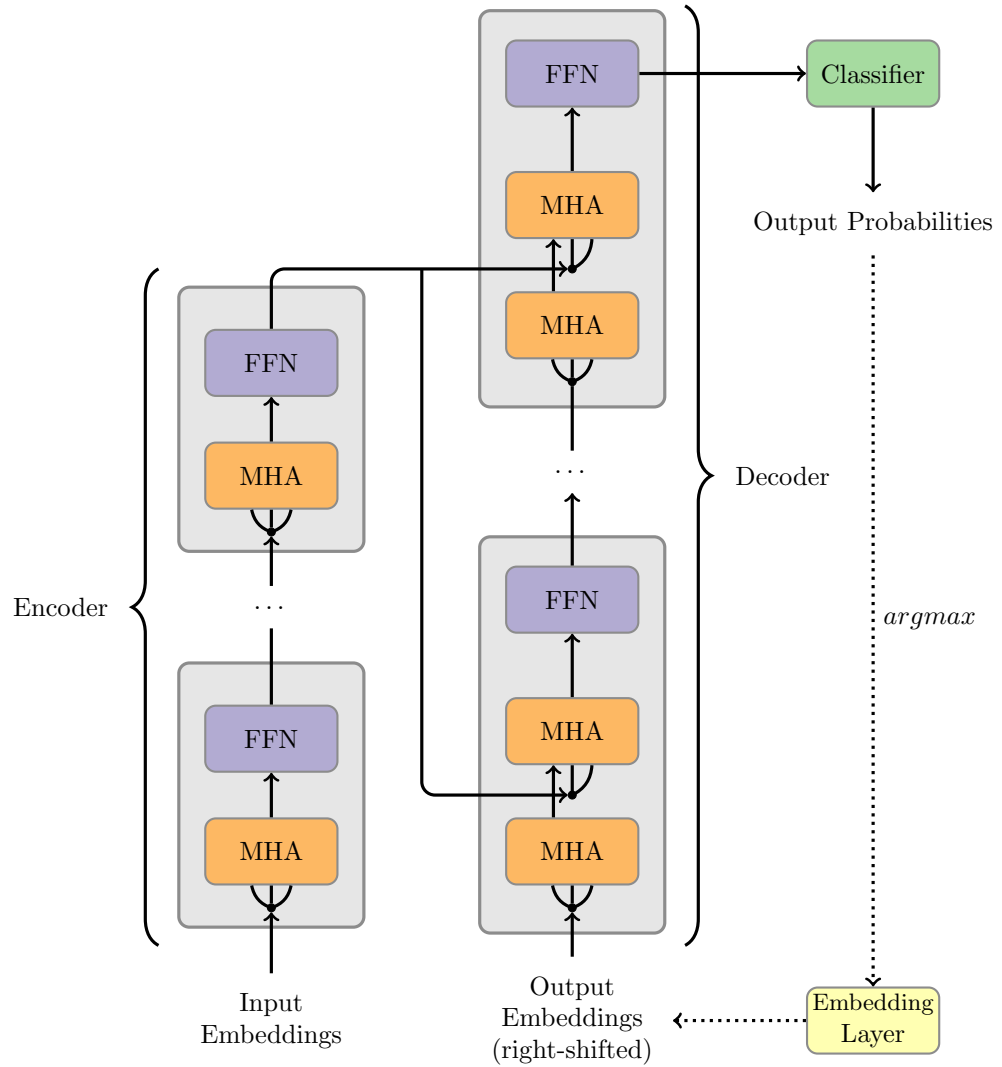

This flow ensures that the final output is informed by the full input sequence but also the preceding output sequence.
It is not bound to equal sequence lengths between input and output, and requires no lossy compression, unlike recurrent encoder/decoder architectures.
By masking the right-shifted output embeddings, the forward pass of the Transformer has minimal computational cost during training time, allowing faster training of deeper models.
During inference, temporal dependency is reintroduced by the need to operate on output embeddings which are not known in advance.
An abstract schematic view of the Transformer is given in Figure~\ref{fig:transformer}.

\subsection{Model}
The supertagging model's architecture follows the standard encoder-decoder paradigm commonly employed by sequence-to-sequence models. 
It accepts a sequence of words as input, and produces a longer sequence of atomic symbols as output.
A high-level overview, together with an example input/output pair, are presented in Figure~\ref{fig:modelandio}.
The source code for our model is publicly available and can be found at \url{https://github.com/konstantinosKokos/Lassy-TLG-Supertagging}.
The custom implementation of the Transformer used by the model can be found at \url{https://github.com/konstantinosKokos/UniversalTransformer}.

\begin{figure}
\centering
\begin{subfigure}[b]{1\textwidth}
\begin{tikzpicture}[every text node part/.style={align=center},
 every node/.style={transform shape},
 scale=0.6,
block/.style={rectangle, inner sep=0pt, minimum width=120pt, minimum height=60pt, rounded corners, ultra thick},
str/.style={rectangle, inner sep=0pt, minimum width=120pt, minimum height=20pt},
arrow/.style={->, ultra thick},
pwise/.style={circle, inner sep=0pt, minimum size=10pt},
smallblock/.style={circle, inner sep=5pt, minimum size=12pt, rounded corners, thick}]

	\node[str] (sentence) at (0, 5) {Input Sentence};		
	\node[str] (symbols) at (7.5, 5) {Output Sequence};

    \definecolor{first}{RGB}{82,82,82}
    \definecolor{enc}{RGB}{253,192,134}
    \definecolor{dec}{RGB}{127,201,127}
    \definecolor{dec2}{RGB}{140,220,140}
    \definecolor{emb}{RGB}{190,174,212}

	\node[block, draw=black, fill=gray!10, draw=gray!130] (elmo) at (0,8) {\textcolor{gray!110}{ELMo}};
	\node[block, draw=black, fill=enc] (te) at (0,12) {\textbf{Encoder}};
	\node[block, draw=black, fill=emb] (se) at (7.5,8) {\textbf{Embedding}};
	\node[block, draw=black, fill=dec2] (td2) at (7.75, 12.25) {};
	\node[block, draw=black, fill=dec] (td) at (7.5,12) {\textbf{Decoder}};
	\node[block, draw=black, fill=emb] (set) at (15,12) {\textbf{Embedding}\\ (transposed)};	
	\node[smallblock, draw=black] (ss) at (15, 8) {$\sigma$};
	\node[smallblock, draw=black] (am) at (12, 8) {$\alpha$};
	\node[str] (out) at (15,5) {Output Probabilities};

	\draw (symbols) edge [arrow, gray!130] node[right] {M symbols} (se);
	\draw  (sentence) edge [arrow, gray!130] node[left] {N words} (elmo);
	\draw  (elmo) edge [arrow, gray!130] node[left] {Sentence Embedding\\ $\mathbb{R} ^ {N \times 1024}$} (te);
	\draw  (se) edge [arrow] node[right] {Symbol Embeddings\\ $\mathbb{R} ^ {M \times 1024}$} (td);
	\draw ($(te.east) + (0, 0.5)$) edge [arrow] node[above] {Encoder Keys\\ $\mathbb{R}^ {N \times 1024}$} ($(td.west) + (0, 0.5)$);
	\draw ($(te.east) + (0, -0.5)$) edge [arrow] node[below] {Encoder Values\\ $\mathbb{R}^ {N \times 1024}$} ($(td.west) + (0, -0.5)$);\
	\draw ($(td.east) + (0.25, 0)$) edge [arrow] node[above] {Decoder Values\\ $\mathbb{R}^ {M \times 1024}$} (set);
	\draw (set) edge [arrow] node[right] {Class Weights} (ss);
	\draw (ss) edge [arrow] (out);
	\draw (set.south) [dotted, very thick] .. controls +(-1,0) and +(-0.5, 0.5) .. (am.north);
	\draw (am.south) [dotted, very thick, ->] .. controls +(0,-0.5) and +(2, ) .. (symbols.east);
\end{tikzpicture}
\caption{The model architecture, where $\sigma$ and $\alpha$ denote the \textit{sigsoftmax} and \textit{argmax} functions respectively, grayed out components indicate non-trainable components and the dotted line depicts the information flow during inference.}
\label{fig:model}
\end{subfigure}
\vspace{10pt}
\begin{subfigure}[b]{1\textwidth}
\begin{minipage}{1\textwidth}
\gll {is (\textit{is})} \quad {er (\textit{there})} \quad {een (\textit{a})} \quad {toepassing (\textit{use})} \quad {voor (\textit{for})} \quad {lineaire (\textit{linear})} \quad {logica (\textit{logic})}\\
$\textsc{np}\myrightarrow{su}\textsc{s}_\text{main}$
\quad
$\textsc{s}_\text{main}\myrightarrow{mod}\textsc{s}_\text{main}$
\quad
\small $\textsc{np}\myrightarrow{det}\textsc{np}$
\quad
\small $\textsc{np}$
\quad
\small $\textsc{np}\myrightarrow{obj1}\textsc{np}\myrightarrow{mod}\textsc{np}$
\quad
\small $\textsc{np}\myrightarrow{mod}\textsc{np}$
\quad
\small $\textsc{np}$\\
\trans $\myrightarrow{su},\textsc{np}, \textsc{s}_\text{main}, \text{\#}, \myrightarrow{mod},\textsc{s}_\text{main}, \textsc{s}_\text{main}, \text{\#},\myrightarrow{det},\textsc{np}, \textsc{np}, \text{\#},\textsc{np},\myrightarrow{obj1},\textsc{np}, \myrightarrow{mod},\textsc{np}, \textsc{np}, \text{\#}\myrightarrow{mod},\textsc{np}, \textsc{np}, \text{\#},\textsc{np},\text{\#}$
\end{minipage}
\caption{Input-output example pair for the sentence ``is er een toepassing voor lineaire logica?'' (\textit{is there a use for linear logic?}). The first two lines present the input sentence and the types that need to be assigned to each word. The third line presents the desired output sequence, with types decomposed to atomic symbol sequences under polish notation, and \# used as a type separator.}
\label{fig:io}
\end{subfigure}
\caption[Supertagging Architecture and I/O Pair]{Supertagging Architecture~(\ref{fig:model}) and an example input-output pair~(\ref{fig:io}).}
\label{fig:modelandio}
\end{figure}

\paragraph{Language Model Pretraining}
Empirical evidence~\cite{D17-1039} suggests that sequence-to-sequence architectures benefit from unsupervised pretraining of their encoder and decoder components as independent language models.
Language models are statistical models that estimate the probability distribution of sequences (e.g. sentences).
They can be used to either rank the probability of a whole sequence, or as generators, for instance predicting a sequence's continuation given a variable-length prefix thereof, by taking the maximum-likelihood estimate of the conditional distribution $
p(w_t | w_{0},\dots,w_{t-1})
$.

Adhering to this, we incorporate a pretrained Dutch ELMo into our encoder~\cite{dutch_elmo}.
ELMo was originally proposed as an architecture for producing deep contextualized word representations~\cite{peters2018deep}, where deep is reference to the use of a stacked, bi-directional LSTM to process the input sequence. 
A weighted combination of the representations of each layer is then utilized as input by downstream task-specific models, where the weighting terms are functions to be learned by the task models.
The variation we are using was trained as a task-agnostic language model on large-scale corpora, and the 1024-dimensional representations it constructs were proven highly adequate for parsing tasks.
The choice of ELMo's output as our sentence-level input carries relays the added strength of utilizing subword-level information; a character-level LSTM participates in the construction of ELMo's lowest-level (context-independent) token representations.

Constructing a decoder-side language model is less of a straightforward decision.
The size of Lassy-Small prohibits pretraining, as this would require splitting into two disjoint subsets (thus reducing the amount of data available to the end-to-end supertagger); this necessitates the use of a different corpus.
Our extraction algorithm is applicable to the silver-standard Lassy-Large, the size of which is certainly appealing for such an endeavour.
However, the quality of its annotations is lacking, reducing the potential benefits from pretraining.
Further, its larger size, in conjunction with the more irregular types (a by-product of the frequency of erroneous annotations) would be a confounding factor in evaluating our model's ability to construct new types.
Considering the above, we refrain from pretraining the decoder in the current setting.

\paragraph{Transformer} Even though ELMo acts as an encoder already, its representations are not necessarily optimal to use as is.
Adapting its representations to the domain is a costly process; ELMo counts a very large number of parameters, making it prone to overfit our dataset.
Instead, we treat it as a static function and process its output with a single-layer Transformer encoder of 3 attention heads.
In practice, since ELMo's parameters are hard-set and not affected by the backward pass, we may precompute the embeddings of our sentences in advance and feed those onto the rest of the network.

The decoding process is accomplished by 2-layer Transformer decoder. 
As the decoder needs to cast its predictions from a broader range of contexts, due to it processing information at a higher granularity scale, we increase its number of attention heads to 8.

We follow the original Transformer formulation in all but one point; we model the position-wise feed-forward transformation that is internal to the encoder and decoder layers as a two-layer, dimensionality preserving network.
We replace the linear rectifier of the intermediate layer with the empirically superior and statistically grounded Gaussian Error Linear Unit~\cite{gelu}:
\[
\text{GELU}(x) = 0.5x 
\left ( 
    1+\text{tanh}
        \left (
            \sqrt{2/\pi}
                \left (
                    x+0.044715x^3
                \right )
        \right )
\right )
\]

Overall, the network is tasked with modeling the probability distribution of the next atomic symbol at timestep $t$, $a_t$, conditional on all previous predictions $a_0,\dots,a_{t-1}$ and the full input sequence $w_0,\dots,w_\tau$, as parameterized by its trainable weights $\theta$:
\[
p_\theta(a_t|a_0,\dots,a_{t-1},w_0,\dots,w_\tau)
\]

\paragraph{Embedding} Since there are no pretrained embeddings for the output tokens, we train an atomic symbol embedding layer alongside the rest of the network.
The Transformer's formulation gives us no say on the dimensionality of the output space, as it has to match that of the input--- 1024.
This is not optimal, as the number of unique output tokens is one order of magnitude lower than the dimensionality, making the scale of the representations redundant.
To recover from this and make further use of the extra parameters, we use the transpose of the embedding matrix as our output classifier. 
Concretely, the embedding matrix is a linear map from $\mathbb{R}^{|\mathcal{A}|}$ to $\mathbb{R}^{1024}$, where $\mathcal{A}$ the set of atomic symbols used by the supertagger.
Hence, it's transpose is a linear map from $\mathbb{R}^{1024}$ to $\mathbb{R}^{|\mathcal{A}|}$, which may be reused to convert the transformer's 1024-dimensional output back into class weights.
These weights are converted into probabilities in the $|\mathcal{A}|$-simplex by application of the sig-softmax function~\cite{kanai2018sigsoftmax}, a softmax variant that enjoys stronger statistical approximation properties:
\[
\text{sigsoftmax}(x_i) = \frac{e^{x_i}\sigma(x_i)}{\sum_j{e^{x_j}\sigma(x_j)}}
\]

\subsection{Digram Encoding}
Predicting type sequences one atomic symbol or connective at a time provides the vocabulary to construct new types, but results in elongated target output sequence lengths. Note that if lexical categories are, on average, made out of $c$ atomic symbols, the overall output length is a constant factor of the sentence length, i.e. there is no change of complexity class with respect to a traditional supertagger.
As a countermeasure, we experiment with {\it digram encoding}, creating new atomic symbols by iteratively applying pairwise merges of  the most frequent intra-type symbol digrams~\cite{bpe}, a practice already shown to improve generalization for translation~\cite{bpe2} and language modeling tasks~\cite{gpt}. 
To evaluate performance, we revert the merges back into their atoms after obtaining the predictions.

With no merges, the model has to construct types and type sequences using only atomic types and connectives.
As more merges are applied, the model gains access to extra short-hands for subsequences within longer types, reducing the target output length, and thus the number of interactions it has to capture.
This, however, comes at the cost of a reduced number of full-type constructions effectively seen during training, while also increasing the number of implicit rules of the type-forming context-free grammar.
If merging is performed to exhaustion, all types are compressed into single symbols corresponding to the indivisible lexical types present in the treebank. 
The model then reduces to a traditional supertagger, never having been exposed to the internal type syntax, and loses the potential to generate new types.

\subsection{Experiments}
\paragraph{Training}
In all described experiments, the model is run on the subset of sample sentences that are at most 20 words long. 
We use a train/val/test split of 80/10/10; it is worth pointing out that the training set contains only $\sim$85\% of the overall unique types, the remainder being present only in the validation and/or test sets.
Training takes place with a batch size of 128, and sentences are padded to the maximum in-batch length. 
Training to convergence takes, on average, eight hours \& 300 epochs for the training set of 45000 sentences on a GTX1080Ti. 
Results are averaged over 5 repetitions.

Accuracy is reported on the type-level; that is, during evaluation, we predict atomic symbol sequences, then collapse subtype sequences into full types and compare the result against the ground truth. 
Notably, a single mistake within a type is counted as a completely wrong type. 

For the training algorithm, we adopt the training scheme originally proposed by Vaswani et al~\cite{vaswani2017attention}, which is unique in two ways.
First, rather than using standard cross-entropy as the objective function, it instead optimizes Kullback-Leibler divergence, a measure of distance between probability distribution.
The distributions compared are the network's predictions and an artificial distribution of the ground truth.
The latter is constructed by subtracting a fixed percentage of the probability mass from the true label, which is then uniformly spread across the remaining labels.
This practically forces the network to be less certain of its predictions, increasing its generalization capacity.
Moreover, it utilizes an adaptive learning rate, linearly increasing over a number of training iterations, then exponentially decaying until termination.
We set the amount of redistributed probability mass to 20\%, double that of the original proposal; a change that is crucial in order to discourage the network from memoizing common type patterns.
We apply dropout in between all network connections, at a rate of 20\%.

\paragraph{Results}
Our experiments involve a fully constructive model employing no merges ($\text{M}_0$), a fully merged one i.e. a traditional supertagger, ($\text{M}_\infty$), and three in-between models trained with 50, 100 and 200 merges ($\text{M}_{50}$, $\text{M}_{100}$ and $\text{M}_{200}$ respectively).
Table~\ref{table:numbers} displays the models' accuracy. 
In addition to the overall accuracy, accuracy over different bins of type frequencies is displayed, as measured in the training data: unseen, rare (1-10), medium (10-100) and high-frequency ($>$ 100) types.

\begin{table}
\centering
\newcommand{\ra}[1]{\renewcommand{\arraystretch}{#1}}
\ra{1.1}
\hspace{-10pt}
\begin{tabularx}{0.49\textwidth}{@{}Xsssss@{}}
{} &  \multicolumn{5}{c}{\centering Type Accuracy}\\
\cmidrule{2-6}
\multicolumn{1}{l}{}  &  \small Overall & \small Unseen & \small  Freq & \small Freq & \small Freq \\
\multicolumn{1}{l}{Model}  &  {} & \small Types & \small  1-10 & \small 10-100 & \small $>$100 \\
\cmidrule[0.001em]{1-6}
\centering $\text{M}_{0}$  & \textbf{ 88.05} & \textbf{19.2} & \textbf{45.68} & \textbf{65.62} & 89.93\\
\cmidrule[0.001em]{1-6}
\centering $\text{M}_{50}$  & 88.03 & 15.97 & 43.69 & 64.33 & \textbf{90.01}\\
\centering $\text{M}_{100}$ & 87.87 & 15.02 & 41.61 & 63.71 & 89.9 \\
\centering $\text{M}_{200}$ & 87.54 & 11.7 & 39.56 & 62.4 & 89.64\\
\cmidrule[0.001em]{1-6}
\centering $\text{M}_{\infty}$ & 87.2 & - & 23.91 & 59.03 & 89.89\\
\end{tabularx}
\caption[Supertagger Performance]{Model performance at different merge scales, with respect to training set type frequencies. $\text{M}_i$ denotes the model at $i$ merges, where $\text{M}_\infty$ means the fully merged model. For the fully merged model there is a 1 to 1 correspondence between input words and output types, so we do away with the separation symbol.}
\label{table:numbers}
\end{table}

Table~\ref{table:numbers} shows that all constructive models perform overall better than $\text{M}_{\infty}$, owing to a consistent increase in their accuracy over unseen, rare, and mid-frequency types.
This suggests significant benefits to using a representation that is aware of the type syntax.
Additionally, the gains are greater the more transparent the view of the type syntax is, i.e. the fewer the merges.
The merge-free model $\text{M}_0$ outperforms all other constructive models across all but the most frequent type bins, reaching an overall accuracy of 88.05\% and an unseen category accuracy of 19.2\%.

What is also interesting to examine is the models' ``imaginative'' precision, i.e.,  how often do they generate new types to analyze a given input sentence, and, when they do, how often are they right (Table~\ref{table:imagination}). 
Although all constructive models are eager to produce types never seen during training, they do so to a reasonable extent. 
Similar to their accuracy, an upwards trend is also seen in their precision, with $\text{M}_0$ getting the largest percentage of generated types correct. 

Together, the results indicate that the type-syntax is not only learnable, but also a representational resource that can be utilized to tangibly improve a supertagger's generalization and overall performance.

\begin{table}
\centering
\noindent
\newcommand{\ra}[1]{\renewcommand{\arraystretch}{#1}}
\ra{1.1}
\begin{tabularx}{0.49\textwidth}{@{}Xsss@{}}
\multicolumn{1}{l}{Model}  &  New Types & Unique & Correct \small (\%)\\
\multicolumn{1}{l}{}  &  Generated  &  &  \\
\cmidrule[0.01em]{1-4}
\centering $\text{M}_{0}$  & 213.6 & 199.2 &  \textbf{44.39} \small (\textbf{20.88}) \\
\centering $\text{M}_{50}$  & 186.6 & 174.2 & 37.89 \small (20.3) \\
\centering $\text{M}_{100}$  & 187.8 & 173.4 & 34.31 \small (18.27) \\
\centering $\text{M}_{200}$  & 190.4 & 178.8 & 27.46 \small (14.42) \\
\end{tabularx}
\caption[Supertagger Unseen Type Precision]{Repetition-averaged unseen type generation and precision.}
\label{table:imagination}
\end{table}

\paragraph{Baselines}
In order to evaluate our model's performance in comparison to established supertagging practices, we experimented with RNN-based encoder-decoder architectures. 
We tried training a single-layer BiGRU encoder over the ELMo representations, connected to a single-layer GRU decoder, following~\cite{gru}; the model took significantly longer to train and yielded far poorer results.
We hypothesize that the encoder's fixed length representation is unable to efficiently capture all of the information required for decoding a full sequence of atomic symbols, inhibiting learning.

As an alternative, we tried a separable LSTM decoder operating individually on the encoder's representations of each word. 
Even though this model was faster to train and performed marginally better compared to the previous attempt, it still showed no capacity for generalization over rare types. 
This is unsurprising, as this approach assumes that the decoding task can be decomposed at the type-level; crucially, the separable decoder's prediction over a word cannot be informed by its predictions spanning other words, an information flow that evidently facilitates learning and generalization.

\subsection{Analysis}
\paragraph{Type Syntax}
To assess the models' acquired grasp of the type syntax, we inspect type predictions in isolation.
Across all merge scales and consistently over all trained models, all produced types (including unseen ones) are \textit{well-formed}, i.e. they are indeed words of the type-forming grammar.
Further, the types constructed perfectly follow along our implicit notational conventions such as the obliqueness hierarchy.

Even more interestingly, for models trained on non-zero merges it is often the case that a type is put together using the correct atomic elements that together constitute a merged symbol, rather than the merged shorthand trained on.
Judging from the above, it is apparent that the model gains a functionally complete understanding of the type-forming grammar's syntax, i.e. the means through which atomic symbols interact to produce types.

\paragraph{Sentence Syntax}
\begin{figure}
\centering
\begin{minipage}{1\textwidth}

\gll {in (\textit{to})}  {hoeverre (\textit{what-degree})} \quad {zal (\textit{will})} \quad {het (\textit{the})} \quad {rapport (\textit{report})} \quad {dan (\textit{then})} \quad {nog (\textit{still})} \quad {een (\textit{a})} \quad {rol (\textit{role})} \quad {spelen (\textit{play})}\\ $\textsc{\textbf{bw}}\myrightarrow{\textbf{obj1}}\textbf{((}\textsc{\textbf{inf}}\myrightarrow{\textbf{mod}}\textsc{\textbf{inf}}\textbf{)}\rightarrow\textsc{\textbf{sv1}}\textbf{)}\myrightarrow{\textbf{body}}\textsc{\textbf{whq}}$
\quad 
\small $\textsc{bw}$
\quad
\small $\textsc{inf}\myrightarrow{vc}\textsc{np}\myrightarrow{su}\textsc{sv1}$
\quad
\small $\textsc{n}\myrightarrow{det}\textsc{np}$
\quad
\small $\textsc{n}$
\quad
\small $\textsc{inf}\myrightarrow{mod}\textsc{inf}$
\quad
\small $\textsc{inf}\myrightarrow{mod}\textsc{inf}$
\quad
\small $\textsc{n}\myrightarrow{det}\textsc{np}$
\quad
\small $\textsc{n}$
\quad
\small $\textsc{np}\myrightarrow{obj1}\textsc{inf}$\\
\end{minipage}
\caption[Unseen Type Example]{Type assignments for the correctly analyzed wh-question ``in hoeverre zal het rapport dan nog een rol spelen'' (\textit{to what extent will the report still play a role}) involving a particular instance of \textit{pied-piping}. The type of ``in'' was never seen during training; it consumes an adverb as its prepositional object, to then provide a third-order type that turns a verb-initial clause with a missing infinitive modifier into a wh-question. Such constructions are a common source of errors for supertaggers, as different instantiations require unique category assignments.}
\label{fig:electricity}
\end{figure}

Beyond the spectrum of single types, we examine type assignments in context.

We first note a surprising ability to correctly analyze syntactically complex constructions requiring higher-order reasoning, even in the presence of unseen types.
An example of such an analysis is shown in Fig~\ref{fig:electricity}.

For erroneous analyses, we observe a strong tendency towards self-consistency.
In cases where a type construction is wrong, types that interact with that type (as either arguments or functors) tend to also follow along with the mistake.
On one hand, this cascading behavior has the effect of increasing error rates as soon as a single error has been made.
On the other hand, however, this is a sign of an implicitly acquired notion of phrase-wide well-typedness, and exemplifies the learned long-range interdependencies between types through the decoder's auto-regressive formulation.
On a related note, we recognize the most frequent error type as misconstruction of conjunction schemes. 
This was, to a degree, expected, as coordinators display an extreme level of lexical ambiguity, owing to our extracted grammar's massive type vocabulary. 

\begin{figure}[t]
    \centering
    \hspace{-70pt}
    \includegraphics[scale=0.29]{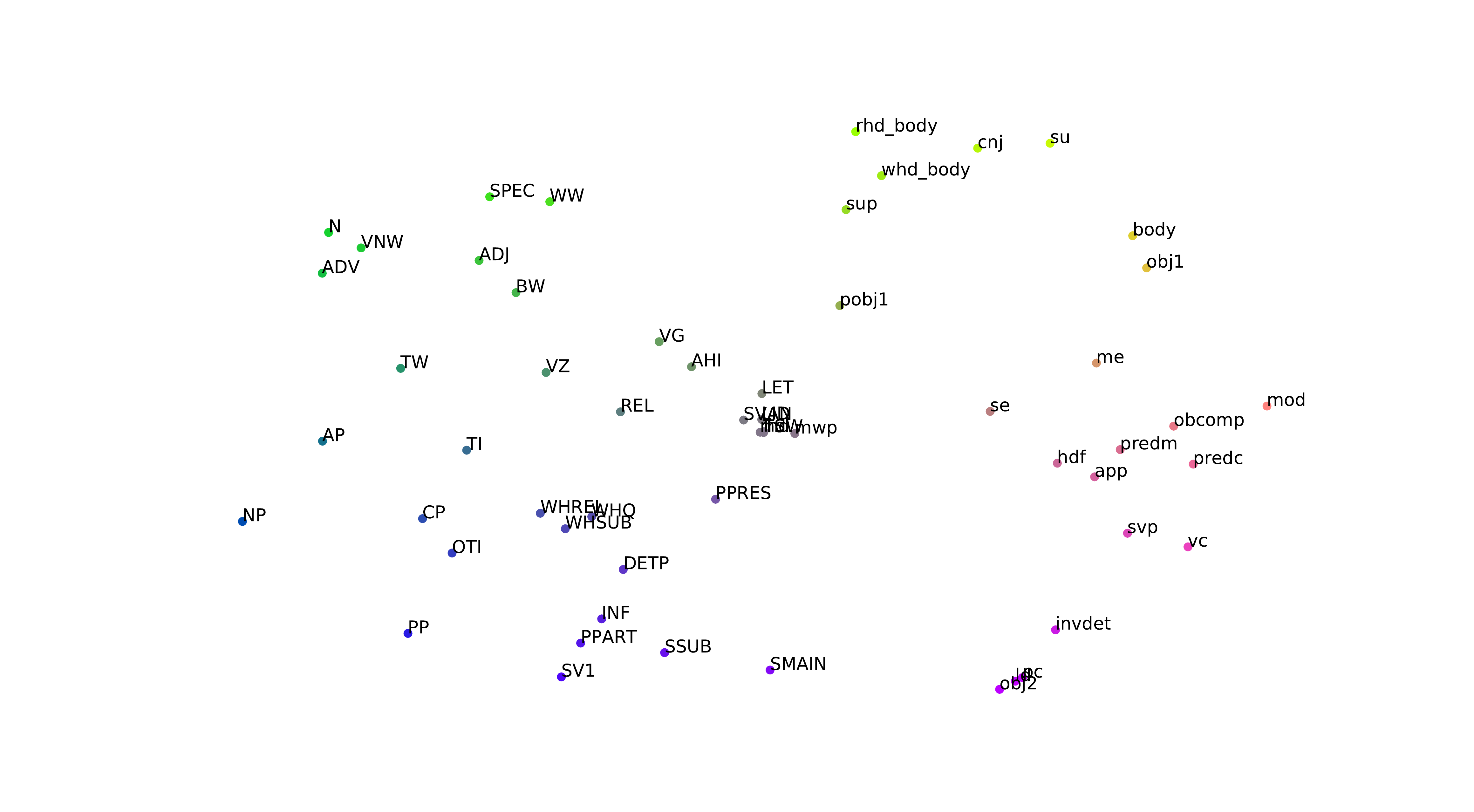}
    \caption[tSNE of Atomic Symbol Embeddings]{2-d tSNE projection of the embedding space. Two major clusters are formed by binary connectives (lowercase, east) and atomic types (uppercase, west). Smaller clusters are also noticeable between sentential types, indirect verb arguments, secondary clauses, complementizers, preliminary arguments, phrasal types etc. Refer to Tables~\ref{table:lex}, \ref{table:colors} for a legend.}
    \label{fig:symbol_tsne}
\end{figure}

\paragraph{Output Embeddings}
Finally, we draw  attention to a thus far ignored aspect of the architecture. 
Our network trains not only the encoder-decoder stacks, but also an embedding layer of atomic symbols.
We can extract this layer's outputs to generate vectorial representations of atomic types and binary connectives, which essentially are high-dimensional character-level embeddings of the type language.
Figure~\ref{fig:symbol_tsne} displays a 2-dimensional tSNE~\cite{maaten2008visualizing} reconstruction of the embedding space, where some degree of structure is immediately apparent.
Considering that dense supertag representations have been shown to benefit parsing~\cite{kasai-etal-2017-tag}, our atomic symbol embeddings may be further utilized by downstream tasks, as a highly refined source of type-level information.

\section{Summary} 
This chapter presented supertagging and its evolution during the last two decades.
We saw how supertagging is a crucial component to categorial parsing, and the means through which it may become more generally applicable and accurate.
Our novel contribution lies in the reformulation of the supertagging task, from sequence labeling (with labels being the type vocabulary) to sequence transduction (with outputs being the elements primitive to type construction).
Such a reformulation, paired with the strong idea of self-attention and an encoder-decoder architecture allowed us to lift the closed-world assumption, yielding a highly general architecture capable of dynamically constructing types beyond a prespecified lexicon.
This work shows that the type-forming syntax internal to categorial grammars can be fully acquired by neural networks, bypassing the limiting factor of type sparsity, and paving the way for exploring the realistic applications of richer type systems in natural language parsing.

\chapter{Parsing}
\label{chapter:parsing}

This chapter presents presents preliminary results on parsing using the grammar specification of Chapter~\ref{chapter:tlg} and the extraction results of Chapter~\ref{chapter:extraction}.
A version of this manuscript has been accepted for presentation at the third workshop on Semantic Spaces at the Intersection of NLP, Physics, and Cognitive Science (SemSpace2019).

\section{Background}
Natural language parsing has been a subject of extensive research during the last decades.
Current state-of-the-art relies on recurrent neural networks equipped with complex data structures that are used to rank the potential parsing steps~\\\cite{dyer-etal-2016-recurrent, dozat2016deep, liu-zhang-2017-order}.
It is not uncommon for combinations of models or training tasks to also be used in order to further improve performance~\cite{clark2018semi, fried2017improving}.
Despite their impressive results, such models are usually complicated to reason about, inconvenient to adapt to different domains and computationally expensive, making them difficult to train and employ.

In light of the above, we take an alternative approach towards parsing that seeks to maximally utilize the proof-theoretic properties of our grammar while remaining as simple and cost-efficient as possible.

\section{Framework}
The key insight is that a parse is a proof, and a proof is simply a sequence of logical rule applications.
If we were to model these rules, namely implication introduction and elimination, and follow them in a backwards-search fashion, we would obtain a proof-faithful parser.

An alternative problem formulation arises if one adopts an Abstract Categorial Grammar (ACG) perspective~\cite{de2001towards}.
In brief, ACG posits the modeling of syntax as the homomorphic translation from a source ``abstract'' logic, which specifies syntactic relations by assigning MILL types to lexical constants, onto a target ``surface'' grammar, which specifies the realization of these abstract syntactic elements as strings and functions over them.
From this viewpoint, our types are (dependency-enhanced) abstract syntactic types of the source logic; what remains to be done is to provide (or rather, learn) the postulated homomorphism that accomplishes the translation onto the surface strings.

With the above in mind, we draw an abstract picture of a backwards-search parsing process in Algorithm~\ref{alg:parse}.

\begin{algorithm}{
\caption{Parse Step}\label{alg:parse}
\Comment{Performs a backward step of the proof reconstruction.}
\begin{algorithmic}[1]
\Procedure{ParseStep}{\textsc{Premises}}
    \State $\textsc{Goal}\gets \textsc{InferGoal}(\textsc{Premises})$
    \While{\textsc{CanIntroduce}()}
        \State (\textsc{premises}$, $\textsc{Goal}$) \gets \textsc{ApplyIntro}(\textsc{premises}, \textsc{Goal})$
    \EndWhile
    \State $(\textsc{PremisesLeft}, \textsc{PremisesRight}) \gets \textsc{ApplyElim}(\textsc{Premises}, \textsc{Goal})$
    \State $\textbf{return} \ (\textsc{PremisesLeft}, \textsc{PremisesRight})$
\EndProcedure
\end{algorithmic}
}
\end{algorithm}

The procedure defined may be decomposed into several subroutines, outlined in the next paragraphs.

\paragraph{Goal Inference}
First, given the types at the left-hand side of a judgement as assumptions, we need to obtain the conclusion, i.e. the proof's goal type at the current step.
Although this is not a strict requirement for the general framework, it is mandatory if we wish to perform introduction online, rather than as a post-processing step, which in turn allows us to inform the elimination module of the goal type at each prior time step.

\paragraph{Introduction}
The next step involves determining whether an introduction rule may be applied. 
Generally, any time the goal type is a complex type, we can safely replace it by its result (i.e. the type on the right side of the main implication connective) while adding its argument (i.e. the type on the left side of the main connective) onto the list of premises, a functionality implemented by the \textsc{ApplyIntro} routine.
This is not the case, however, if the main connective carries a name corresponding to a modifier label, or the argument to be introduced was just eliminated at the immediately prior step of the backwards proof search (thus avoiding infinite loops); \textsc{CanIntroduce} can be designed as a minimal stateful program that keeps track of these two conditions and informs the parser accordingly.

\paragraph{Elimination}
Finally, given that our complex types are the signatures of binary functors, elimination may be treated as the splitting of a sequence (\textsc{Premises}) into two disjoint, possibly non-contiguous, subsequences (\textsc{PremisesLeft} and \textsc{PremisesRight}), where the elements of the first together form the argument that is to be consumed by the function formed by the elements of the second.

\paragraph{Overview}
In practical terms, we first provide our system with a phrase (the full sequence of premises), out of which the phrasal type is derived.
After zero or more applications of an introduction rule followed by one application of an elimination rule, the phrase is split into two disjoint (possibly non-contiguous) sequences of premises.
If any of the resulting sequences has an length of one, it corresponds to an axiom leaf; otherwise, it requires further processing, which is accomplished by repeating the same process anew.
This iteration progressively yields a binary branching tree, that is in one-to-one correspondence with the underlying proof constructed bottom-up (with introduction rules telescoped).

\section{Implementation}
\subsection{Count Invariance}
As mentioned earlier, goal inference is only needed if we to perform introduction steps during proof construction and thus keep track of the goal formula at each proof step.
A very simple way to implement the \textsc{InferGoal} routine, which is nevertheless sufficient for our preliminary experiments, is by utilizing the  \emph{count invariance} property~\cite{DBLP:journals/jphil/Benthem91} that requires a balanced number of input and output occurrences of individual atomic types.
Concisely, we may imagine each complex type as a fractional, with the result type as its nominator and the argument type as its denominator.
Out of a multiset of assumptions, the conclusion may then be derived by performing a multiplication over the corresponding types, with fractionals maximally simplified afterwards.
This process (combined with our ability to deterministically arrange functor types via the argument ordering scheme of subsection~\ref{subsec:extraction_alg}) yields a unique type which is generally the correct one, modulo a few edge cases.

Things become less straightforward when considering modifiers; being polymorphic instances of the $\textsc{x} \to \textsc{x}$ scheme, their fractional representations are self-canceling, thus invisible to the count invariance.
This rarely, however, poses a problem, if we take into account that a) the bottom step of the proof will never have a modifier type as its conclusion, given that modifiers are context-specific (for instance, an adjectival phrase in isolation will be typed as $\textsc{ap}$ rather than $\textsc{np} \to \textsc{np}$) and b) the goal type may also be inferred by taking the goal type of the prior parse step and ``subtracting'' the goal type of the eliminated argument of the current parse step.

Even after this alteration, some cases still remain unsettled.
Third or higher order types require hypothetical reasoning over embedded functors, rather than arguments.
If these happen to be modifiers, they are again invisible to the count invariance, but no immediate way of resolving them is now evident.

Count invariance also struggles under conjunction, as our coordinator types do not specify the exact number of conjuncts but only their general structure; i.e. there is an added variable that the type equations need to be solved against, the value of which would specify the exact instantiation of the coordinator type. 

Given the prototype nature of the parsing experiments, we refrain from designing a more complicated algorithm targeted towards these cases.
Instead, for the time being we simply ignore samples that involve such constructions.
For the sake of argument, many directions are possible as future work; for instance utilizing heuristics that account for higher-order lexical type assignments, or even a full replacement with a neural, lexically informed module responsible for type inference.

\subsection{Elimination as Binary Sequence Classification}
\label{subsec:model}
At this point, recall that our grammar specification assumes associativity and commutativity as universally holding, a design choice that limits the type system's complexity and enhances its learnability.
The non-directionality of $\rightarrow$ means that the splitting done by \textsc{ApplyElim} is not deterministic.
In practice, the type assignments made by the supertagger (even when fully correct) may admit more proofs (or parses) than linguistically desired.
To constrain the search space over parses to just those that are linguistically plausible but also constitute valid proofs, our parser needs to resolve ambiguities by incorporating both lexical and type-level information.
Hence, rather than treating \textsc{Premises} as a sequence of types, we instead expand it to sequence of pairs of words and types.
This allows us to distinguish between elements that share the same type but are anchored to different lexical items --- an addition that, together with our dependency-decorated types, allows for a preferential treatment of particular words with respect to certain dependency roles under different contexts.

With this in mind, we choose to model \textsc{ApplyElim} as a neural function.
Under its current specification, the function's task is to split a sequence into two disjoint subsequences; or equivallently, to assign a binary label for each item within the sequence.
Binary sequence classification is an established problem in machine learning literature, which points to the direct applicability of standard recurrent architectures.
Our network is a standard variant of such an architecture; we allow a bidirectional, two-layer deep Gated Recurrent Unit (GRU)~\cite{gru} to iterate over the vectorial representations of the input sequence.
A linear transformation then converts the high-dimensional vectors of the recurrent unit onto a class weight vector in $\mathbb{R}^2$, which are converted onto class probabilities by the softmax function.

To create vectorial representations of our sequence elements, we first apply the ELMo used by the supertagger onto each word of a sentence.
A word's vector $\overrightarrow{w}$ is then given by ELMo(word, context), where context refers to the initial sentence (prior to any eliminations).
By using a contextual embedding that is informed by the full sentence and letting it persist unchanged throughout the proof, we can provide the parser with an implicit notion of the proof's "past" while drastically reducing the computational costs associated with calculating the word vectors at each step.
Next, we convey the type-level information by associating each unique type with a vector.
Recalling that complex types (in Polish notation) are sequences of atomic types and binary connectives, for which we already have embeddings as produced by the supertagger, we construct complex type embeddings by iterating another GRU over the vector sequences that correspond to each complex type.

A word-type pair's vector is then simply the concatenation of its word and type vectors.
In cases where an element participating in an elimination is not lexically grounded (i.e. types generated by prior introduction rules), we simply set its word vector to zeros.
Finally, in order to inform the network of the conclusion type (which we have already derived), we further concatenate its vector onto each word-type pair, essentially converting the recurrence into a function that is parametric with respect to the goal type.

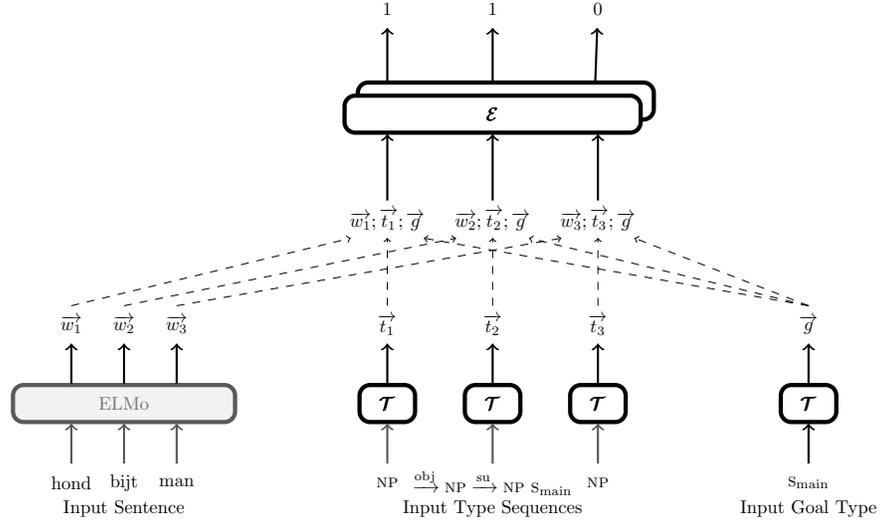
\begin{figure*}[t]
\centering
\begin{tikzpicture}[every text node part/.style={align=center},
 every node/.style={transform shape},
 scale=0.7,
block/.style={rectangle, inner sep=0pt, minimum width=120pt, minimum height=20pt, rounded corners, ultra thick},
str/.style={rectangle, inner sep=0pt, minimum width=120pt, minimum height=20pt},
arrow/.style={->, thick},
pwise/.style={circle, inner sep=0pt, minimum size=10pt},
smallblock/.style={circle, inner sep=5pt, minimum size=12pt, rounded corners, thick},
mediumblock/.style={rectangle, inner sep=0pt, minimum width=30pt, minimum height=20pt, rounded corners, ultra thick},
mini/.style={rectangle, inner sep=0pt, minimum width=30pt, minimum height=20pt, thick, draw=black,}]

%	\node[str] (sentence) at (0, 5) {Input Sentence};
%	\node[str] (symbols) at (7.5, 5) {Input Type Sequences};
%	\node[str] (goal) at (13, 5) {Input Goal};
\node[str] (sentence) at (4, -0.5) {Input Sentence};
\node[str] (dog) at (3, 0) {hond};
\node[str] (bites) at (4, 0) {bijt};
\node[str] (man) at (5, 0) {man};

\node[block, draw=gray!130, fill=gray!10] (elmo) at (4, 1.5) {\textcolor{gray!110}{ELMo}};

\draw (dog) edge[arrow, gray!130] ($(elmo.south) + (-1, 0)$);
\draw (bites) edge[arrow, gray!130] (elmo);
\draw (man) edge[arrow, gray!130] ($(elmo.south) + (1, 0)$);

\node[str] (w1) at (3, 3) {$\overrightarrow{w_1}$};
\node[str] (w2) at (4, 3) {$\overrightarrow{w_2}$};
\node[str] (w3) at (5, 3) {$\overrightarrow{w_3}$};

\draw ($(elmo.north) + (-1, 0)$) edge[arrow, black] (w1);
\draw (elmo) edge[arrow, black] (w2);
\draw ($(elmo.north) + (1, 0)$) edge[arrow, black] (w3);

\node[str] (types) at (11, -0.5) {Input Type Sequences};
\node[str] (su) at (9, 0) {\textsc{np}};
\node[str] (verb) at (11, 0) {$\myrightarrow{obj} \textsc{np} \myrightarrow{su} \textsc{np} \  \textsc{s}_\text{main}$};
\node[str] (obj) at (13, 0) {\textsc{np}};

\node[str] (t1) at (9, 3) {$\overrightarrow{t_1}$};
\node[str] (t2) at (11, 3) {$\overrightarrow{t_2}$};
\node[str] (t3) at (13, 3) {$\overrightarrow{t_3}$};

\node[mediumblock, draw=black] (T1) at (9, 1.5) {$\pmb{\mathcal{T}}$};
\node[mediumblock, draw=black] (T2) at (11, 1.5) {$\pmb{\mathcal{T}}$};
\node[mediumblock, draw=black] (T3) at (13, 1.5) {$\pmb{\mathcal{T}}$};

\draw (su) edge[arrow, gray!130] (T1);
\draw (T1) edge[arrow, black] (t1);
\draw (verb) edge[arrow, gray!130] (T2);
\draw (T2) edge[arrow, black] (t2);
\draw (obj) edge[arrow, gray!130] (T3);
\draw (T3) edge[arrow, black] (t3);

\node[str] (goal) at (17, -0.5) {Input Goal Type};
\node[str] (g) at (17, 0) {$\textsc{s}_\text{main}$};
\node[str] (gt) at (17, 3) {$\overrightarrow{g}$};
\node[mediumblock, draw=black] (T4) at (17, 1.5) {$\pmb{\mathcal{T}}$};
\draw (g) edge[arrow, black] (T4);
\draw (T4) edge[arrow, black] (gt);

\node[str] (wtg1) at (9, 5) {$\overrightarrow{w_1};\overrightarrow{t_1};\overrightarrow{g}$};
\node[str] (wtg2) at (11, 5) {$\overrightarrow{w_2};\overrightarrow{t_2};\overrightarrow{g}$};
\node[str] (wtg3) at (13, 5) {$\overrightarrow{w_3};\overrightarrow{t_3};\overrightarrow{g}$};

\draw (w1.north) edge[arrow, ultra thin, dashed] ($(wtg1.south) + (-0.7, 0)$);
\draw (w2.north) edge[arrow, ultra thin, dashed] ($(wtg2.south) + (-0.7, 0)$);
\draw (w3.north) edge[arrow, ultra thin, dashed] ($(wtg3.south) + (-0.7, 0)$);

\draw (t1.north) edge[arrow, ultra thin, dashed] (wtg1.south);
\draw (t2.north) edge[arrow, ultra thin, dashed] (wtg2.south);
\draw (t3.north) edge[arrow, ultra thin, dashed] (wtg3.south);

\draw (gt.north) edge[arrow, ultra thin, dashed] ($(wtg1.south) + (0.7, 0)$);
\draw (gt.north) edge[arrow, ultra thin, dashed] ($(wtg2.south) + (0.7, 0)$);
\draw (gt.north) edge[arrow, ultra thin, dashed] ($(wtg3.south) + (0.7, 0)$);

\node[block, draw=black, minimum width=160pt] (RNN2) at (11.25, 7.25) {};
\node[block, draw=black, minimum width=160pt, fill=white] (RNN) at (11, 7) {$\pmb{\mathcal{E}}$};
\draw (wtg1) edge[arrow, black] ($(RNN.south) + (-2, 0)$);
\draw (wtg2) edge[arrow, black] ($(RNN.south) + (-0, 0)$);
\draw (wtg3) edge[arrow, black] ($(RNN.south) + (2, 0)$);

\node[str] (l1) at (9, 9) {1};
\node[str] (l2) at (11, 9) {1};
\node[str] (l3) at (13, 9) {0};
-
\draw ($(RNN2.north) + (-2.25, 0)$) edge[arrow, black] (l1);
\draw ($(RNN2.north) + (-0.25, 0)$) edge[arrow, black] (l2);
\draw ($(RNN2.north) + (1.7, 0)$) edge[arrow, black] (l3);

\end{tikzpicture}
\caption[Neural Elimination Module]{The \textsc{ApplyElim} neural architecture, where $\mathcal{T}$ refers to the type-level GRU, and $\mathcal{E}$ to the premise-level GRU. Vector concatenation is denoted by $;$. For the example word input ``hond bijt man'', ``man'' is the element to be eliminated, therefore getting the label 0.}
\label{fig:elimnet}
\end{figure*}

A visual presentation of the network is shown in Figure~\ref{fig:elimnet}.

\subsection{Experiments}
\paragraph{Implementation Details}
We use a 1-layer unidirectional GRU with an input and hidden dimensionality of 1024 as our type embedder.
For the premise-level GRU, we set its hidden dimensionality to 256, its number of layers to 2 and apply a recurrent dropout of 0.5 for regularization.
We train our network using a cross-entropy loss and an AdamW optimizer \cite{adamW} with a learning rate of 10${}^{-3}$ and a weight decay of 10${}^{-4}$.
The source code for the model described can be found at \url{https://github.com/konstantinosKokos/Lassy-TLG-Parsing}.

\paragraph{Training}
To train the network, we precompute the contextualized vectors for each word participating in an elimination. 
Since the word vectors do not change within any single proof, we may then treat each elimination as an independent data point, allowing multiple eliminations (possibly from different sentences) to be processed in parallel.
This gives us the ability to avoid complex data structure manipulation during training time, abolishing the need for CPU instructions that insert computational overhead.
In effect, the neural component of the parser is based solely on highly optimized tensor operations and requires no more than a couple of minutes to train, despite its high expressivity.
This marks a significant improvement over modern parsing architectures, which commonly involve a stack containing partial derivations which is continuously written to and read from, in both the temporal axis (the parse steps) and the sentential axis (the neural iteration over the words).

\paragraph{Inference}
The inference process is identical to training, except for the fact that the input is no longer an independent sample, but rather the production of a previous application of the network (or, the initial phrase).
End-to-end inference has quadratic complexity with respect to the input length (linear number of eliminations, linear iteration complexity per elimination).

\paragraph{Experimental Setup}
We run preliminary experiments on a subset of our data to evaluate the framework's potential.
We limit our experiments to sentences involving types of at most order 2 and no conjunctions for the reasons outlined earlier.
The resulting dataset counts 33\,000 sentences (approximately half of the original), out of which 340\,000 instaces of eliminations are generated.
We train on the first 80\% of the sentences, and report results on the remaining 20\%.
In order to assess the parser in isolation, we use the gold extracted types rather than the types assigned by the supertagger.

\paragraph{Ablations} 
We perform a number of ablations to gain an understanding of the relative influence of each extra source of information to the model.
Table~\ref{table:results} reports our results.

\begin{table}
\centering
\noindent
\newcommand{\ra}[1]{\renewcommand{\arraystretch}{#1}}
\ra{1.1}
\begin{tabularx}{0.75\textwidth}{@{}l|mmmm@{}}
Model \quad & \centering Full & \centering Full-g & \centering Full-g-t & \multicolumn{1}{m}{\centering Full-g-w}  \\
\hline
Accuracy \quad \quad & \centering \textbf{97.15} & \centering 95.3 & \centering 87.77 & \multicolumn{1}{m}{\centering 94.2} \\
\end{tabularx}
\caption[Elimination Module Performance]{Percentage of eliminations correctly analyzed by each model, where Full refers to the model  described in~\ref{subsec:model}, and -g, -t, -w refer to a model where the goal, type and word-level inputs have been removed respectively.}
\label{table:results}
\end{table}

\paragraph{Analysis}
As all other parsing subroutines considered are deterministic, the neural elimination module is the factor that decides the upper boundary of our system's performance.
Evidently, the model achieves highly competitive accuracy scores despite its simple formulation and implementation.
Further, it manages to successfully incorporate all informational sources, as suggested by its increased performance when allowed access to more inputs.
When trained only on word or type information, it still manages adequate accuracy scores, attesting to the high quality of both the word vectors as yielded by the pretrained ELMo, and also the rich informational content of the atomic embeddings as produced by the supertagger.

The raw numbers presented are of course not fully indicative of the system's performance in a realistic setting, as many potential error sources have been artificially removed.
First off, the types considered are the correct ones, whereas they would normally need to be produced by the supertagger.
To increase its robustness against wrong tags, the training process could emulate these by confounding the input types in a controlled manner (alternatively, the two could be jointly trained on a shared portion of the dataset).
Further, eliminations are performed independently, meaning that errors from prior eliminations within a sentence are not propagated upwards through the proof.
Finally, the sentences analyzed are on the easier side of the spectrum, as they involve either no or limited hypothetical reasoning and no conjunctions (which were also the major source of error for the supertagger).
Nevertheless, these first results testify to the strength of the general approach and show that an undirectional type logic provides fertile grounds for parsing, confirming our original hypotheses and earlier design decisions.

\section{Summary}
In this chapter, we introduced an abstract framework for the construction of type-logical proofs in a backwards fashion.
We then instantiated this framework with a basic goal inference component capable of dealing with a portion of possible input judgements, and an elimination component in the form of a simple recurrent network.
This elementary instantiation serves as a concrete proof-of-concept rather than a completely developed parser.
Yet, its very high preliminary results are a clear indication that efficient and highly accurate parsing with our undirectional type logic (and the type lexicon extracted) is within reach.

\chapter{Conclusion}
\label{chapter:conclusion}
This thesis set out to design, extract and experimentally validate a type-logical grammar for written Dutch, aimed towards semantic compositionality.

We began by introducing type-logical grammars and the recurring patterns within them.
Having weighted the pros and cons between established variants, we chose to base our grammar on the Lambek-van Benthem Calculus.
This decision gave our grammar a direct equivalence to the simply-typed $\lambda$-calculus, making it highly fitted for future semantic interpretations.
It also trivialized the treatment of crossing and long-range dependencies, as well as any issues pertaining to word order permutations.
At the expense of these benefits, our type logic becomes hard to perform proof-search over and permits more derivations than the language allows.
To reconcile these shortcomings, we enriched the logic with dependency annotations, anticipating their future utilization in coordination with word-level information.

Next, we implemented an algorithm tasked with performing type assignments on the syntactically annotated sentences of Lassy, the written Dutch corpus, according to our type logic.
To ameliorate incompatiblities between our desired analyses and the ones provided by the corpus, we layed out a number of corpus transformation, each specific to a particular syntactic construction.
By applying them to the corpus we utilized otherwise unusable sentences, ensuring the maximum number of type assignments without making any compromises on their quality.

Albeit our designed grammar's lack of directionality, the type system still proved highly refined, owing to the numerous dependency decorations and atomic types but also the large variety in type structures.
Its fine-grained nature had the side effect of a previously unseen degree of type sparsity, challenging its learnability by standard supertagging architectures.
Rather than ignore rare types or artificially deflate the type system's complexity, we proposed a generative, attention-based supertagging architecture.
Our model proved able to fully acquire the type syntax, learning to construct types inductively, thus bypassing the inherent limitations of established models with respect to type sparsity.

Finally, in order to ascertain the type system's potential as a backbone to parsing, we ran some first experiments on backwards proof-search using a simple recurrent architecture, simultaneously informed by word- and type-level information.
The network proved highly optimal, both in terms of computational efficiency and raw performance.
Although incomplete, these first results suggest that, despite their lack of directionality, our grammar's types suffice for structural disambiguation, when used in tandem with the lexical content of a sentence.

To review, this thesis has produced a dependency-aware type-logical grammar variant, a means for its data-driven extraction, a methodology for constructive supertagging and a (still in the works) framework for structurally ambiguous type-logical parsing.

\paragraph{Future Directions}
Some of the research questions posed have only been given partial answers.
Regarding parsing, it remains to be seen how to best integrate supertagging and parsing, and how to overcome the limitations of the current parsing framework.
Regarding the extraction, further work on exception cases would now yield diminishing returns; however, any bit of extra coverage, no matter how small, is worth the effort as it turns the grammar more general.
Investigating non-compositional phenomena (i.e. unheaded structures) is also an open challenge.
A great degree of empirical experimentation may still be performed over the extraction parameters (e.g. reducing the range of the atomic type or implication label translation tables) to evaluate their effect on downstream tasks.
On a related note, a sanity check could be carried out using pure ILL types to assert the significance of the dependency decorations in parsing.
The positive effect is an opposing force to the type system's complexity (and the associated drops in supertagging accuracy); thus the two need to be carefully weighed to find an optimal balance.

Aside from these questions, the thesis opens many possible directions for future explorations.
Given the experimentally validated treebank, the principal focus now turns on how to best utilize type-logical derivations and their corresponding computational terms to inform semantics; i.e. which candidate semantic interpretations exist for our types and their interactions, and how can they best benefit from this work?
Moreover, it is important to consider the broader applicability of the methods presented; e.g. whether other languages could benefit from non-directional type logics, or other grammar formalisms from a constructive supertagging paradigm.

Such inquiries are no small endeavour and could not be fit under a single thesis.
We hope, however, to have sufficiently motivated the interested reader in pondering the beauty of a type-theoretic view on language, the means through which it can be practically employed and its potential implications for large-scale natural language processing.

\bibliographystyle{alpha}
\bibliography{sources}
\end{document}